%% file: main.tex
\documentclass[10pt,twocolumn,letterpaper]{article}
\usepackage{iccv}              

\definecolor{confblue}{rgb}{0.21,0.49,0.74}
\usepackage[pagebackref,breaklinks,colorlinks,allcolors=confblue]{hyperref}
\usepackage[pdftex,outline]{contour}

\usepackage{multirow}
\usepackage{multicol}
\usepackage{makecell}
\usepackage{amsmath}
\usepackage{stmaryrd}
\usepackage{tikz}
\usepackage{siunitx}
\usepackage{fontawesome5}

\newcommand{\green}[1]{\textcolor{ForestGreen}{#1}}
\newcommand{\venue}[1]{{\tiny\textcolor{gray}{[#1]}}}
\newcommand*{\y}{\checkmark}
\newcommand*{\n}{\textcolor{gray}{--}}
\newcommand*{\blarrow}{\rotatebox[origin=c]{180}{$\Lsh$}}
\newcommand{\ours}{DEAR}
\newcommand{\ourspp}{DEARLi}
\newcommand{\pens}{$\mathbf{P}_\mathrm{ENS}$}

\newcommand{\gray}[1]{\textcolor{gray}{#1}}

\newcommand{\thingtt}{\textit{thing}}
\newcommand{\stufftt}{\textit{stuff}}
\newcommand{\voidtt}{\textit{background}}

\newcommand{\myres}[2]{{#1}$\times${#2}}
\DeclareSIUnit\k{k}
\DeclareSIUnit{\pp}{\textup{p.p.}}

\definecolor{gold}{HTML}{FBF2D2}
\definecolor{silver}{HTML}{DDDDDD}
\definecolor{bronze}{HTML}{EED2B8}
\definecolor{goldD}{HTML}{D9AE13}
\definecolor{silverD}{HTML}{909090}
\definecolor{bronzeD}{HTML}{9A5F26}

\newcommand{\medal}[3]{\tikz[anchor=base, baseline]{\node[rounded corners=2pt,fill=#1,draw=#2,inner sep=1.5pt] {#3};}}
\newcommand{\bm}[2]{
    \ifcase#1\or
    {\medal{gold}{goldD}{\textbf{#2}}}
    \or 
    {\medal{silver}{silverD}{#2}}
    \or 
    {\medal{bronze}{bronzeD}{#2}}
    \else 
        #2
    \fi\ignorespaces
}

\definecolor{ggreen}{HTML}{11F083}
\definecolor{gyellow}{HTML}{F4B400}
\definecolor{gblue}{HTML}{4285F4}
\definecolor{gred}{HTML}{DB4437}

\definecolor{chaircolor}{HTML}{CC4603}
\definecolor{benchcolor}{HTML}{C2FF00}
\definecolor{armchaircolor}{HTML}{08FFD6}
\definecolor{bedcolor}{HTML}{CC05FF}
\definecolor{pillowcolor}{HTML}{00EBFF}

\newcommand{\drawboundingboxes}[3]{%
    \begin{tikzpicture}
    \node[anchor=south west, inner sep=0] (image) at (0,0) {#1};
    \begin{scope}[x={(image.south east)}, y={(image.north west)}]
    \foreach \xone/\yone/\xtwo/\ytwo/\bbcolor/\labeltext in {#3} {
        \draw[rounded corners,line width=0.5mm,\bbcolor] (\xone, {1-\yone}) rectangle (\xtwo, {1-\ytwo});
    }
    \foreach \xc/\yc/\bbcolor/\labeltext/\labcolor in {#2} {
        \node[inner sep=2pt,anchor=center,text=\bbcolor,font=\sffamily\tiny,opacity=0.5,text opacity=1.0,rounded corners,fill=\labcolor]
        at (\xc, {1-\yc})
           {\labeltext};
    }
    \end{scope}
    \end{tikzpicture}
}

\title{DEARLi: Decoupled Enhancement of Recognition and Localization\\ for Semi-supervised Panoptic Segmentation}

\author{%
Ivan Martinović\textsuperscript{1}
\qquad
Josip Šarić\textsuperscript{2} 
\qquad
Marin Oršić\textsuperscript{1} 
\qquad
Matej Kristan\textsuperscript{2} 
\qquad
Siniša Šegvić\textsuperscript{1}\\
    {\normalsize \begin{tabular}{cc}
         {\textsuperscript{1}Faculty of Electrical Engineering and Computing}&
         {\textsuperscript{2}Faculty of Computer and Information Science}  \\
         University of Zagreb & University of Ljubljana \\
         {\tt\small name.surname@fer.hr} & {\tt\small name.surname@fri.uni-lj.si}\\
    \end{tabular}}
}

\begin{document}
\maketitle
\input{sections/0_abstract}    
\input{sections/1_introduction}
\input{sections/2_related_work}
\input{sections/3_method}

\input{sections/4_experiments}

\input{sections/6_conclusion}
\input{sections/acknowledgments}
\small
\bibliographystyle{ieeenat_fullname}
\bibliography{main}

\input{sections/appendix}
\end{document}

%% file: sections/0_abstract.tex
\begin{abstract}
Pixel-level annotation is expensive and time-consuming.
Semi-supervised segmentation methods
address this challenge by learning models
on few labeled images alongside
a large corpus of unlabeled images.
Although foundation models
could further account for label scarcity,
effective mechanisms for
their exploitation remain underexplored.
We address this by devising
a novel semi-supervised panoptic approach
fueled by two dedicated foundation models.
We enhance recognition by complementing
unsupervised mask-transformer consistency 
with zero-shot classification of CLIP features.
We enhance localization
by class-agnostic decoder warm-up
with respect to SAM pseudo-labels.
The resulting decoupled enhancement of recognition
and localization (\ourspp{})
particularly excels
in the most challenging semi-supervised scenarios
with large taxonomies and limited labeled data.
Moreover, \ourspp{} outperforms
the state of the art
in semi-supervised semantic segmentation
by a large margin
while requiring 8$\times$ less GPU memory,
in spite of being trained only for the panoptic objective.
We observe 29.9 PQ
and 38.9 mIoU on ADE20K
with only 158 labeled images.
The source code
is available at \texttt{\href{https://github.com/helen1c/DEARLi}{github.com/helen1c/DEARLi}}.
\end{abstract}

%% file: sections/1_introduction.tex
\section{Introduction}\label{sec:introduction}
Panoptic segmentation
assigns a semantic label
to each image pixel
while also distinguishing
instances of object classes~\cite{kirillov19cvpr}.
This is a 
core capability required
across a wide range of applications 
such as autonomous driving~\cite{zendel22cvpr},
remote sensing~\cite{carvalho22rsensing},
medical imaging~\cite{cha21jcm},
and maritime obstacle detection~\cite{zust23iccv}.
Strong application potential
has driven considerable
research efforts~\cite{kirillov19cvpr2,cheng20cvpr,cheng21neurips},
which relied on
vast amounts 
of annotated data
~\cite{lin14eccv,cordts16cvpr}.
However, 
manual panoptic annotation 
is a time-consuming 
and tedious task, 
often requiring 
more than an hour
per image~\cite{cordts16cvpr,sakaridis21iccv}.

\input{fig/fig_tex/01_fig_1}

In related fields such as
image classification~\cite{sohn20neurips},
optical flow estimation~\cite{lai17neurips}, 
object detection~\cite{xu21iccv}, 
and semantic segmentation~\cite{ouali20cvpr},
semi-supervised learning has been extensively studied 
to reduce the data annotation requirements.
However, only a few 
studies have considered
semi-supervised 
panoptic segmentation~\cite{li18eccv,chen20eccv,qi22eccv,hu23iccv}.
Importantly, these studies
do not tackle the
most challenging yet
practical scenario,
which involves 
very few
labeled images and
large class taxonomies.
This setup makes
labeled examples
per class extremely
scarce and thus
requires methods with
very strong generalization
capabilities.

Foundation models offer 
a unique source of 
auxiliary learning signal 
that could be exploited 
to address the exposed issue.
In particular, 
their robust representations
obtained through 
large-scale pretraining
can provide potential 
to improve 
generalization for 
underrepresented classes.
Recently, SemiVL~\cite{hoyer24eccv} leveraged
CLIP~\cite{radford21icml} 
for backbone 
initialization
and
pseudo-label acquisition
in semi-supervised
semantics. 
However, dense zero-shot 
CLIP predictions
are considerably noisy
due to poor localization~\cite{xu23cvpr,yu23neurips}. 

Thus, new techniques 
are required
to selectively pry out the skills 
that are embedded in foundation models
and relevant for a specific 
performance aspect 
of the trained network.
Effective knowledge 
distillation for 
panoptic segmentation,
in particular
should focus on two 
key aspects: 
(i) recognition, and (ii) localization.
Distilling recognition should 
aid generalization across 
a diverse class taxonomy, 
while distilling localization should 
improve detection of 
both semantic and instance boundaries,
thereby mitigating biases 
introduced by 
learning from
limited annotations.

We address the 
aforementioned challenges 
by proposing
\underline{D}ecoupled 
\underline{E}nh\underline{A}ncement
of \underline{R}ecognition 
and \underline{L}ocal\underline{i}zation
(\ourspp{}) 
-- a novel semi-supervised
panoptic segmentation method 
driven by 
orthogonal contribution 
from two dedicated foundation models
within the mask transformer framework~\cite{cheng21neurips}.
First, we exploit
a vision-language model~\cite{radford21icml}
exclusively for
recognition signals
by ensembling its
zero-shot
mask-wide posteriors~\cite{xu23cvpr}
with mask-transformer
classification.
Second, we leverage
class-agnostic SAM~\cite{kirillov23iccv}
to generate segmentation signals.
We achieve this 
through decoder warm-up
and show that it is possible to
improve a mask transformer
with localization-specific
pre-training.
As shown in \Cref{fig:figure_1_predictions}, 
the recognition-enhanced model \ours{} 
improves mask classification, 
while the localization enhancement in \ourspp{} 
further improves the mask segmentation.

Beyond our methodological contributions,
we report
the first extensive study
on semi-supervised 
panoptic segmentation
in scenarios 
with low annotation budgets
and rich class taxonomies.
The results indicate
that semi-supervised \ourspp{}
consistently outperforms supervised
counterparts trained on
4$\times$ more labeled data.
On ADE20K, \ours{} surpasses
the state of the art
in semi-supervised semantic
segmentation~\cite{hoyer24eccv}
by 7 mIoU points on average,
despite being trained
for the panoptic objective.
Remarkably, our method achieves
these gains using 
8$\times$ less GPUs.

%% file: fig/fig_tex/01_fig_1.tex
\newcommand{\bedxone}     {0.06}
\newcommand{\bedyone}     {0.32}
\newcommand{\bedxtwo}     {0.28}
\newcommand{\bedytwo}     {0.51}
\newcommand{\benchxone}   {0.35}
\newcommand{\benchyone}   {0.56}
\newcommand{\benchxtwo}   {0.67}
\newcommand{\benchytwo}   {0.91}
\newcommand{\chairxone}   {0.72}
\newcommand{\chairyone}   {0.58}
\newcommand{\chairxtwo}   {0.98}
\newcommand{\chairytwo}   {0.98}
\newcommand{\chairtwoxone}{0.78}
\newcommand{\chairtwoyone}{0.64}
\newcommand{\chairtwoytwo}{0.98}
\newcommand{\chairtwoxtwo}{0.93}

\newcommand{\bedxc}{0.26}
\newcommand{\bedyc}{0.72}
\newcommand{\benchxc}{0.54}
\newcommand{\benchyc}{0.80}
\newcommand{\chaironexc}{0.87}
\newcommand{\chaironeyc}{0.91}
\newcommand{\chairtwoxc}{0.90}
\newcommand{\chairtwoyc}{0.64}

\newcommand{\armchairstr}{armchair}
\newcommand{\bedstr}{bed}
\newcommand{\benchstr}{bench}
\newcommand{\chairstr}{chair}
\newcommand{\voidstr}{not found}

\newcommand{\ffigw}{0.47\linewidth}
\newcommand{\tsize}{\footnotesize}
 \begin{figure}[t]
     \centering
     \begin{tabular}{@{}c@{\hspace{4pt}}c@{}}
        \drawboundingboxes
        {
            \includegraphics[width=\ffigw]{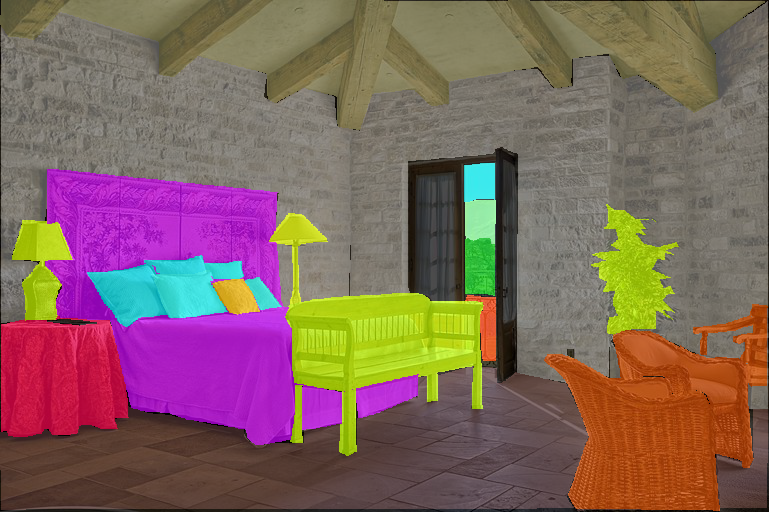}
        }
        {
            \bedxc/\bedyc/bedcolor/\bedstr/black,
            \benchxc/\benchyc/benchcolor/\benchstr/black,
            \chaironexc/\chaironeyc/chaircolor/\chairstr/white,
            \chairtwoxc/\chairtwoyc/chaircolor/\chairstr/white}
        {}
        &
        \drawboundingboxes
        {
            \includegraphics[width=\ffigw]{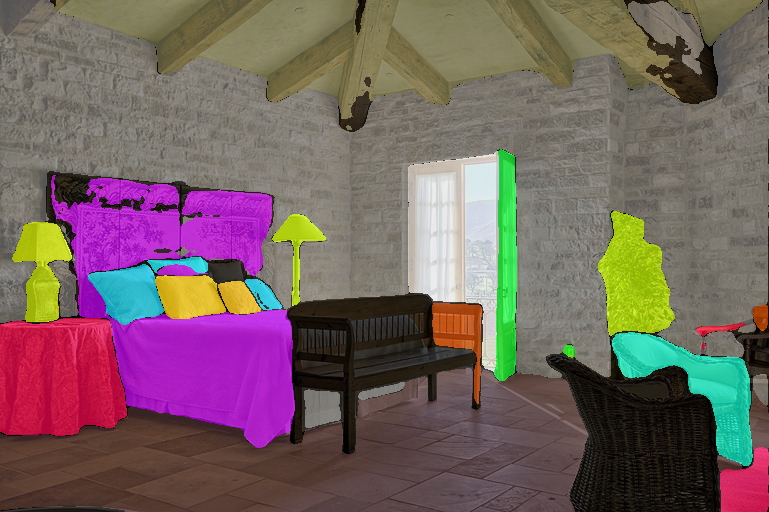}
        }{
            \bedxc/\bedyc/bedcolor/\bedstr/black,
            \benchxc/\benchyc/black/\voidstr/white,
            {\chaironexc-0.02}/{\chaironeyc}/black/\voidstr/white,
            {\chairtwoxc - 0.05}/\chairtwoyc/armchaircolor/\armchairstr/black}
        {
            \bedxone/\bedyone/\bedxtwo/\bedytwo/gred/\bedstr,
            \benchxone/\benchyone/\benchxtwo/\benchytwo/gred/\benchstr,
            \chairxone/\chairyone/\chairxtwo/\chairytwo/gred/\chairstr
        }
        \\[-0.4em]
        \tsize Ground truth 
       & \tsize Semi-supervised M2F~\cite{cheng22cvpr} \\

        \drawboundingboxes
        {
            \includegraphics[width=\ffigw]{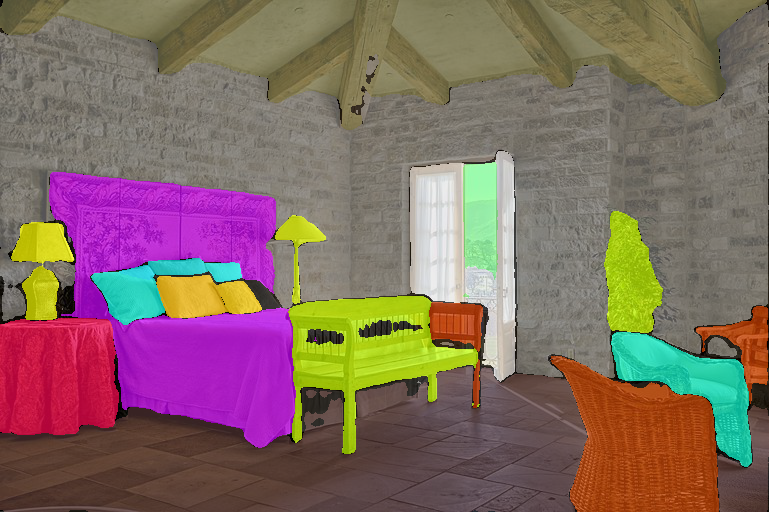}
        }{
            \bedxc/\bedyc/bedcolor/\bedstr/black,
            \benchxc/\benchyc/benchcolor/\benchstr/black,
            {\chaironexc-0.04}/{\chaironeyc}/chaircolor/\chairstr/white,
            {\chairtwoxc - 0.05}/\chairtwoyc/armchaircolor/\armchairstr/black}
        {
           \bedxone/\bedyone/\bedxtwo/\bedytwo/ggreen,
            \benchxone/\benchyone/\benchxtwo/\benchytwo/gred/\benchstr,
            \chairxone/\chairyone/\chairxtwo/\chairytwo/gred/\chairstr
        }
        &
        \drawboundingboxes
        {
            \includegraphics[width=\ffigw]{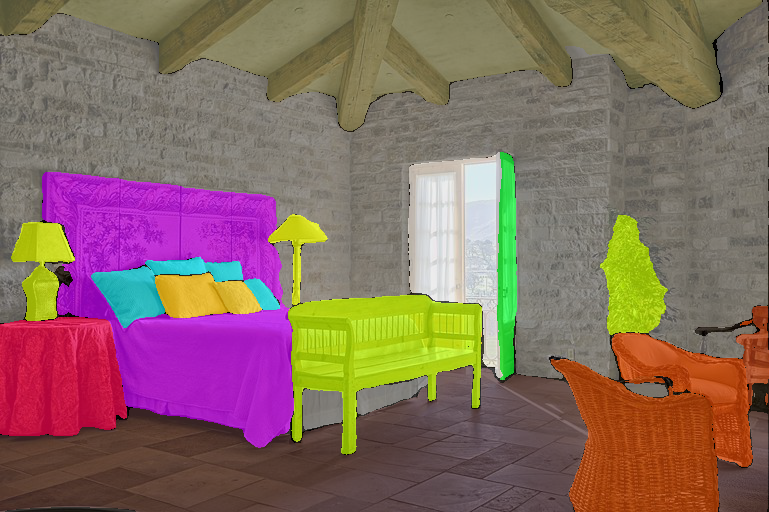}
        }{
            \bedxc/\bedyc/bedcolor/\bedstr/black,
            \benchxc/\benchyc/benchcolor/\benchstr/black,
            \chaironexc/\chaironeyc/chaircolor/\chairstr/white,
            \chairtwoxc/\chairtwoyc/chaircolor/\chairstr/white}
        {
            \bedxone/\bedyone/\bedxtwo/\bedytwo/ggreen/\bedstr,
            \benchxone/\benchyone/\benchxtwo/\benchytwo/ggreen/\benchstr,
            \chairxone/\chairyone/\chairxtwo/\chairytwo/ggreen/\chairstr
        }
        \\[-0.4em]
        \tsize \ours & \tsize \ourspp{} \\
     \end{tabular}
    \caption{
        Unlike semi-supervised-trained
        state-of-the-art panoptic model~\cite{cheng22cvpr},
        our recognition-enhanced method \ours{}
        correctly detects \emph{chair} and \emph{bench} segments, 
        while additional localization enhancement in \ourspp{} 
        further improves masks accuracy. 
     }
     \label{fig:figure_1_predictions}
 \end{figure}

%% file: sections/2_related_work.tex
\section{Related Work}
\label{sec:related_work}

\textbf{Panoptic segmentation.}
Early panoptic approaches either
adapt instance~\cite{he17iccv,kirillov19cvpr}
or semantic segmentation
methods~\cite{chen18eccv,cheng20cvpr}.
These approaches introduce
additional outputs that
require expensive
heuristic-based decoding.
Inefficient modeling can be
avoided by associating instance pixels with
object queries~\cite{carion20eccv}
through cross-attention~\cite{vaswani17neurips}.
This incorporates a
special kind of
inductive bias where
similar pixels
of an object class
should belong to
the same instance.
Mask Transformers (MT)~\cite{cheng21neurips,wang21cvpr,cheng22cvpr,yu22eccv}
apply this idea to
the panoptic segmentation task.
MaskFormer~\cite{cheng21neurips}
produces a fixed number
of mask embeddings
together with the
corresponding
mask-wide class posteriors. 
It recovers dense
mask-assignment
maps by scoring high resolution
features with mask embeddings.
Mask2Former (M2F)~\cite{cheng22cvpr}
drives mask embeddings towards
particular instances by attending
queries to multiscale
features through
masked cross-attention.
Our method builds upon Mask2Former
due to reasonable priors,
high accuracy and
acceptable training efficiency.
Our multi-stage learning
pipeline (\cref{fig:ssl-pipeline})
extends mask transformers with 
orthogonal recognition
and localization enhancement,
which was not previously explored.

\textbf{Dense prediction using vision-language models.}
Joint language-image
representations~\cite{radford21icml,jia21pmlr}
led to advances in tasks
such as text-to-image generation~\cite{ramesh21icml,rombach22cvpr},
zero-shot classification~\cite{radford21icml,jia21pmlr,wortsman22cvpr,zhai23iccv},
and captioning~\cite{yang23cvpr2}.
However, nonlinear attention
before the output
projection hampers
dense spatial
inference with
vision transformers~\cite{li22iclr,zhou22eccv,wang24eccv,lan24eccv}.
Therefore,
we pair the M2F decoder with 
frozen ConvNeXt-CLIP~\cite{liu22cvpr,cherti23cvpr}
backbone
and collect per-mask
embeddings from frozen
features by mask pooling~\cite{ghiasi22eccv,xu23cvpr,yu23neurips}.
Our model architecture
is similar to FC-CLIP~\cite{yu23neurips}
that considers
supervised open-vocabulary
segmentation~\cite{liang23cvpr,ding22cvpr,jiao23neurips},
but does not 
learn on unlabeled images.
FC-CLIP employs a frozen
CLIP backbone with modified M2F decoders.
In contrast,
we propose class-agnostic
pre-training of the
standard M2F decoder,
and show that
geometric ensembling
of M2F and CLIP posteriors~\cite{yu23neurips}
excels on
underrepresented classes in the
semi-supervised context. 

\textbf{Semi-supervised Segmentation.}
\input{fig/fig_tex/03_figure_method}
Only a few methods
consider semi-supervised
panoptics~\cite{chen20eccv,qi22eccv,martinovic2024mc}.
However, these approaches
rely on a substantial amount
of annotated data 
and
outdated backbones.
In contrast, we focus on 
low-label regimes 
in combination with rich 
class taxonomies. 
Our setup enables 
direct comparison
with the related semantic
segmentation works,
as discussed next.
These approaches leverage consistency~\cite{bortsova19miccai,ouali20cvpr,lai21cvpr,grubisic23sensors,mai24cvpr},
co-training~\cite{blum98cclt,qiao18eccv},
self-training~\cite{yuan21iccv,yang22cvpr,ke22itip} and
adversarial training~\cite{goodfellow14neurips,souly17iccv,mittal19pami,qi19cvpr}.
Many of these methods rely on pseudo-labels~\cite{zou21iclr,liang23iccv,wang24cvpr,wu23pami,sun24cvpr}.
However, pseudo-label
quality might
hurt training due to 
poor accuracy
or class imbalance~\cite{zou21iclr,liang23iccv,wu23pami,hu21neurips}.
Our method addresses
this issue through geometric ensembling~\cite{ghiasi22eccv,gu2022iclr,kuo2023iclr,xu23cvpr,yu23neurips}.
Our method is closely
related to methods
that enforce consistency under input augmentations~\cite{xie20neurips,zhong21iccv,sohn20neurips,yang23cvpr}.
We leverage
the Mean Teacher
framework~\cite{tarvainen17neurips}
where the teacher
corresponds to 
the exponential moving average
of the student.
Our teachers receive
weakly perturbed images, 
stop the gradients
and produce hard pseudo-labels~\cite{french20bmvc,sohn20neurips,zhong21iccv,wang24cvpr2,hu24cvpr}.
SemiVL~\cite{hoyer24eccv}
learns consistency~\cite{yang23cvpr}
with CLIP initialization
and introduces a loss to
maintain alignment between
visual and language embeddings.
In contrast, our approach avoids
feature drift
as the backbone remains frozen.
Compared to SemiVL,
our method performs zero-shot
classification on mask-pooled
convolutional CLIP features
instead of individual ViT tokens,
which reduces noise
in pseudo-labels.
In addition, SemiVL’s decoder
builds upon
per-pixel vision-language 
similarities,
which limits its ability
to capture weak correlations.
Finally, our approach supports both
panoptic and semantic segmentation.

%% file: fig/fig_tex/03_figure_method.tex
\begin{figure*}[t!]
	\centering
	\includegraphics[width=\linewidth]{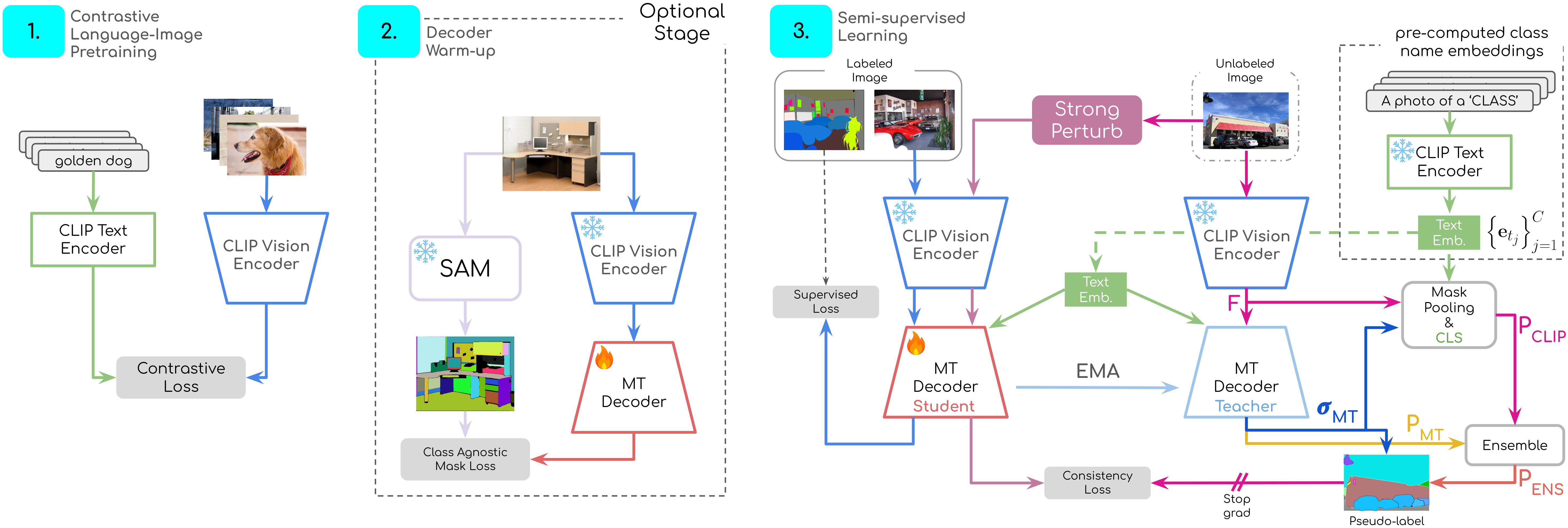}
	\caption{
        Overview of our three-stage
        semi-supervised learning pipeline.
        The first stage corresponds 
        to large-scale contrastive
        language-image
        pre-training (CLIP~\cite{radford21icml,cherti23cvpr}).
        The second stage enhances the 
        localization by class-agnostic
        mask transformer (MT)
        decoder warm-up.
        The third stage trains 
        the decoder on labeled
        and unlabeled images
        with Mean Teacher consistency.
        We enhance the 
        teacher recognition by ensembling the mask-wide posterior of the mask transformer
        with zero-shot classification
        of mask-pooled CLIP features.}
\label{fig:ssl-pipeline}
\end{figure*}

%% file: sections/3_method.tex
\section{Method}
\label{section:decoupled_enhacement_of_recognition_and_localization}
\newcommand{\lsup}{\mathcal{L}^\mathrm{SUP}}
\newcommand{\lunl}{\mathcal{L}^\mathrm{UNL}}
Semi-supervised learning (SSL)
involves a small 
amount of labeled data
$\mathcal{X}^l=\left\{\left(x_i^l, y^l_i\right)\right\}_{i=1}^{N_l}$
and a large amount of unlabeled data 
$\mathcal{X}^u=\left\{x_i^u\right\}_{i=1}^{N_u}$,
where $N_l \ll N_u$.
The goal is to utilize
both subsets to achieve
better generalization performance.
These algorithms typically 
optimize a 
compound loss on
labeled and unlabeled samples:
\begin{align}
    \label{eq:ssl}
	\mathcal{L}^\mathrm{SSL} = \lsup + \lunl.
\end{align}
The first component
corresponds to 
the standard supervised loss 
$\lsup_i = \mathcal{L}(o_i, y^{GT}_i)$,
where $y^{GT}_i$ denotes 
the ground truth
and $o_i=h_{\theta_{stud}}\left(x_i^l\right)$ 
represents the model prediction
computed in the 
corresponding 
labeled example $x_i^l$.
We express the contribution
of the unlabeled data
with the consistency loss
$\lunl_i = \mathcal{L}(o_i^u, \hat{y}^u_i)$~\cite{xie20neurips}.
The model output 
$o_i^u=h_{\theta_{stud}} \left(\mathcal{S}(x_i^u)\right)$
is computed with respect to strongly 
perturbed unlabeled example 
$\mathcal{S}(x_i^u)$,
while the pseudo-label
$\hat{y}^u_i=h_{\theta_{teach}} \left(\mathcal{W}(x_i^u)\right)$
is derived from the same,
but weakly perturbed example $\mathcal{W}(x_i^u)$.
The teacher $h_{\theta_{teach}}$
and the student $h_{\theta_{stud}}$
follow the same architecture
and therefore can be plugged into 
the Mean Teacher framework~\cite{tarvainen17neurips}.
Each iteration t sets $\theta_{teach}$
to the exponential moving average (EMA) 
of $\theta_{stud}$ 
with decay
rate
$\gamma$:
\begin{equation}
        \label{eq:ema}
        \theta_{teach}^{t} = \gamma \theta_{teach}^{t-1} + \left(1-\gamma\right) \theta_{stud}^{t-1}. 
\end{equation}

\Cref{fig:ssl-pipeline}
provides an overview
of our 
training pipeline.
The first stage
corresponds to
large-scale 
contrastive
language-image
pre-training
(CLIP)~\cite{radford21icml}.
The resulting model serves
as a backbone 
feature extractor
of our panoptic models. 
These models 
build upon 
mask transformers
~\cite{wang21cvpr, yu22eccv, cheng22cvpr, li23cvpr} 
in order to enable 
elegant integration 
of separate recognition 
and segmentation
foundation models,
as explained in~\ref{subsection:panoptic-mask-transformers}.
The second stage 
enhances the localization
by warming-up 
the mask transformer decoder
on class-agnostic regions, 
which we discuss in~\ref{subsec:m-sam-train}.
The third stage
trains the mask
transformer decoder
on the target dataset
according to the semi-supervised objective (Eq.~\ref{eq:ssl}).
Here, we enhance the recognition
by ensembling the standard
teacher`s mask-wide posteriors
with zero-shot classification 
of mask-pooled CLIP features 
as discussed in~\cref{subsec:m-vlm-integration}.  

\subsection{Panoptic Mask Transformers}
\label{subsection:panoptic-mask-transformers}
Mask transformers~\cite{wang21cvpr, yu22eccv, cheng22cvpr, li23cvpr}
start from $N$ learnable queries,
attend them to 
translation equivariant features,
and predict mask embeddings 
of \thingtt{} and \stufftt{} classes.
Mask embeddings are 
then projected
onto mask-wide logits 
$\mathbf{P} \in \mathbb{R}^{N \times \left(C+1\right)}$
and used to score  
high resolution features
into sigmoid-activated 
localization maps 
$\boldsymbol{\sigma} \in \mathbb{R}^{N \times H \times W}$.
The C+1-th \textit{no-object}
class allows to discard excess queries
since $N$ is usually larger than the 
number of segments in the image.
Building on the set 
prediction framework~\cite{carion20eccv},
mask transformers establish
bipartite matching $\mathcal M$ 
between predicted masks
and ground truth segments
in order to compute the loss.
The compound loss consists of
recognition and localization terms:
\begin{align}
	\label{eq:m2f_loss}
	\mathcal{L} = 
	\sum_i^N 
	\mathcal{L}_{cls}(\mathbf{P}_i, y^{GT}_{\mathcal{M}(i)}) + 
	\hspace{-1.5em}
	\sum_{ y_{\mathcal{M}(i)}^{GT} \neq C + 1}{
		\hspace{-1.5em}
		\mathcal{L}_{loc}(\boldsymbol{\sigma}_i, \boldsymbol{\sigma}^{GT}_{\mathcal{M}(i)})}.
\end{align}
Here $\mathcal{L}_{cls}$ 
denotes segment-wide 
recognition loss
expressed with 
standard cross-entropy,
while $\mathcal{L}_{loc}$ 
corresponds
to per-pixel
localization loss
expressed as a combination
of the dice loss and 
binary cross-entropy.

\subsection{Vision-Language Enhanced Recognition}
\label{subsec:m-vlm-integration}
Mask transformers 
can easily overfit 
to limited labeled data. 
This leads
to biased pseudo-labels
in semi-supervised learning.
To address this issue,
we integrate mask transformer
with a CLIP foundation model,
which has less pronounced
biases due to
large-scale pretraining.

We begin by
extracting image features
with a frozen CLIP backbone.
While parameter freezing
limits the learning capacity, 
it offers several benefits.
First, it preserves zero-shot
mask classification ability
since the model embeds
images into the
joint vision-text
feature space~\cite{radford21icml}.
Second, it decreases 
the risk of overfitting 
and bias absorption 
due to insufficient 
labeled training data~\cite{zhou22eccv,wang24eccv,lan24eccv}.
Third, it
reduces the training footprint
since backpropagation through the CLIP backbone
becomes unnecessary.

Next, 
we fix the final layer weights 
of the mask classifier 
to the class embeddings 
from the CLIP text encoder. 
Consequently, 
the mask decoder must learn 
to map mask embeddings 
into the language space 
for accurate classification.
Built on Mask2Former~\cite{cheng22cvpr}, 
we call this model M2F-Lang.
 
Finally, rather than relying 
solely on the mask transformer 
for pseudolabel generation, 
we ensemble its 
classification probabilities
with those from 
zero-shot evaluated CLIP.
Since CLIP 
provides only 
image-level recognition,
we first describe the
mask pooling~\cite{ghiasi22eccv, xu23cvpr, yu23neurips} operation
that enables zero-shot
mask-level classification.
Given dense CLIP features 
$\mathbf{F}\in \mathbb{R}^{H' \times W' \times D}$ 
and a binary mask 
$\mathbf{M}_i \in {\left\{0, 1\right\}}^{H \times W}$,
mask pooling recovers 
mask-aggregated
visual embedding 
$\mathbf{e}_{v_i}\in \mathbb{R}^D$ 
as follows:
\begin{equation}
	\label{eq:mask-pooling}
	\mathbf{e}_{v_i}=\mathcal{MP}\left(\mathbf{F}, \mathbf{M}_i\right) = \frac{\sum_{r,c}^{HW}\mathbf{F}[r,c,:]\cdot \mathbf{M}_i[r,c]}{\sum_{r,c}^{HW} \mathbf{M}_i[{r,c}]}.
\end{equation}
We downsample the masks to match feature resolution.
The visual embedding \(\mathbf{e}_{v_i}\) 
is then compared against 
a set of class embeddings 
$\left \{ \mathbf{e}_{t_j} \right \}_{j=1}^C$
obtained by applying 
the CLIP text encoder \(\mathcal{T}\) 
to class names. 
This comparison yields 
a class probability distribution, 
computed as the softmax of cosine similarities scaled by a temperature parameter $\tau$:  
\begin{equation}\label{eqn:zero_shot}
    \mathbf{p}_{\mathrm{CLIP}}^i = \mathrm{softmax}([{\mathbf{e}_{v_i}^T\mathbf{e}_{t_1}}, {\mathbf{e}_{v_i}^T \mathbf{e}_{t_2}}, ..., {\mathbf{e}_{v_i}^T \mathbf{e}_{t_C}}], {\tau}).
\end{equation}
The next paragraph explains how CLIP mask-pooling can enhance
pseudo-labels 
for semi-supervised learning.

Given a weakly augmented
unlabeled image
$\mathcal{W}\left(x^u\right)$,
the teacher network first generates
$N$ initial mask candidates.
After removing those
classified as \emph{no object},
a set of $N'$ masks remain,
defined with sigmoid-activated
localization maps
$\boldsymbol{\sigma}_{\mathrm{MT}} \in \mathbb{R}^{N' \times H \times W}$
and class probabilities
$\mathbf{P}_{\mathrm{MT}}\in \mathbb{R}^{N' \times C}$. 
We then calculate a binary mask
$\mathbf{M}_i\in\left\{0,1\right\}^{H \times W}$
for each of the $N'$ masks 
by thresholding the corresponding
sigmoid mask:
$\mathbf{M}_i=\llbracket {\boldsymbol{\sigma}_{\mathrm{MT}}}_i \geq 0.5 \rrbracket$.
Next, we recover 
the cached CLIP features
$\mathbf{F} \in \mathbb{R}^{H' \times W' \times D}$
from the frozen backbone
and retrieve zero-shot
class distributions 
$\mathbf{P}_{\mathrm{CLIP}}\in \mathbb{R}^{N' \times C}$
via mask pooling over all masks.
Finally, 
we determine the
ensembled mask-wide posteriors 
as a weighted geometric mean 
of the mask transformer posteriors
and the zero-shot posteriors (\ref{eqn:zero_shot})
~\cite{ghiasi22eccv,xu23cvpr,yu23neurips}:
\begin{equation}
    \label{eq:ensembling}
    \mathbf{P}_{\mathrm{ENS}} = \left(\mathbf{P}_{\mathrm{MT}}\right)^\alpha \odot \left(\mathbf{P}_\mathrm{CLIP}\right)^{1-\alpha}.
\end{equation}
These ensembled probabilities
are then fed to the 
standard panoptic inference~\cite{cheng21neurips}
to recover hard pseudo-labels
for consistency training.
Such pseudo-label acquisition
exploits only the
recognition signal from CLIP
as boundaries are retrieved 
from the mask-transformer,
which was originally intended.
Although the student learns
from enhanced pseudo-labels,
we observe 
slight improvements
when including the 
ensembling during inference
(\cf \cref{tab:pq_ablations_appendix} in suppl.).

\noindent\textbf{Validating CLIP zero-shot mask classification}.
As a proof of concept,
we test the CLIP zero-shot 
recognition capabilities
in the panoptic context.
In particular, 
we measure the 
panoptic quality
of a model 
with CLIP features
$\mathbf{F}\in \mathbb{R}^{H' \times W' \times D}$ 
and the oracle mask 
proposal generator.
We extract 
dense CLIP features
with MaskCLIP~\cite{zhou22eccv}
for ViT~\cite{dosovitskiy21iclr,radford21icml}
and simply remove
global average pooling 
for ConvNeXt~\cite{cherti23cvpr}.
The oracle 
generator retrieves
a binary mask candidate
$\mathbf{M} \in {\left\{0, 1\right\}}^{H \times W}$
for each ground truth segment.
We recover the per-mask
class distributions
with mask pooling 
(Eqs.~\ref{eq:mask-pooling} and~\ref{eqn:zero_shot}).

\Cref{fig:zero-shot-mask-cls}
presents the resulting 
panoptic quality
on ADE20K and COCO-Panoptic.
We observe that 
zero-shot classification
of mask-pooled CLIP features
delivers attractive performance.
This shows that
mask transformer can 
benefit from the ensembling.
Just as important, 
ConvNeXt models
significantly outperform 
ViT on both datasets.
This motivates us to 
reconsider the ViT backbone
used in the previous
state-of-the-art for 
semi-supervised 
semantics~\cite{hoyer24eccv}.
Inspired by these findings,
we select the frozen OpenCLIP
ConvNeXt~\cite{cherti23cvpr} 
pretrained on
LAION-2B~\cite{schuhmann22neurips} as 
the backbone 
and 
\input{fig/fig_tex/03_fig_teaser}
include zero-shot classification
of mask-pooled CLIP features
in our pseudo-label
generation procedure.

\subsection{Class-agnostic Mask-Decoder Warm-up}
\label{subsec:m-sam-train}
The proposed
ensembling 
with zero-shot predictions
assumes 
adequate mask candidates
with accurate boundaries.
Failure of this assumption 
would devalue the benefits
of the proposed ensembling.
This risk is especially pertinent to
semi-supervised setups  
where the model may struggle 
to learn precise object boundaries 
due to limited labeled data.

We address this challenge
by distilling objectness
from the Segment Anything Model
(SAM)~\cite{kirillov23iccv}.
Given an unlabeled image,
SAM can produce 
a set of class-agnostic
binary masks 
$\{\mathbf{M}_i\}_{i=1}^{N_\mathrm{SAM}}$
corresponding to different 
regions in the image.
A naive approach
to obtain panoptic pseudo-labels
classifies each SAM mask $\mathbf{M}_i$ 
with CLIP using the mask pooling 
(Eqs.~\ref{eq:mask-pooling} and~\ref{eqn:zero_shot}).
However, this approach 
yields only 8.8 PQ on ADE20K. 
This poor performance is a result of a significant granularity
mismatch between SAM predictions and the target dataset,
as illustrated in~\cref{fig:mismatch}.

We propose a simple 
yet effective solution
to leverage SAM's rich 
objectness knowledge
while simultaneously addressing
the granularity mismatch issue.
The key idea 
is to precondition
the mask decoder 
prior to the main
semi-supervised stage. 
Specifically, we first generate
class-agnostic pseudo-labels using SAM
for both labeled images $\mathcal{X}^l$
and unlabeled images $\mathcal{X}^u$.
These class-agnostic pseudo-labels
are then used to warm up the mask decoder,
as illustrated in the middle pane
of~\cref{fig:ssl-pipeline}.
During this warm-up phase,
the optimization focuses solely on
the localization loss term
$\mathcal{L}_{loc}$~(\cf~Eq.~\ref{eq:m2f_loss}),
leaving classification and granularity refinement
to be learned later from 
the available labeled subset.
The proposed Decoder Warm-up (DeWa) stage
enhances both recall and boundary alignment
of the mask decoder,
as demonstrated in our experiments.
\newcommand{\myh}{1.55cm}
\begin{figure}[h]
    \centering
    \begin{tabular}{@{}c@{\hspace{0.025cm}}c@{\hspace{0.1cm}}c@{\hspace{0.025cm}}c@{}}
         \scriptsize Ground-truth& \scriptsize SAM predictions & \scriptsize Ground-truth& \scriptsize SAM predictions \\
         \includegraphics[height=\myh,keepaspectratio]{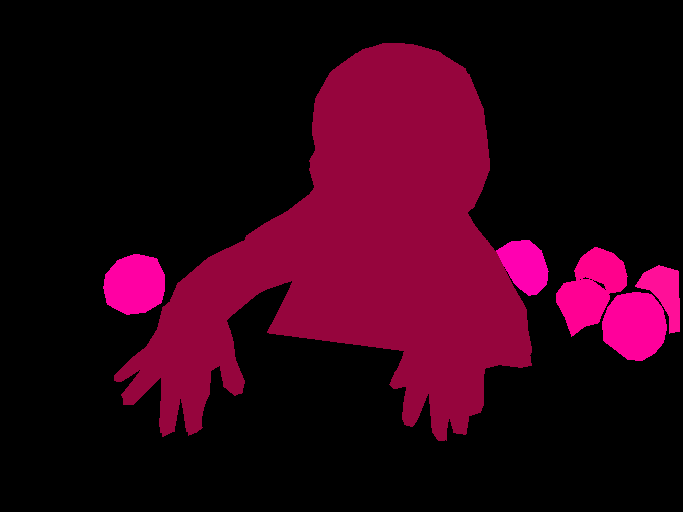} & 
         \includegraphics[height=\myh,keepaspectratio]{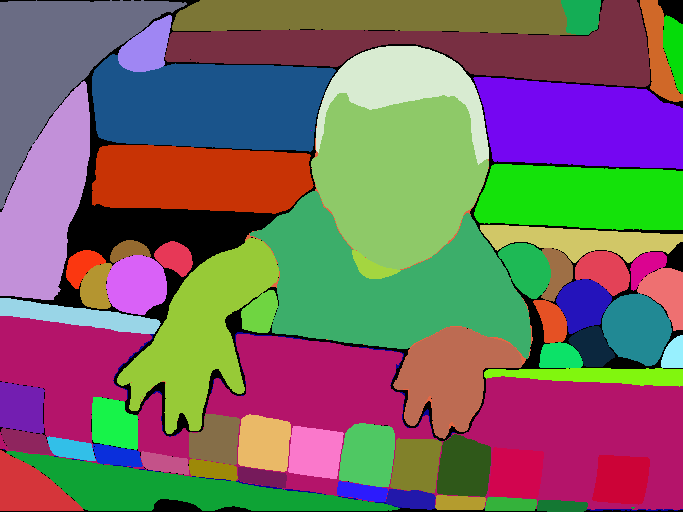}&
         \includegraphics[height=\myh,keepaspectratio]{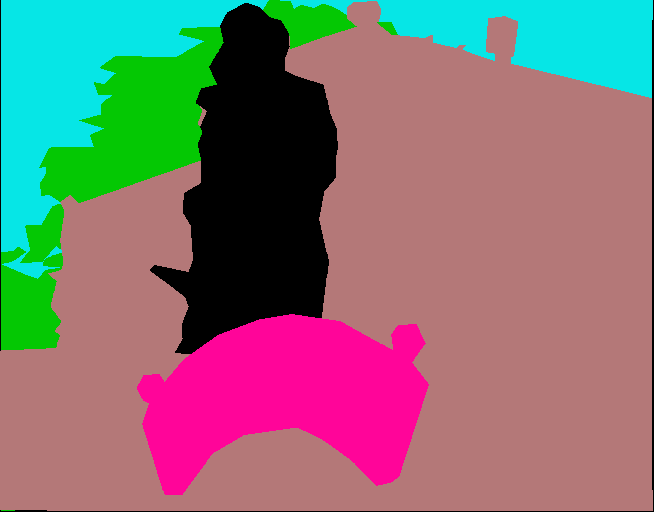}&
         \includegraphics[height=\myh,keepaspectratio]{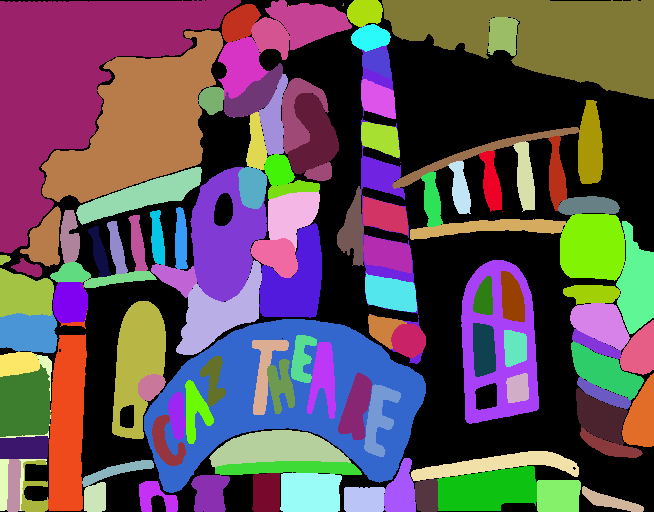} \\
    \end{tabular}
    \caption{Illustration of granularity mismatch between
    ADE20K GT labels and
    class-agnostic predictions generated by SAM~\cite{kirillov23iccv}.}
    \label{fig:mismatch}
\end{figure}

%% file: fig/fig_tex/03_fig_teaser.tex
\begin{figure}[b]
	\centering
	\includegraphics[width=\linewidth]{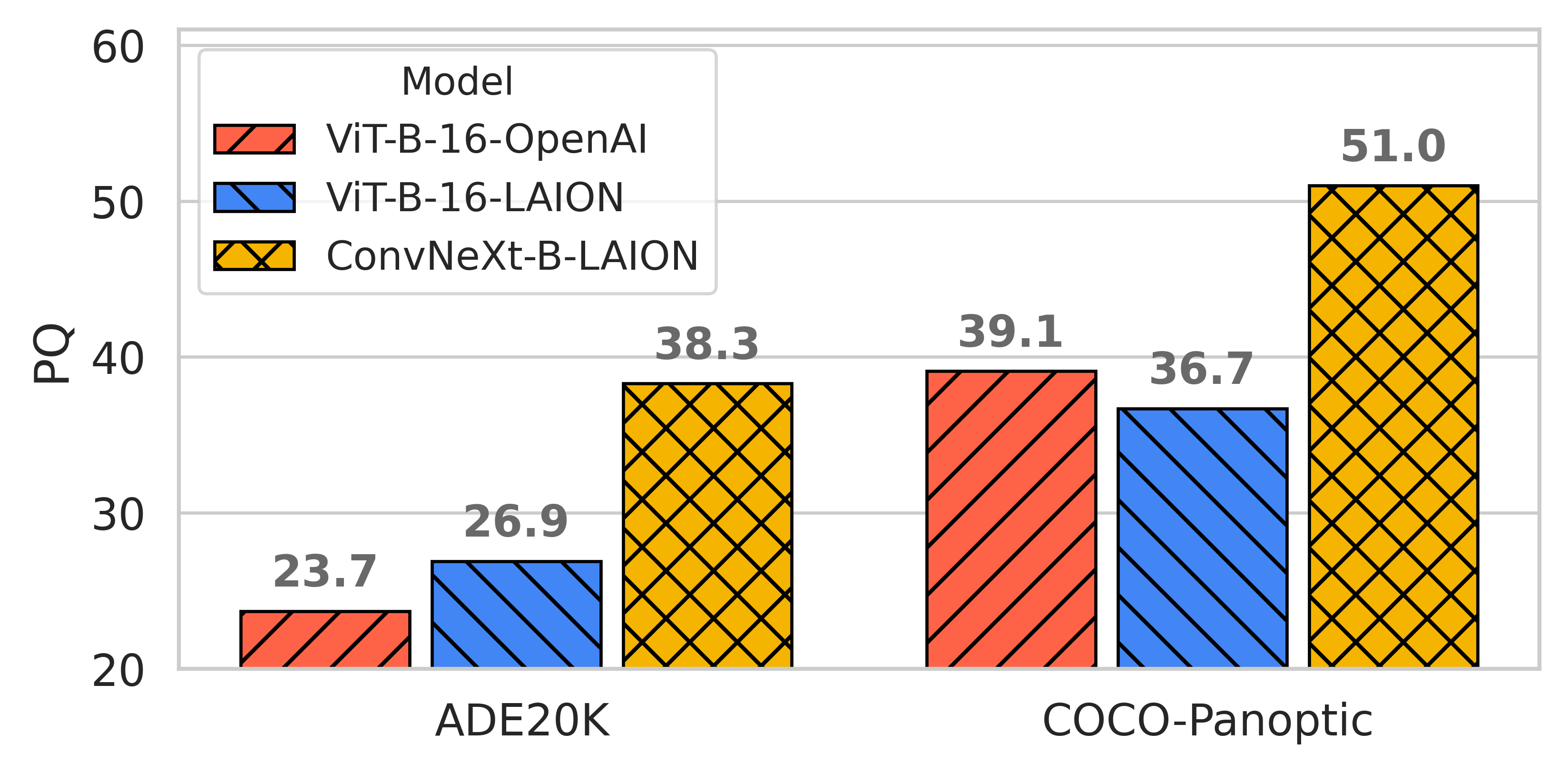}
	\caption{Zero-shot panoptic
                quality with ground-truth masks
                and mask-pooled CLIP
                features.
                Each mask is classified
                with respect to textual CLIP embeddings
                of class name descriptions.}
	\label{fig:zero-shot-mask-cls}
\end{figure}

%% file: sections/4_experiments.tex
\section{Experiments}
\input{tables/04_main_ablation_pq_sq_rq}
\label{sec:experiments}
\textbf{Datasets.}
We conduct experiments on:
ADE20K, COCO-Panoptic, and COCO-Objects.
The ADE20K dataset~\cite{zhou17cvpr}
comprises \num{20210}
training and \num{2000} validation images
with \num{150} semantic classes.
COCO-Panoptic~\cite{lin14eccv} includes
\SI{118}{\k} training and~\SI{5}{\k} validation images,
with \num{80}~\thingtt{}
and \num{53}~\stufftt{} classes.
COCO-Objects~\cite{lin14eccv}
corresponds to the same dataset with
all~\stufftt{} classes
remapped to the \voidtt{} class.
We consider this dataset to ensure fair comparison
with semi-supervised semantic segmentation methods.

\noindent\textbf{Architecture.}
Our experiments
build upon ConvNeXt-B~(CN-B)~\cite{liu22cvpr}
and assume CLIP~\cite{radford21icml,cherti23cvpr}
pre-training on LAION-2B~(L2B)~\cite{schuhmann22neurips}.
We complement this backbone with the
M2F~\cite{cheng22cvpr} decoder.
Models with language-based recognition (M2F-Lang) 
produce classification logits
using dot product between
mask-wide class embeddings
and precomputed language embeddings.
Effectively, 
frozen language embeddings operate as a hand-crafted linear projection.

\noindent\textbf{Training.}
Our semi-supervised approach
uses Mean Teacher~\cite{tarvainen17neurips}
across weakly and strongly
perturbed unlabeled images (\cref{section:decoupled_enhacement_of_recognition_and_localization}).
Weak perturbations are random scaling (\num{0.1} to \num{2.0}),
cropping and horizontal flipping.
Strong perturbations
further include color jittering,
Gaussian blur, grayscaling and CutMix~\cite{yun19iccv}.
We train for~\num{80000}
iterations in all semi-supervised experiments.
Training batches consist of~\num{8} labeled and~\num{8} unlabeled images.
The student and teacher networks share the same architecture.
Teacher network parameters are updated using EMA with $\gamma = 0.999$.

We use identical dataset partitions
as in the prior work~\cite{yang23cvpr,hoyer24eccv}.
We train on crop sizes~\myres{640}{640}
and \myres{512}{512} for ADE and COCO, respectively.
We use the AdamW~\cite{kingma15iclr,loshchilov19iclr}
optimizer with
weight decay~\num{0.05}
and learning rate~\num{0.0001}.
Unless otherwise specified,
we freeze the backbone and train only
the M2F decoder.
We set geometric ensembling
factor $\alpha=0.6$ across
all experiments.
For language-based
recognition, we
adopt prompt templates~\cite{ghiasi22eccv,gu2022iclr,xu23cvpr}.
In M2F-Lang experiments, we do not reject
segments with low max-softmax~\cite{yu23neurips}.
We warm-up the decoder (DeWa) with batch size \num{16}
for~\SI{80}{\k}~and~\SI{160}{\k}~iterations on ADE and COCO, respectively.
Class-agnostic masks
are generated with the 
ViT-H SAM~\cite{kirillov23iccv}
and default
hyperparameters.
All experiments
with a frozen backbone
are conducted on a single A100-\SI{40}{\giga\byte} GPU.

\noindent\textbf{Evaluation.} 
Most of our experiments
report panoptic quality (PQ),
averaged across all dataset classes~\cite{kirillov19cvpr}.
We compare with previous semantic segmentation
approaches in terms of mean
intersection over union (mIoU).
We report performance
of the checkpoint 
from the final training
iteration on the ADE20K/COCO
validation set.

\subsection{Panoptic Segmentation Results and Ablations}
\label{subsection:panseg_results}
\Cref{tab:pq_ablations}~outlines the development
path of our method.
It presents performance
increments and component ablations
from our mask-recognition baseline (M2F)
over the CLIP enhanced~\ours{},
to the CLIP and SAM enhanced~\ourspp{}.
All experiments report PQ performance on ADE20K val
and train on standard
semi-supervised
partitions~\cite{yang23cvpr}.
We evaluate how
CLIP initialization,
language-based recognition,
per-class probability geometric
ensembling~(\pens{})
and decoder warm-up 
(DeWa) contribute to our method.

We observe that
semi-supervised learning (SSL\y)
consistently outperforms
fully supervised counterparts.
This provides a sanity check for further experiments.
Gains from semi-supervised learning
are more pronounced
in low-label data regimes.
Comparison between the first two sections in~\cref{tab:pq_ablations}
reveals that LAION-2B CLIP initialization
consistently surpasses ImageNet1k~\cite{deng09cvpr} initialization
(from~\SI{3}{\pp}~PQ on 1/128 to \SI{6.6}{\pp}~PQ on 1/8).
Backbone fine-tuning improves performance
in supervised experiments across
the first two sections.
However, this roughly doubles the training memory requirements.
More importantly, fine-tuning the backbone
introduces vision-language misalignment
that degrades performance in low-label regimes
(preliminary experiments show \SI{-4.8}{\pp}~PQ on 1/128).
Hence, all experiments from 
the last section freeze the backbone.

The last section focuses
on learning with fixed
language embeddings (M2F-Lang).
Here, mask-wide logits correspond
to cosine similarity between
retrieved mask-wide embeddings and
precomputed language embeddings.
Recall that mask-wide class embeddings
can be recovered either from the M2F decoder
or through mask pooling of CLIP features.
Ensembling the corresponding two posteriors
leads to consistent and substantial improvement
with respect to the decoder-only posterior (M2F-Lang).
M2F-Lang~+~\pens~(eval only) retains some improvement
even when ensembling posteriors
exclusively during inference. 
We observe that semi-supervised learning on unlabeled images
leads to similar or greater improvements
as mask-wide posterior ensembling.
Importantly, 
incorporating geometric ensembling
into pseudo-labels generation 
provides additional significant gains,
as the student benefits from a stronger learning signal.
This leads us to our method,~\ours{},
which consistently improves
upon evaluation-only ensembling
across all partitions.
\ours{} performs especially well
in the most challenging regime
with only 158 labeled images.
Finally, the inclusion
of class-agnostic decoder warm-up (\ourspp{})
further improves the performance
across all data partitions.
This supports our hypothesis
that incorporating
class-agnostic pre-training
with SAM pseudo-labels
can improve final panoptic quality,
despite the granularity mismatch between
SAM pseudo-labels and
ADE20K taxonomy (\cf\cref{fig:mismatch}).
For more comprehensive ablations,
see supplement (\cref{tab:pq_ablations_appendix}) .

\Cref{tab:performance_coco_panoptic} shows similar
findings on COCO-Panoptic.
Posterior ensembling for pseudo-label generation (\ours{})
consistently outperforms baseline across all partitions.
Moreover, SAM distillation within the decoder warm-up
(\ourspp{}) yields further gains,
with a notable +\SI{4.1}{\pp}~PQ
on the most challenging setup with only~\num{232} labeled images.
Note that~\ourspp{} surpasses its supervised counterpart (M2F-Lang+\pens{}) while using 4$\times$ less labeled data.
\input{tables/04_coco-panoptic-pq}
\subsection{Comparison with the State of the Art}
\label{subsec:comparison-with-the-state-of-the-art}
\input{tables/04_sota-comparison-vlm-ade20k}
To the best of our knowledge,
our method is the first to
address semi-supervised
panoptic segmentation
with standard partitioning~\cite{yang23cvpr} of
the common segmentation
datasets with rich taxonomies.
Consequently,
we cannot present
a direct comparison
with previous semi-supervised
panoptic approaches.
Instead, we compare~\ours{}
with the state of the art in semi-supervised semantic segmentation
using the same labeled data partitions.
In order to measure semantic segmentation performance,
we evaluate our panoptic models
with semantic inference, without retraining~\cite{cheng21neurips}.
We consider this
comparison relevant and fair
since panoptic
M2F models typically underperform
with respect to native
semantic segmentation models
on that particular task~\cite{cheng22cvpr}. 
For an exact comparison,
see~\cref{tab:semseg-training} in the supplement.

We are particularly interested
in the comparison with SemiVL~\cite{hoyer24eccv}, 
since it also leverages CLIP.
\Cref{tab:sota_ade,tab:sota_coco} 
indicate that~\ours{}
beats all previous methods by a large margin
across all partitions on ADE20K and COCO-Objects.
We additionally include \ourspp{} in these tables
even though comparison with SemiVL
might be unfair, since \ourspp{}
distills knowledge from SAM.
Nevertheless, it may prove as a useful baseline
for future semi-supervised approaches
with class-agnostic pre-training.

We observe that our baseline M2F+SSL with 
ImageNet1k initialization
outperforms all previous baselines
with the same pre-training,
while even exceeding SemiVL in some assays.
This suggests that M2F+SSL is a strong baseline
and strengthens the value of our contributions.
SemiVL with a convolutional backbone
performs roughly the same as
SemiVL with a transformer backbone (\cf~\cref{tab:sota_ade})
or even worse (\cf~\cref{tab:sota_coco}).
This suggests that the observed advantages 
of the convolutional backbone (\cf~\cref{fig:zero-shot-mask-cls})
are likely due to direct segment prediction 
and mask pooling~\cite{cheng22cvpr, xu23cvpr}
being present in our architecture.
The supplement contains
details of SemiVL~\cite{hoyer24eccv}
training atop CN-B-L2B (\cref{sec:suppl-semivl}),
as well as semantic 
segmentation ablations of \ourspp{}~(\cref{tab:iccv-miou-ablation}).
\\ 
\noindent \textbf{Comparison on COCO-Objects.} 
Many previous works in semi-supervised semantic segmentation 
report experiments on the COCO-Objects dataset, 
which includes \num{80}~\thingtt{} classes and one~\voidtt{} class.
We enable training of our panoptic
models on these~\num{81} classes
by remapping all~\stufftt{} classes from COCO-Panoptic
to the~\voidtt{} class.
\Cref{tab:performance_coco_objects} shows that
class-agnostic decoder warm-up 
consistently enhances panoptic
performance as in previous setups.
Moreover,~\Cref{tab:sota_coco} shows that~\ours{} again
consistently outperforms current state of the art~\cite{hoyer24eccv}
across all partitions, while~\ourspp{} confirms
the advantage from~\Cref{tab:performance_coco_objects}.
A qualitative comparison with
state-of-the-art methods on several examples
is provided in the supplement.
\input{tables/04_coco-objects-pq}
\input{tables/04_sota-comparison-vlm-coco-objects}

\subsection{Additional Ablations and Analysis}
\label{subsection:additional_ablations}
\noindent\textbf{Decoder warm-up.} 
\Cref{tab:ablation_pretraining-stage} 
validates the performance of
our method with different
decoder warm-up procedures.
We keep the backbone frozen.
We start with randomly initialized decoder,
as used in~\ours{}.
Row 2 shows that decoder initialization
with supervised
pre-training
(M2F-Lang from~\cref{tab:pq_ablations})
actually decreases the performance,
which suggests the presence of overfitting.
Row 3 trains on random 110k images from SA1B,
the dataset that was
originally used to train SAM~\cite{kirillov23iccv}.
This experiment follows hyperparameters
from~\ourspp{} (row 4).
Comparable performance of the last two rows 
suggests that~\ours{} benefits from
decoder warm-up,
even in the
presence of domain shift
from the SA1B.
We highlight the
last section improvement
as a valuable contribution
with plenty of applications.

\noindent \textbf{Geometric ensembling ablation.}
\Cref{tab:iccv-ablations-pens} 
validates pseudolabel generation 
with $\mathbf{P}_{\mathrm{MT}}$, 
$\mathbf{P}_{\mathrm{CLIP}}$, 
or $\mathbf{P}_{\mathrm{ENS}}$ on ADE20K.
Here, we use \ours{}
with 
standard M2F inference 
in the student
(w/o ensembling).
We observe consistent performance
improvements with $\mathbf{P}_{\mathrm{ENS}}$, 
which justifies proposed mechanism of 
knowledge distillation from CLIP.
\input{tables/04_additional-ablation-different-initialization}
\input{tables/iccv_table_pmt_ablation}

\noindent \textbf{Gains on underrepresented classes.} 
\Cref{fig:underrepresented-classes-gains} compares M2F+SSL (\cf~\cref{tab:pq_ablations}, 6th row)
with~\ourspp{} when trained
on ADE 1/128 and ADE 1/64
semi-supervised setups.
We group classes by pixel ratio,
with~\num{30} classes per group
from least to most frequent.
Notably,~\SI{20}{\percent} of classes
occupy over~\SI{80}{\percent}
of labeled pixels in both setups,
reflecting a long-tailed distribution.
We observe substantial gains
of~\ourspp{}
in underrepresented classes
and smaller gains
in frequent classes.
As expected, this difference decreases
in partitions containing more labeled data.
\input{fig/fig_tex/04_figure_grouped_improvements_v2}

\noindent \textbf{Increasing the backbone.}
\Cref{fig:increasing-backbone-size} validates 
the panoptic quality
of~\ours{} and~\ourspp{}
on ADE and COCO-Panoptic
when increasing
the frozen backbone size from
ConvNeXt-B to ConvNeXt-L (\ie~88M $\rightarrow$ 197M).
The larger
backbone consistently
improves results.
\ourspp{} surpasses~\ours{} in all configurations.
\ourspp{} particularly excels in 
very low-labeled regimes,
achieving performance
improvements comparable
to doubling the backbone capacity.
The supplement includes exact measurements. 
\input{tables/04_basevslarge_fig}

\noindent\textbf{Validating} $\mathbf{\alpha}$\textbf{.} 
\Cref{tab:analysis_alpha} evaluates panoptic
quality of~\ours{}
on ADE20K as the geometric ensembling
factor $\alpha$ varies.
We find that $\alpha=0.6$
offers the best balance across data splits,
with higher values improving performance
on larger partitions (e.g., 1/8).
This aligns with expectations,
as higher $\alpha$ increases
reliance on learned classification,
which tends to be more accurate
with more labeled data. 
\input{tables/04_additional-ablation_performance-for-different-alphas-ade20k}

%% file: tables/04_main_ablation_pq_sq_rq.tex
\begin{table*}[t]
    \scriptsize
    \centering
    \begin{tabular}{lccc@{\hspace{3pt}}c@{\hspace{3pt}}cc@{\hspace{3pt}}c@{\hspace{3pt}}cc@{\hspace{3pt}}c@{\hspace{3pt}}cc@{\hspace{3pt}}c@{\hspace{3pt}}cc@{\hspace{3pt}}c@{\hspace{3pt}}c}
        \toprule
        \multirow{2}{*}{Method} & \multirow{2}{*}{SSL} & \multirow{2}{*}{Backbone} & \multicolumn{3}{c}{1/128 (158)} & \multicolumn{3}{c}{1/64 (316)} & \multicolumn{3}{c}{1/32 (632)} & \multicolumn{3}{c}{1/16 (1263)} & \multicolumn{3}{c}{1/8 (2526)} \\
        & & & \textbf{mPQ}  & mSQ & mRQ  & \textbf{mPQ}  & mSQ & mRQ & \textbf{mPQ}  & mSQ & mRQ  &  \textbf{mPQ}  & mSQ & mRQ  & \textbf{mPQ}  & mSQ & mRQ \\
        \midrule
        \multirow{3}{*}{M2F}        &\n& \faSnowflake~CN-B-IN1k                 & 7.4 & 32.7 & 9.3 & 13.3  & 52.2 & 16.7 & 17.8 & 62.7& 21.7& 22.1 & 65.1 & 27.0 & 25.5 & 69.3 &30.7 \\
                                    &\n& \faFire     ~CN-B-IN1k                 & 9.3 & 38.6 & 11.7 & 15.0 & 55.2 & 18.6 & 20.5 & 62.9& 25.2& 25.0 & 68.9 & 30.5 & 29.9 & 73.2 &35.9 \\
        & \y & \faFire     ~CN-B-IN1k                 & 16.4 & 45.7 & 20.4 & 22.8 & 60.4 & 27.9 & 27.8 & 70.2 & 34.0 & 30.1 & 71.4 & 36.6 & 33.6 & 75.1 & 40.3 \\
        \midrule
        \multirow{3}{*}{M2F}                                      &\n& \faSnowflake~CN-B-L2B                  & 10.0 & 38.7 & 12.4 & 16.4 & 56.2 & 20.1 & 23.4 & 65.4 & 28.5 & 28.7 & 69.5 & 34.9 & 34.2 & 73.7 & 40.9 \\
                                              &\n& \faFire     ~CN-B-L2B                  & 12.9 & 42.8 & 16.1 & 19.3 & 59.1 & 23.9 & 26.0 & 69.1 & 31.7 & 31.1 & 72.0 & 37.7 & 36.3 & 75.6 & 43.7 \\

                                    & \y &  \faFire    ~CN-B-L2B                  & 19.4 & 46.2 & 24.0 & 28.4 & 63.7 & 34.6 & 33.2 & 71.3 & 40.2 & 36.8 & 75.9 & 44.3 & 40.2 & 78.8 & 48.2 \\
        \midrule
        M2F-Lang                                 &\n& \multirow{6}{*}{\faSnowflake~CN-B-L2B} & 11.7 & 43.7 & 14.7 & 19.1 & 61.4& 23.9 & 26.3 & 70.2&32.1 & 31.6 &72.8 & 38.3& 36.6 & 76.7 & 44.0 \\
         \blarrow \ + \pens{} (eval only)                       &\n&                                        & 18.3 & 63.7 & 23.3 & 23.9 & 69.5 & 30.0 & 29.0 & 73.3 & 35.4 & 33.6 & 75.6 &40.8 & 37.8 &77.6 &45.7 \\
        M2F-Lang                         &\y&                                        & 18.3 & 55.0 & 22.6 & 26.4 & 68.0& 32.3& 32.5 & 76.1 & 39.4 & 35.5& 74.5 &42.8 &39.2 &77.2 &47.2 \\
        \blarrow \ + \pens{} (eval only)           &\y&                                        & {21.6} & 62.6 & 26.7 & {30.2} & 73.9 & 37.0 & {33.9} & 76.2 & 41.1 & 36.7 & 77.6 & 44.2& 40.2 & 77.2 & 48.4 \\
        \quad \blarrow \  + \pens{} (\textbf{\ours})&\y&                                        & \bm2{27.7} & 70.3 & 34.4 & \bm2{32.3} & 74.0 & 39.6 & \bm2{34.8} & 75.4 & 42.1 & \bm2{38.3} & 79.3 & 46.2 & \bm2{40.6} & 80.7 & 48.7 \\
        \qquad \blarrow \ + DeWa (\textbf{\ourspp}) &\y&                                        & \bm1{29.9} & 74.0 & 36.6 & \bm1{34.6} & 75.4 & 41.2 & \bm1{36.3} & 77.1 & 43.5 & \bm1{39.2} & 80.3 & 47.1 & \bm1{41.6} & 80.6 & 49.5 \\
        \bottomrule
    \end{tabular}
    \caption{
    Impact of our contributions to panoptic performance on ADE20K.
    Top two sections start from the supervised baseline
    and show improvements due to backbone fine-tuning 
    (\faSnowflake $\rightarrow$ \faFire),
    semi-supervised learning with Mean Teacher consistency (SSL)
    and LAION-2B pre-training (L2B).
    Bottom section involves 
    language-based recognition (M2F-Lang)
    and shows improvements due to SSL,
    geometric ensembling with CLIP during inference (eval-only)
    and pseduolabels generation (\pens{}),
    and class-agnostic Decoder Warm-up (DeWa).}
    \label{tab:pq_ablations}
\end{table*}

%% file: tables/04_coco-panoptic-pq.tex
\begin{table}[th]
    \centering
    \footnotesize
    \setlength{\tabcolsep}{3pt}
    \begin{tabular}{lcccccc}
        \toprule
        \multirow{2}{*}{Method} &\multirow{2}{*}{SSL}&  1/512 &    1/256 &    1/128 &    1/64 &    1/32 \\
        & &  (232) & (463) & (925) & (1849) & (3697) \\
        \midrule
        M2F-Lang                 & \n & 15.0 & 26.0& 33.2 & 37.8 & 41.1 \\
        \ \blarrow \ + \pens{} (eval only)       & \n & 22.9 & 30.7& 36.3 & 39.5 & 42.5 \\
        \midrule
         M2F-Lang                 & \y & 27.6 & 34.7& 38.9 & 42.0 & 44.2 \\
        \ \blarrow \ + \pens{} (\textbf{\ours}) &\y& 34.7 & 38.6 & 40.8 & 43.0 & 44.8 \\
        \ \ \ \ \blarrow \ + DeWa (\textbf{\ourspp{}})  &\y& \bm1{38.8} & \bm1{41.3} & \bm1{43.1} & \bm1{44.5} & \bm1{46.4} \\
         \bottomrule
    \end{tabular}
    \caption{PQ performance on \textbf{COCO-Panoptic}. 
    All experiments leverage frozen \faSnowflake~CN-B-L2B backbone.
    Top section shows the supervised model
    and improvement due to inference only ensembling.
    Bottom section starts from semi-supervised learning
    with decoder only posteriors,
    and presents improvements due to synergy of
    SSL and \pens{}
    (\ours{})
    and 
    decoder warm-up (\ourspp{}).}
    \label{tab:performance_coco_panoptic}    
\end{table}

%% file: tables/04_sota-comparison-vlm-ade20k.tex
\begin{table}[b]
    \centering
    \scriptsize
    \setlength{\tabcolsep}{3pt}
    \begin{tabular}{llc@{\hspace{3pt}}c@{\hspace{3pt}}c@{\hspace{3pt}}c@{\hspace{3pt}}c}
        \toprule
        \multirow{2}{*}{Method} &  \multirow{2}{*}{Net} & 1/128 & 1/64 & 1/32 & 1/16 & 1/8 \\
        & & (158) & (316) & (632) & (1263) & (2526) \\
        \midrule
        CutMix~\cite{french20bmvc} \venue{BMVC'20}     & R101 &  --  &  --  & 26.2  & 29.8  & 35.6 \\
        AEL~\cite{hu21neurips} \venue{NeurIPS'21}     & R101 &  --  &  --  & 28.4  & 33.2  & 38.0 \\
        UniMatch~\cite{yang23cvpr} \venue{CVPR'23}  & R101 & 15.6 & 21.6 & 28.1  & 31.5  & 34.6 \\
        UniMatch~\cite{yang23cvpr,hoyer24eccv} \venue{CVPR'23}   & ViT-B/16 & 18.4 & 25.3 & 31.2 & 34.4 & 38.0 \\
        \text{M2F+SSL (\cf\cref{tab:pq_ablations})} & \text{CN-B-IN1k} & \text{19.9} & \text{29.4} & \text{34.0} & \text{37.4} &  \bm3{41.2}\\
        \midrule
        SemiVL~\cite{hoyer24eccv} {\venue{ECCV'24}} &  \underline{ViT-B/16} & {28.1} & \bm3{33.7} & {35.1} & {37.2} & {39.4} \\
        {SemiVL}$^\dagger$~\cite{hoyer24eccv} {\venue{ECCV'24}} &  CN-B-L2B &
        \bm3{30.2} & \text{33.3} & \bm3{36.3} & \bm3{37.7} & {40.7} \\[0.6em]
        \multirow{2}{*}{\textbf{\ours}} &   \multirow{2}{*}{CN-B-L2B} &\bm2{36.5} &\bm2{40.5} &\bm2{42.8} & \bm1{45.8} &\bm2{47.5} \\
        & &  \green{(+8.4)} & \green{(+6.8)} & \green{(+7.7)} & \green{(+8.6)} & \green{(+8.1)} \\[0.5em]
        \midrule
        \multirow{2}{*}{\textbf{\ourspp}}   & \multirow{2}{*}{CN-B-L2B} &\bm1{38.9} &\bm1{42.0} &\bm1{44.3} & \bm2{45.0} &\bm1{48.1} \\
        & &  \green{(+10.8)} & \green{(+8.3)} & \green{(+9.2)} & \green{(+7.8)} & \green{(+8.7)} \\
        \bottomrule
    \end{tabular}
        \caption{
            Comparison with the state of the art in semi-supervised semantic segmentation (mIoU) on \textbf{ADE20K}.
            $\dagger$ indicates our experiments with public source code. \underline{Underline} denotes CLIP-WiT~\cite{radford21icml} initialization.
            Improvements over SemiVL-ViT-B/16 are in~\textcolor{Green}{green}.
            Gold, silver and bronze denote the best results.
        }
    \label{tab:sota_ade}
\end{table}

%% file: tables/04_coco-objects-pq.tex
\begin{table}[t]
    \centering
    \scriptsize
    \setlength{\tabcolsep}{3pt}
    \begin{tabular}{lccccc}
        \toprule
        \multirow{2}{*}{Method}  &     1/512 &    1/256 &    1/128 &    1/64 &    1/32 \\
        & (232) & (463) & (925) & (1849) & (3697) \\
        \midrule
        \faSnowflake ~\ours{} & 37.9 & 41.0 & 42.9 & 45.0 & 46.5 \\
        \midrule
        \faSnowflake ~\ourspp{}  & \bm1{42.0} & \bm1{43.7} & \bm1{45.4}&  \bm1{47.6}& \bm1{48.7} \\
        \bottomrule
    \end{tabular}
    \caption{Semi-supervised PQ performance on \textbf{COCO-Objects}.}
    \label{tab:performance_coco_objects}
    
\end{table}

%% file: tables/04_sota-comparison-vlm-coco-objects.tex
\begin{table}[t]
    \centering
    \scriptsize
    \setlength{\tabcolsep}{3pt}
    \begin{tabular}{llc@{\hspace{3pt}}c@{\hspace{3pt}}c@{\hspace{3pt}}c@{\hspace{3pt}}c}
        \toprule
        \multirow{2}{*}{Method}               &     \multirow{2}{*}{Net} &     1/512 &    1/256 &    1/128 &    1/64 &    1/32 \\
        & & (232) & (463) & (925) & (1849) & (3697) \\
        \midrule
        PseudoSeg~\cite{zou21iclr,zhong21iccv} \venue{ICLR'21} &                      XC-65 &            29.8 &           37.1 &           39.1 &           41.8 &           43.6 \\
        PC$^2$Seg~\cite{zhong21iccv} \venue{ICCV'21} &                      XC-65 &            29.9 &           37.5 &           40.1 &           43.7 &           46.1 \\
        CISC-R~\cite{wu23pami} \venue{TPAMI'23} & XC-65 & 32.1 & 40.2 & 42.2 & \n & \n \\
        UniMatch~\cite{yang23cvpr} \venue{CVPR'23} &                      XC-65 &            31.9 &           38.9 &           44.4 &           48.2 &           49.8 \\
        UniMatch~\cite{yang23cvpr,hoyer24eccv} \venue{CVPR'23} & ViT-B/16 & 36.6 & 44.1 & 49.1 & 53.5 & 55.0 \\
        LogicDiag~\cite{liang23iccv} \venue{ICCV'23} &                      XC-65 &            33.1 &           40.3 &           45.4 &           48.8 &           50.5 \\
        AllSpark~\cite{wang24cvpr} \venue{CVPR'24} & MiT-B5 & 34.1 & 41.7 & 45.5 & 49.6 & \n \\
        S4Former~\cite{hu24cvpr} \venue{CVPR'24} & DeiT-B & 35.2 & 43.1 & 46.9 & \n & \n \\
        \text{M2F+SSL} & \text{CN-B-IN1k} & \text{38.6} & \text{46.0} & \text{50.8} & \text{54.6} &  \text{55.7}\\
        \midrule
        {SemiVL}~\cite{hoyer24eccv} \venue{ECCV'24} & \underline{ViT-B/16} &            \bm3{50.1} &           \bm3{52.8} &           \bm3{53.6} &           \bm3{55.4} &           \bm3{56.5} \\
        {SemiVL}$^\dagger$~\cite{hoyer24eccv} \venue{ECCV'24} & {{CN-B-L2B}} &            \text{47.6} &           \text{49.1} &           \text{50.1} &           \text{52.6} &           \text{52.9} \\[0.6em]
        \multirow{2}{*}{\textbf{\ours}} & \multirow{2}{*}{{CN-B-L2B}} &  \bm2{52.9} &           \bm2{53.9} &           \bm2{56.2} &           \bm2{58.7} &           \bm2{59.3} \\
        &   & \green{(+2.8)} & \green{(+1.1)} & \green{(+2.6)} & \green{(+3.3)} & \green{(+2.8)} \\
        \midrule
        \multirow{2}{*}{\textbf{\ourspp}}  & \multirow{2}{*}{{CN-B-L2B}} &           \bm1{54.6} &           \bm1{55.1} &           \bm1{57.0} & \bm1{59.1} &           \bm1{60.2} \\
        &                                                                   & \green{(+4.5)} & \green{(+2.3)} & \green{(+3.4)} & \green{(+3.7)} & \green{(+3.7)} \\
        \bottomrule
    \end{tabular}
        \caption{
            Comparison with the state of the art in semi-supervised semantic segmentation (mIoU) on \textbf{COCO-Objects}.
            $\dagger$ indicates our experiments with public source code.
            \underline{Underline} denotes CLIP-WiT~\cite{radford21icml} initialization.
        }
    \label{tab:sota_coco}
\end{table}

%% file: tables/04_additional-ablation-different-initialization.tex
\begin{table}[t!]
    \centering
    \footnotesize
    \setlength{\tabcolsep}{3pt}
    \begin{tabular}{lccccc}
        \toprule
        \multirow{2}{*}{Method}  & 1/128 & 1/64 & 1/32 & 1/16 & 1/8 \\
        &  (158) & (316) & (632) & (1263) & (2526) \\
        \midrule
        Random init (\textbf{\ours})             & 27.7& 32.3 & 34.8 & 38.3& 40.6\\ 
        Labeled-only init                & 26.7 & 31.7 & 34.8 & 38.0 & 40.4 \\
        \midrule
        SA1B (1\%) - pretraining                  & \bm1{30.2} & \bm2{34.0} & \bm1{36.3} & \bm2{38.7} & \bm1{41.8}\\
        SAM pseudo-labels (\textbf{\ourspp})      & \bm2{29.9} & \bm1{34.6} & \bm1{36.3} & \bm1{39.0} & \bm2{41.6} \\
        \bottomrule
    \end{tabular}
        \caption{Validation of decoder warm-up procedures on ADE20K.}
    \label{tab:ablation_pretraining-stage}

\end{table}

%% file: tables/iccv_table_pmt_ablation.tex
\begin{table}[h!]
    \centering
    \footnotesize
    \setlength{\tabcolsep}{3pt}
    \begin{tabular}{l@{\ }rccccc}
    
        \toprule
         Method& &  1/128 &    1/64 &    1/32 &    1/16 &    1/8 \\
        \midrule
        $\mathbf{P}_{\mathrm{MT}}$   &  \gray{(teacher only)}   & 18.3 & 26.4 & 32.5 & 35.5 & 39.2\\[-0.2em]
        $\mathbf{P}_{\mathrm{CLIP}}$ &\gray{(teacher only)} & 24.1 & 28.3 & 32.5 & 36.1 & 39.2\\[-0.2em]
        \midrule
        $\mathbf{P}_{\mathrm{ENS}}$  & \gray{(Eq. \ref{eq:ensembling}, teacher only)}  & \bm1{27.3} & \bm1{31.5} & \bm1{34.4} & \bm1{38.0} & \bm1{40.3} \\[-0.2em]
        \bottomrule
    \end{tabular}
    \caption{Ablation of posterior ensembling on ADE20K (PQ).}
        \label{tab:iccv-ablations-pens}    
\end{table}

%% file: fig/fig_tex/04_figure_grouped_improvements_v2.tex
\begin{figure}[h!]
    \footnotesize
    \centering
    \begin{tabular}{c}
         \includegraphics[width=0.95\linewidth]{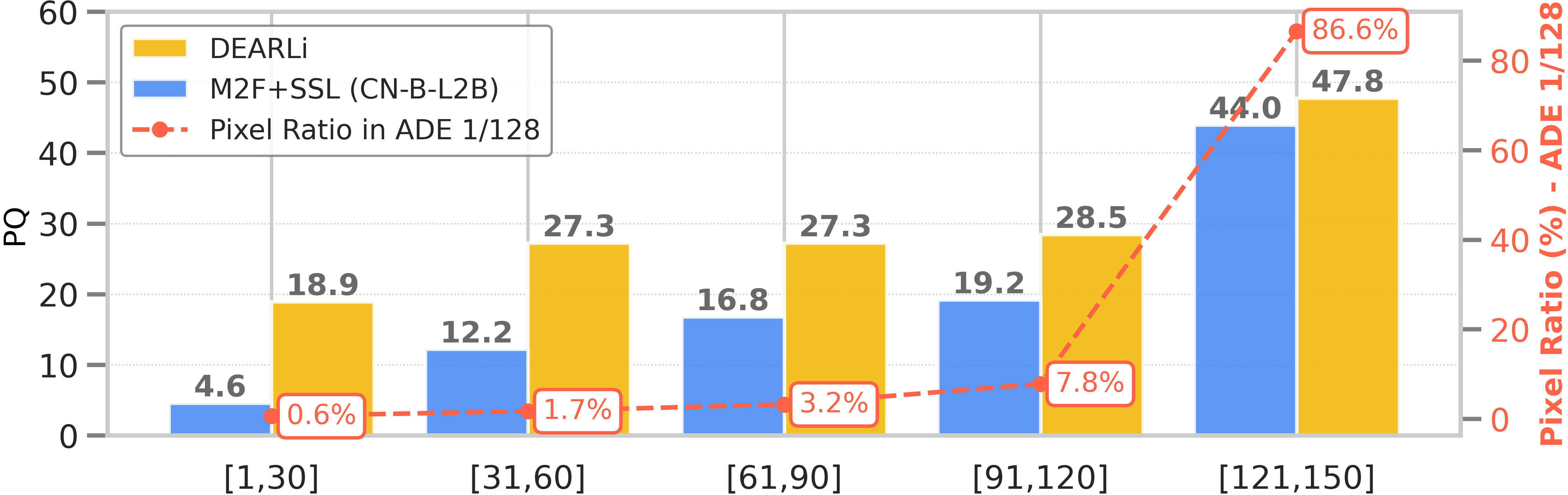}\\
         \includegraphics[width=0.95\linewidth]{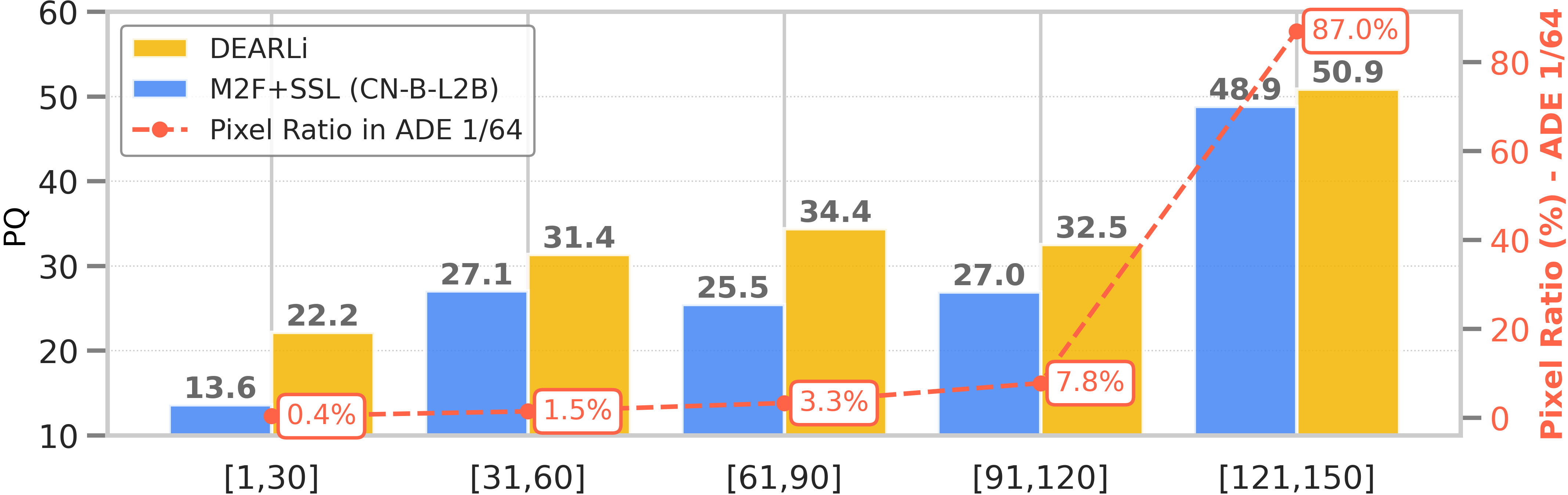} 
    \end{tabular}
    \caption{
    PQ performance on ADE 1/128 (top)
    and ADE 1/64 (bottom)
    for five groups of classes ranked 
    according to per-pixel label incidence.
    \ourspp{} improves upon M2F+SSL
    in all groups, but most significantly
    on underrepresented classes. 
    }
    \label{fig:underrepresented-classes-gains}
\end{figure}

%% file: tables/04_basevslarge_fig.tex
\begin{figure}[h]
    \centering
    \begin{tabular}{c@{\hspace{3pt}}c}
             \includegraphics[width=0.465\linewidth]{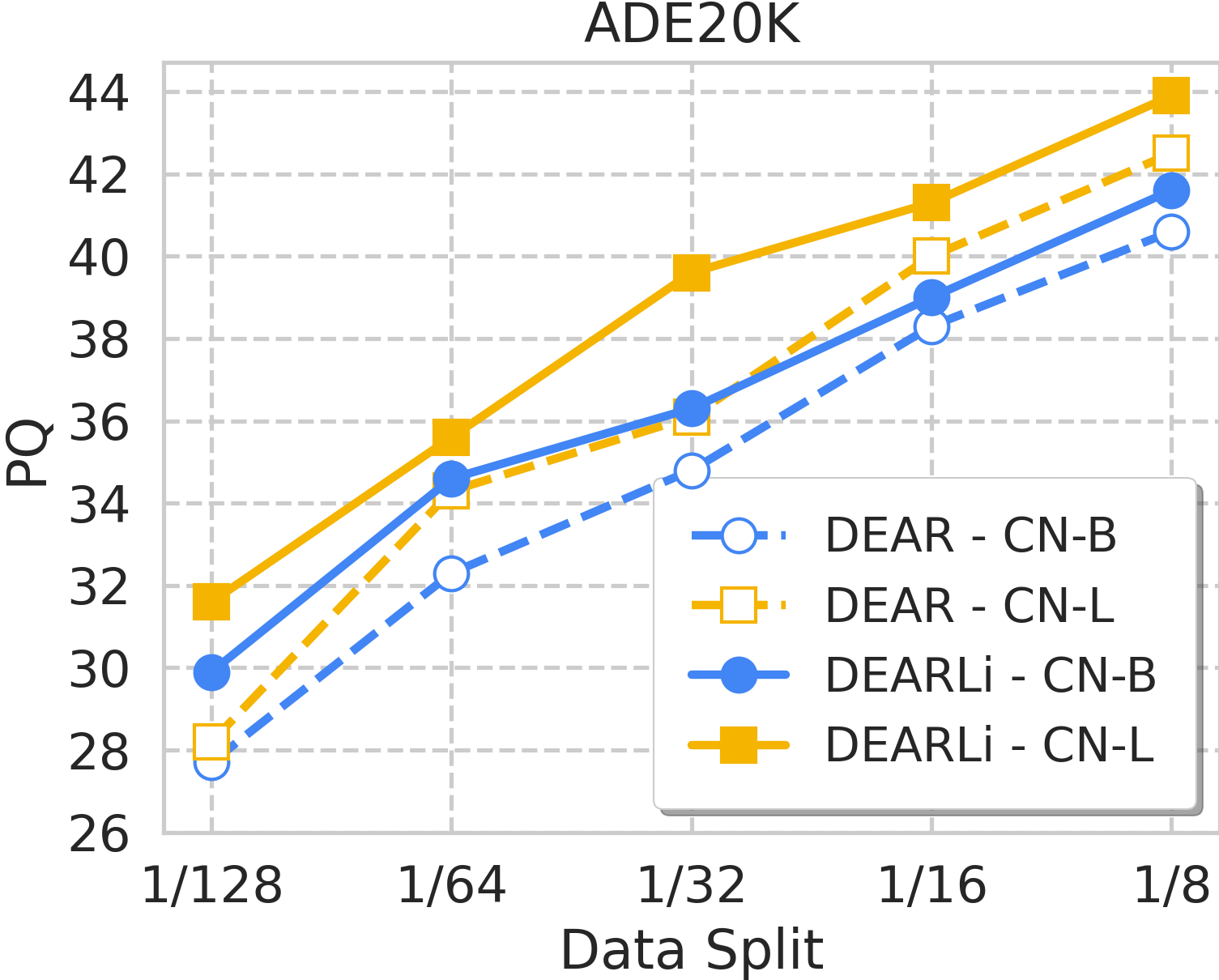}
         &   
             \includegraphics[width=0.465\linewidth]{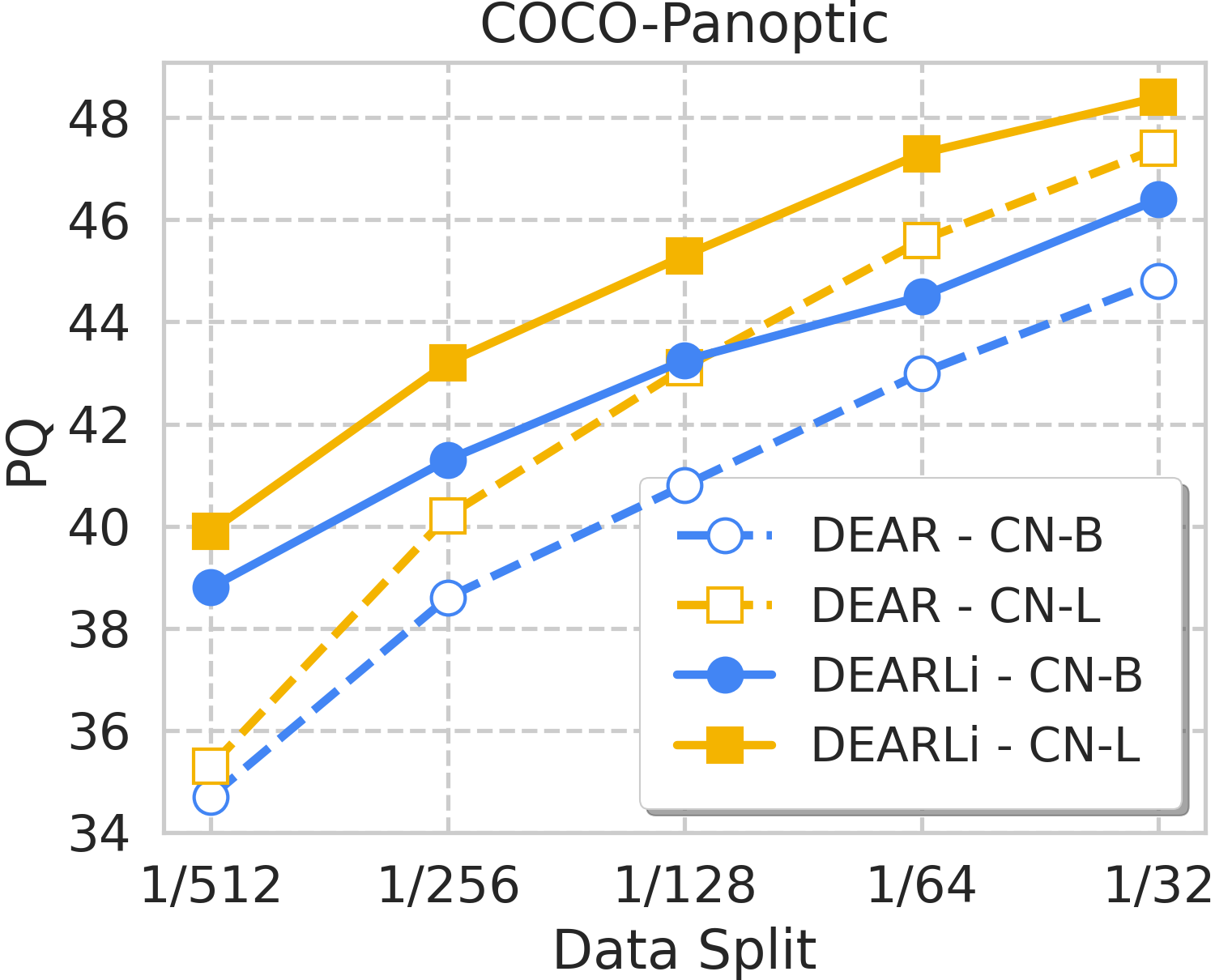}
             \\ 
    \end{tabular}
    \caption{
    PQ performance of \ours{} and \ourspp{}
    with two different backbones
    on ADE20K~(left) 
    and COCO-Panoptic~(right).}    
    \label{fig:increasing-backbone-size}
\end{figure}

%% file: tables/04_additional-ablation_performance-for-different-alphas-ade20k.tex
\begin{table}[ht!]
    \centering
    \footnotesize
    \begin{tabular}{lcccccc}
        \toprule
        & 1/128 & 1/64 & 1/32 & 1/16 & 1/8 \\
        &  (158) & (316) & (632) & (1263) & (2526) \\
        \midrule
        $\alpha=0.5$ & 26.70& 31.22 & 34.60 & 36.88 & 39.66 \\
        $\alpha=0.6$ & \bm1{27.67} & \bm1{32.27} & 34.77 & \bm1{38.27} & 40.62 \\
        $\alpha=0.7$ & 24.35 & 31.05 & \bm1{34.88} & 37.80 & \bm1{41.42} \\
        \bottomrule
    \end{tabular}
    \caption{
    PQ performance of~\ours{} on ADE20K for varying $\alpha$.
    }
    \label{tab:analysis_alpha}

\end{table}

%% file: sections/6_conclusion.tex
\section{Conclusion}
\label{sec:conclusion}
We have presented
a foundation-model-powered
method for semi-supervised
panoptic segmentation.
Our method assumes decoupled
recognition and localization heads
of a mask transformer baseline
and enhance them separately.
Recognition is enhanced
by ensembling the standard
mask-wide posteriors
with zero-shot classification
of mask-pooled CLIP features.
Localization is improved
through class-agnostic
decoder warm-up
towards SAM pseudo-labels.
Our panoptic models establish
strong semi-supervised baselines
on ADE20K and COCO,
surpassing supervised counterparts
trained with 4$\times$ more labeled data.
They also achieve state-of-the-art
performance in
semi-supervised semantic
segmentation despite optimizing
a panoptic objective,
while requiring 8$\times$ fewer GPUs.

%% file: sections/acknowledgments.tex
\section*{Acknowledgments}
This work has been supported
by Croatian Recovery
and Resilience Fund -
NextGenerationEU (grant C1.4 R5-I2.01.0001),
Croatian Science Foundation
(grant IP-2020-02-5851 ADEPT),
Advanced computing service provided
by the University of Zagreb
University Computing Centre - SRCE,
Slovenian research agency research program \mbox{P2-0214},
and European Union's Horizon Europe
research and innovation programme under the Marie Skłodowska-Curie Postdoctoral Fellowship Programme, SMASH co-funded under the grant agreement No.~101081355.
The SMASH project is co-funded by
the Republic of Slovenia and
the European Union from the
European Regional Development Fund.

%% file: sections/appendix.tex
\clearpage
\setcounter{page}{1}
\maketitlesupplementary
\appendix
\section{Additional ablations and results}
\textbf{Increasing backbone size.}
\Cref{tab:vary-ade,tab:vary-cocop} 
present the performance of \ours{} and \ourspp{} 
with ConvNeXt-B (CN-Base) and ConvNeXt-L (CN-Large) backbones
on ADE20K and COCO-Panoptic. 
These experiments correspond to 
\Cref{fig:increasing-backbone-size} 
from the main manuscript, 
but are tabulated 
for easier comparison 
and reference for future work.
Both \ours{} and \ourspp{} benefit
from increased backbone capacity.
We observe this benefit 
across all data partitions
on both datasets.
Note that a frozen CN-Large
backbone still enables 
\ours{} and \ourspp{} 
to be trained on a single
A100-40GB GPU.
\input{tables/04_additional-ablation_vary_model_size}
\input{tables/04_additional_ablation_vary_model_size_coco}

\textbf{On training stability.}
\Cref{tab:vary-seeds}
presents PQ of
our method
on ADE20K 
across three 
different seeds.
\input{tables/appendix_stability_ade}
The first row corresponds
to the results
of the run presented 
in~\cref{tab:pq_ablations}
of the main manuscript.
The next two rows present
additional runs
with the comparable performance.
Overall, 
the experiments exhibit
small variance, 
which demonstrates 
the training stability 
of our method.

\textbf{Panoptic vs. semantic labels.}
\Cref{tab:semseg-training}
examines the impact of
training labels 
on the semi-supervised 
semantic segmentation 
performance
of our method.
The first row
repeats performance of~\ours{}
from 
the~\cref{tab:sota_ade}
of the main manuscript.
The second row presents
experiments where
the same model
is retrained with semantic segmentation labels,
which requires additional
adaptation to the pseudo-labels
generation procedure for
the semantic segmentation objective.
We observe that the models
achieve comparable performance,
which justifies our comparison
with previous methods in~\cref{tab:sota_ade} from the main manuscript.
This demonstrates that 
the influence of
the training labels is minimal.
Indeed, the semantic segmentation labels
significantly outperform the panoptic labels
in the 1/32 experiment,
while underperforming at 1/128
and performing roughly the same 
in the remaining three experiments.
\input{tables/04_appendix_semseg-training}

\textbf{Semantic segmentaiton ablations.} 
\input{tables/iccv_table_miou_ablation}
\Cref{tab:iccv-miou-ablation}
presents ablations of \ourspp{}
for semantic segmentation.
These models are identical to those
in Tab.~1 of the main manuscript,
but evaluated with standard
M2F semantic segmentation inference.
The trends observed are similar to those
for panoptic segmentation. DEAR significantly
outperforms the M2F-Lang baseline across
all data partitions, while DEARLi
further enhances performance
in 4 out of 5 scenarios. 
Furthermore, we observe (\cf rows 1 and 2)
that our M2F-Lang baseline in 3 out of 5 scenarios
outperforms state-of-the-art 
semantic segmentation
method SemiVL~\cite{hoyer24eccv},
which proves that
our panoptic baseline is indeed strong. 

\input{tables/04_appendix_ablations}
\textbf{Additional panoptic ablations.}
\Cref{tab:pq_ablations_appendix} extends 
\Cref{tab:pq_ablations} from the main manuscript.
The second section
shows experiments with
ImageNet-21k~\cite{ridnik21neurips}
backbone initialization. 
We observe improvements
over the ImageNet-1k initialization
(\cf\ first section)
in all setups
across all data regimes. 
Nevertheless,
pre-training
on LAION-2B~\cite{schuhmann22neurips} still 
prevails (\cf\ third section).
This confirms the benefits
of contrastive
language-image
pretraining 
with respect to the traditional
pre-training on categorical
image-wide labels.
The last section
presents additional
ablations of our 
contributions.
The second row evaluates \ours{}
without ensembling with zero-shot
CLIP features during inference,
using ensembling only in the
teacher to enhance pseudo-labels.
The experiments reveal
that inference-time 
ensembling brings
slight improvement
of 0.3-0.8 PQ points,
depending on the
data partition.
The last row 
shows the contribution
of the decoder warm-up 
(DeWa) to the Mean Teacher 
baseline.
We observe that DeWa
consistently improves
the performance by
1.4-2.7 PQ points.
However, 
it still performs significantly worse 
than \ourspp{}. 
These results highlight the complementary nature 
of the proposed contributions.

\section{SemiVL with a ConvNeXt Backbone}
\label{sec:suppl-semivl}
\input{tables/04_appendix_semivl_reproduction}
\input{tables/04_appendix_semivl_reproduction_coco-objects}
We first successfully reproduced
SemiVL~\cite{hoyer24eccv} experiments
with ViT.
Then, we reconfigure SemiVL
with a ConvNeXt-B backbone by introducing
several straightforward modifications
to the published source code.
First, new language embeddings
are generated using the
appropriate text encoder from OpenCLIP~\cite{cherti23cvpr}.
Second, the upsampling component
of SemiVL is adjusted to
align with the feature dimensions
of ConvNeXt at each stage.
An extra skip connection
and upsampling block
are added to
accommodate the hierarchical
structure of
convolutional backbones.
All other implementation 
details remain consistent
with the ViT-B/16 setup.

SemiVL~\cite{hoyer24eccv} only fine-tunes 
the attention weights of the ViT backbone.
Since ConvNeXt lacks
a directly analogous mechanism,
we freeze the backbone to ensure
a direct comparison with our approach.
We also conduct experiments
with a fully fine-tuned ConvNeXt backbone,
using the same hyperparameters
as described in the original paper.
In~\Cref{tab:sota_ade,tab:sota_coco} (main manuscript), we report experiments with a frozen backbone, while~\Cref{tab:semivl-reproduction-ade,tab:reproduction-coco-objects-semivl} include additional experiment with a fine-tuned backbone.
These results suggest that
SemiVL~\cite{hoyer24eccv}
gains no benefit 
from fine-tuning 
the ConvNeXt backbone.
Note that SemiVL~\cite{hoyer24eccv} 
reports results
from the
best checkpoint on
the validation set.
In our SemiVL reproduction,
we also report results from the best epoch.
As outlined in the main manuscript,
all results for \ours{} and \ourspp{}
are obtained by evaluating the checkpoint
from the final training iteration.

\section{Limitations}
The hyperparameter $\alpha$ for geometric ensembling is manually set to 0.6, which may be suboptimal for different data partitions. From \cref{tab:analysis_alpha} in the main manuscript, we observe that higher values of $\alpha$ increase performance when more labeled data is available. This suggests that making $\alpha$ adaptive based on the quantity of labeled data could be beneficial. Additionally, it may be advantageous to dynamically adjust $\alpha$ during training (e.g., starting with a lower value early on and increasing it in later stages). These challenges present opportunities for future research.
\section{Further Implementation Details}
This section presents implementation details
of image perturbations that are
presented in~\Cref{sec:experiments}
of the main manuscript.
For color jittering,
we use \texttt{torchvision}
implementation with the
following parameters:
\texttt{brightness:(0.2, 1.8)},
\texttt{saturation:(0.2, 1.8)},
\texttt{contrast:(0.2, 1.8)},
and \texttt{hue:(-0.2, 0.2)}.
Gaussian blur (\texttt{sigma:(0.1, 2.0)})
and CutMix~\cite{yun19iccv,sohn20neurips}
are applied with a probability of 0.5,
and grayscaling with a probability 0.2.

The number of mask queries
in Mask2Former~\cite{cheng22cvpr} is fixed at
200 across all experiments.
Models trained without unlabeled data (supervised)
begin with 10k iterations
for the smallest data regime
(\ie, 1/128 for ADE20K and 1/512 for COCO),
with an additional 10k
iterations added for
each subsequent regime.
The batch size in these experiments is 8.
For the decoder warm-up stage,
we generate class-agnostic
pseudo-labels using ViT-Huge SAM~\cite{kirillov23iccv}.
\section{Panoptic Segmentation Examples}
\Cref{tab:ade-visualizations-panoptic,tab:coco-panoptic-visualizations-panoptic}
show panoptic predictions
of our models
on ADE20K and COCO-Panoptic
validation images,
respectively.
The models are trained
on the most challenging
data partitions
with the least amount
of labeled images.
The first two columns 
display the input image 
and the corresponding 
ground truth. 
The next three columns
present overlaid panoptic predictions
of the Mean Teacher baseline (M2F+SSL), 
\ours{} and \ourspp{}. 
Comparing \ours{} with the 
baseline reveals that 
the recognition enhancement 
often rectifies classification errors 
(e.g., \textit{building} to \textit{wall} in the 2nd row of \cref{tab:ade-visualizations-panoptic}, or
\textit{suitcase} to \textit{handbag} in the 6th row of \cref{tab:coco-panoptic-visualizations-panoptic}). 
Additionally, a comparison between \ours{} and \ourspp{} 
highlights how the localization enhancement
refines segmentation boundaries 
(e.g., \textit{chair} legs in the third row of \cref{tab:ade-visualizations-panoptic}).
The last row
of \cref{tab:ade-visualizations-panoptic}
shows an interesting 
failure mode
where our models
isolate parts
of the \textit{house} such as
\textit{door} and \textit{windows}
into separate segments.
While this decision
is not entirely wrong,
it deviates from the 
dataset’s labeling policy
which assigns these classes
to indoor scenes only.
This illustrates
the limitations
of CLIP zero-shot
classification,
as class text embeddings
diverge from the 
labeling policy.
\input{tables/04_appendix_panoptic_predictions}
\input{tables/04_appendix_panoptic_predictions_coco-panoptic}

\section{Comparison with the State of the Art}
\Cref{tab:ade-visualizations-comparison-sota,tab:coco-objects-visualizations-comparison-sota} 
compare semantic segmentation predictions
of our method \ours{} 
with 
the state-of-the-art method SemiVL~\cite{hoyer24eccv}
on ADE20K and COCO-Objects, respectively.
We consider models
trained in 
most challenging 
data partitions
with the least amount
of labeled data.
The columns show 
the input image,
ground truth,
and predictions
of SemiVL~\cite{hoyer24eccv} and \ours{}.
We observe that SemiVL 
occasionally misclassifies 
correctly segmented objects 
(e.g.\ the third and fifth row 
in \cref{tab:ade-visualizations-comparison-sota}). 
Furthermore, 
it sometimes splits 
a single object into 
two different classes 
(e.g., the fourth row in 
\cref{tab:ade-visualizations-comparison-sota} 
and the fifth row in 
\cref{tab:coco-objects-visualizations-comparison-sota}). 
In contrast, 
our model produces 
fewer misclassifications and 
more cohesive segmentations. 
We argue this is due to the 
mask transformer framework, 
which enables zero-shot CLIP 
classification at the mask level and
delivers more consistent predictions than 
in the SemiVL's patch-level
classification approach.
\input{tables/04_appendix_sota_comparison_ade_visualizations}
\input{tables/04_apendix_sota_comparison_coco-objects-visualization}

%% file: tables/04_additional-ablation_vary_model_size.tex
\begin{table}[ht!]
    \centering
    \footnotesize
    \setlength{\tabcolsep}{4.5pt}
    \renewcommand{\arraystretch}{1.1} 
    \begin{tabular}{ll@{\hspace{10pt}}ccccc}
        \toprule
        \multirow{2}{*}{Method} & \multirow{2}{*}{\makecell{Backbone\\Size}} & 1/128 & 1/64 & 1/32 & 1/16 & 1/8 \\
                                                                     & & (158) & (316) & (632) & (1263) & (2526) \\
        \midrule
        \ours{}   & \faSnowflake{} CN-B-L2B  & 27.7 & 32.3 &  34.8 &  38.3 & 40.6   \\
        \ours{}   &\faSnowflake{} CN-L-L2B & 28.2 & 34.3 & 36.1 & 40.0 &  42.5\\
        \midrule
        \ourspp{} &\faSnowflake{} CN-B-L2B  &  29.9  & 34.6  & 36.3  & 39.0  & 41.6  \\
        \ourspp{} &\faSnowflake{} CN-L-L2B & 31.6 & 35.6 & 39.6 & 41.3 &  43.9\\
        \bottomrule
    \end{tabular}
    \caption{Panoptic performance (PQ) on ADE20K when increasing the backbone size. CN-B and CN-L refer to ConvNeXt-Base and ConvNeXt-Large, respectively. L2B denotes the LAION-2B pre-training dataset. \faSnowflake{} represents a frozen backbone.}
    \label{tab:vary-ade}

\end{table}

%% file: tables/04_additional_ablation_vary_model_size_coco.tex
\begin{table}[ht!]
    \centering
    \footnotesize
    \setlength{\tabcolsep}{4.5pt}
    \renewcommand{\arraystretch}{1.1} 
    \begin{tabular}{ll@{\hspace{10pt}}ccccc}
        \toprule
        \multirow{2}{*}{Method} & \multirow{2}{*}{\makecell{Backbone\\Size}} & 1/512 & 1/256 & 1/128 & 1/64 & 1/32 \\
                                                                     & & (232) & (463) & (925) & (1849) & (3697) \\
        \midrule
        \ours{}   & \faSnowflake{} CN-B-L2B  & 34.7 & 38.6 &  40.8 &  43.0 & 44.8   \\
        \ours{}   & \faSnowflake{} CN-L-L2B & 35.3 & 40.2 & 43.1 & 45.6 &  47.4\\
        \midrule
        \ourspp{} & \faSnowflake{} CN-B-L2B  &  38.8  & 41.3  & 43.1  & 44.5  & 46.4  \\
        \ourspp{} & \faSnowflake{} CN-L-L2B & 39.9 & 43.2 & 45.3 & 47.3 &  48.4\\
        \bottomrule
    \end{tabular}
    \caption{Panoptic performance (PQ) on COCO-Panoptic when increasing the backbone size.}
    \label{tab:vary-cocop}

\end{table}

%% file: tables/appendix_stability_ade.tex
\begin{table}[h!]
    \centering
    \footnotesize
    \setlength{\tabcolsep}{3.5pt}
    \renewcommand{\arraystretch}{1.1} 
    \begin{tabular}{lccccc}
        \toprule
        \multirow{1}{*}{Method}& 1/128 & 1/64 & 1/32 & 1/16 & 1/8 \\
        \midrule
         \multirow{3}{*}{\ourspp{}} 
        &\textbf{29.9}&\textbf{34.6}&\textbf{36.3}&\textbf{39.2}&\textbf{41.6}  \\
         &30.5&34.8&36.5&39.1&42.0  \\
         &30.7&34.8&36.1&39.5&41.6  \\
        \midrule
         &$30.4_{\pm0.3}$&$34.7_{\pm0.1}$&$36.3_{\pm0.2}$&$39.3_{\pm0.2}$&$41.7_{\pm0.2}$  \\
        \bottomrule
    \end{tabular}
    \caption{Panoptic performance (PQ) of our final method, \ourspp{}, across three different seeds for standard data partitions of ADE20K. Bolded results correspond to the runs reported in \cref{tab:pq_ablations} of the main manuscript. The last row represents $\text{mean}_{\pm \text{std}}$.}
    \label{tab:vary-seeds}
\end{table}

%% file: tables/04_appendix_semseg-training.tex
\begin{table}[h!]
    \centering
    \footnotesize
    \setlength{\tabcolsep}{2pt}
    \renewcommand{\arraystretch}{1.1} 
    \begin{tabular}{ll@{\hspace{10pt}}ccccc}
        \toprule
        \multirow{2}{*}{Method} & \multirow{2}{*}{\makecell{Backbone}} & 1/128 & 1/64 & 1/32 & 1/16 & 1/8 \\
                                                                     & & (158) & (316) & (632) & (1263) & (2526) \\
        \midrule
        \ours{}   & \multirow{2}{*}{\faSnowflake{} CN-B-L2B}  & {36.5} &{40.5} &{42.8} & {45.8} &{47.5}    \\
        \ours{}-\textit{semseg}   & & 35.8 & 40.9 & 44.8& 45.2 & 47.4 \\
        \bottomrule
    \end{tabular}
    \caption{Semantic performance (mIoU)
    on different ADE20K data partitions.
    \ours{} is trained
    with respect to the
    standard panoptic labels
    as in the main manuscript.
    In contrast,
    \ours{}-\textit{semseg} is trained
    with respect 
    to the semantic
    segmentation labels.
    Notably, the reported performances
    are closely comparable.}
    \label{tab:semseg-training}

\end{table}

%% file: tables/iccv_table_miou_ablation.tex
\begin{table}[b]
    \centering
    \scriptsize
    \setlength{\tabcolsep}{3pt}
    \begin{tabular}{lcccccc}
        \toprule
        {Method} &{SSL}&  1/128 &    1/64 &    1/32 &    1/16 &    1/8 \\
        \midrule
        SemiVL (\cf Tab.~3) \quad \textcolor{gray}{\faSnowflake~CN-B-L2B}    & \y  &
        \bm3{30.2} & \bm3{33.3} & \text{36.3} & \text{37.7} & {40.7} \\
        \midrule
        M2F-Lang \quad \qquad \quad \ \ \ \textcolor{gray}{\faSnowflake~CN-B-L2B}      & \y & 20.9 & 30.1 & \bm3{37.4} & \bm3{40.7} & \bm3{45.4} \\[-0.2em]
        \ \blarrow \ + \pens{} (\textbf{\ours})                         & \y & \bm2{36.5} & \bm2{40.5} & \bm2{42.8} & \bm1{45.8} & \bm2{47.5} \\[-0.2em]
        \ \ \ \ \blarrow \ + DeWa (\textbf{\ourspp})                    & \y & \bm1{38.9} & \bm1{42.0} & \bm1{44.3} & \bm2{45.0} & \bm1{48.1} \\
         \bottomrule
    \end{tabular}
    \caption{Semantic segmentation ablations (mIoU) on ADE20K.}
    \label{tab:iccv-miou-ablation}    
\end{table}

%% file: tables/04_appendix_ablations.tex
\begin{table*}[t!]
    \scriptsize
    \centering
    \begin{tabular}{lccc@{\hspace{3pt}}c@{\hspace{3pt}}cc@{\hspace{3pt}}c@{\hspace{3pt}}cc@{\hspace{3pt}}c@{\hspace{3pt}}cc@{\hspace{3pt}}c@{\hspace{3pt}}cc@{\hspace{3pt}}c@{\hspace{3pt}}c}
        \toprule
        \multirow{2}{*}{Method} & \multirow{2}{*}{SSL} & \multirow{2}{*}{Backbone} & \multicolumn{3}{c}{1/128 (158)} & \multicolumn{3}{c}{1/64 (316)} & \multicolumn{3}{c}{1/32 (632)} & \multicolumn{3}{c}{1/16 (1263)} & \multicolumn{3}{c}{1/8 (2526)} \\
        & & & \textbf{mPQ}  & mSQ & mRQ  & \textbf{mPQ}  & mSQ & mRQ & \textbf{mPQ}  & mSQ & mRQ  &  \textbf{mPQ}  & mSQ & mRQ  & \textbf{mPQ}  & mSQ & mRQ \\
        \midrule
        \multirow{3}{*}{M2F}        &\n& \faSnowflake~CN-B-IN1k                 & 7.4 & 32.7 & 9.3 & 13.3  & 52.2 & 16.7 & 17.8 & 62.7& 21.7& 22.1 & 65.1 & 27.0 & 25.5 & 69.3 &30.7 \\
                                    &\n& \faFire     ~CN-B-IN1k                 & 9.3 & 38.6 & 11.7 & 15.0 & 55.2 & 18.6 & 20.5 & 62.9& 25.2& 25.0 & 68.9 & 30.5 & 29.9 & 73.2 &35.9 \\
        & \y & \faFire     ~CN-B-IN1k                 & 16.4 & 45.7 & 20.4 & 22.8 & 60.4 & 27.9 & 27.8 & 70.2 & 34.0 & 30.1 & 71.4 & 36.6 & 33.6 & 75.1 & 40.3 \\
        \midrule
        \multirow{3}{*}{\textcolor{gray}{M2F}}                                       &\textcolor{gray}{\n}&               \textcolor{gray}{\faSnowflake{} CN-B-IN21k} & \textcolor{gray}{9.7}  & \textcolor{gray}{37.7} & \textcolor{gray}{12.7}  & \textcolor{gray}{14.4} & \textcolor{gray}{53.2} & \textcolor{gray}{17.8} & \textcolor{gray}{20.8} & \textcolor{gray}{64.2} & \textcolor{gray}{25.5} & \textcolor{gray}{25.8} & \textcolor{gray}{67.6} & \textcolor{gray}{31.6} & \textcolor{gray}{30.1} & \textcolor{gray}{72.2} & \textcolor{gray}{36.1} \\
                                               &\textcolor{gray}{\n}&               \textcolor{gray}{\faFire{} CN-B-IN21k} & \textcolor{gray}{11.2} & \textcolor{gray}{41.9} & \textcolor{gray}{14.1}  & \textcolor{gray}{16.7} & \textcolor{gray}{56.6} & \textcolor{gray}{20.6} & \textcolor{gray}{23.3} & \textcolor{gray}{66.4} & \textcolor{gray}{28.4} & \textcolor{gray}{27.9} & \textcolor{gray}{69.8} & \textcolor{gray}{33.7} & \textcolor{gray}{32.9} & \textcolor{gray}{74.4} & \textcolor{gray}{39.3} \\
                                         &\textcolor{gray}{\y}&               \textcolor{gray}{\faFire{} CN-B-IN21k} & \textcolor{gray}{18.7} & \textcolor{gray}{46.5} & \textcolor{gray}{23.1} & \textcolor{gray}{26.3} & \textcolor{gray}{62.1} & \textcolor{gray}{32.0} & \textcolor{gray}{30.9} & \textcolor{gray}{70.0} & \textcolor{gray}{37.1} & \textcolor{gray}{34.3} & \textcolor{gray}{74.0} & \textcolor{gray}{41.2} & \textcolor{gray}{37.6} & \textcolor{gray}{77.1} & \textcolor{gray}{45.1} \\
        \midrule
        \multirow{3}{*}{M2F}                                      &\n& \faSnowflake~CN-B-L2B                  & 10.0 & 38.7 & 12.4 & 16.4 & 56.2 & 20.1 & 23.4 & 65.4 & 28.5 & 28.7 & 69.5 & 34.9 & 34.2 & 73.7 & 40.9 \\
                                              &\n& \faFire     ~CN-B-L2B                  & 12.9 & 42.8 & 16.1 & 19.3 & 59.1 & 23.9 & 26.0 & 69.1 & 31.7 & 31.1 & 72.0 & 37.7 & 36.3 & 75.6 & 43.7 \\
                                    & \y &  \faFire    ~CN-B-L2B                  & 19.4 & 46.2 & 24.0 & 28.4 & 63.7 & 34.6 & 33.2 & 71.3 & 40.2 & 36.8 & 75.9 & 44.3 & 40.2 & 78.8 & 48.2 \\
        \midrule
        M2F-Lang                                 &\n& \multirow{6}{*}{\faSnowflake~CN-B-L2B} & 11.7 & 43.7 & 14.7 & 19.1 & 61.4& 23.9 & 26.3 & 70.2&32.1 & 31.6 &72.8 & 38.3& 36.6 & 76.7 & 44.0 \\
         \blarrow \ + \pens{} (eval only)                       &\n&                                        & 18.3 & 63.7 & 23.3 & 23.9 & 69.5 & 30.0 & 29.0 & 73.3 & 35.4 & 33.6 & 75.6 &40.8 & 37.8 &77.6 &45.7 \\
        M2F-Lang                         &\y&                                        & 18.3 & 55.0 & 22.6 & 26.4 & 68.0& 32.3& 32.5 & 76.1 & 39.4 & 35.5& 74.5 &42.8 &39.2 &77.2 &47.2 \\
        \blarrow \ + \pens{} (eval only)           &\y&                                        & {21.6} & 62.6 & 26.7 & {30.2} & 73.9 & 37.0 & {33.9} & 76.2 & 41.1 & 36.7 & 77.6 & 44.2& 40.2 & 77.2 & 48.4 \\
        \quad \blarrow \  + \pens{} (\textbf{\ours})&\y&                                        & \bm2{27.7} & 70.3 & 34.4 & \bm2{32.3} & 74.0 & 39.6 & \bm2{34.8} & 75.4 & 42.1 & \bm2{38.3} & 79.3 & 46.2 & \bm2{40.6} & 80.7 & 48.7 \\
        \qquad \blarrow \ + DeWa (\textbf{\ourspp}) &\y&                                        & \bm1{29.9} & 74.0 & 36.6 & \bm1{34.6} & 75.4 & 41.2 & \bm1{36.3} & 77.1 & 43.5 & \bm1{39.2} & 80.3 & 47.1 & \bm1{41.6} & 80.6 & 49.5 \\
        \midrule
                M2F-Lang                         &\y&    \multirow{4}{*}{\faSnowflake{} CN-B-L2B}                                   & 18.3 & 55.0 & 22.6 & 26.4 & 68.0& 32.3& 32.5 & 76.1 & 39.4 & 35.5& 74.5 &42.8 &39.2 &77.2 &47.2 \\
        \textcolor{gray}{\blarrow \ + \pens{} (teacher only)}           &\textcolor{gray}{\y}&                                        & \textcolor{gray}{27.3} & \textcolor{gray}{68.7} & \textcolor{gray}{34.0} & \textcolor{gray}{31.5} & \textcolor{gray}{72.9} & \textcolor{gray}{38.6} & \textcolor{gray}{34.4} & \textcolor{gray}{74.5} & \textcolor{gray}{41.5} & \textcolor{gray}{38.0} & \textcolor{gray}{78.2} & \textcolor{gray}{45.8}& \textcolor{gray}{40.3} & \textcolor{gray}{79.2} & \textcolor{gray}{48.4} \\
                M2F-Lang                         &\y&                                        & 18.3 & 55.0 & 22.6 & 26.4 & 68.0& 32.3& 32.5 & 76.1 & 39.4 & 35.5& 74.5 &42.8 &39.2 &77.2 &47.2 \\
                \textcolor{gray}{\blarrow \ + DeWa}           &\textcolor{gray}{\y}&&
                \textcolor{gray}{20.5} & \textcolor{gray}{55.6} & \textcolor{gray}{25.1} &
                \textcolor{gray}{29.1} & \textcolor{gray}{71.1} & \textcolor{gray}{35.0} &
                \textcolor{gray}{33.9} & \textcolor{gray}{75.6} & \textcolor{gray}{40.6} &
                \textcolor{gray}{36.9} & \textcolor{gray}{75.3} & \textcolor{gray}{44.3} &
                \textcolor{gray}{40.8} & \textcolor{gray}{78.3} & \textcolor{gray}{48.6} \\
        \bottomrule
    \end{tabular}
    \caption{
Extension of~\cref{tab:pq_ablations} from the main manuscript. The new experiments are shown in \textcolor{gray}{gray}. Pre-training on ImageNet-21k performs in between pre-training on ImageNet-1k and LAION-2B~\cite{schuhmann22neurips}.  Ensembling with zero-shot CLIP helps more when applied during training within the teacher (pseudo-labels generation). Decoder warm-up (DeWa) contributes with and without zero-shot CLIP ensembling.}
    \label{tab:pq_ablations_appendix}
\end{table*}

%% file: tables/04_appendix_semivl_reproduction.tex
\begin{table}[b]
    \centering
    \scriptsize
    \setlength{\tabcolsep}{4pt}
    \begin{tabular}{llc@{\hspace{3pt}}c@{\hspace{3pt}}c@{\hspace{3pt}}c@{\hspace{3pt}}c}
        \toprule
        \multirow{2}{*}{Method} &  \multirow{2}{*}{Net} & 1/128 & 1/64 & 1/32 & 1/16 & 1/8 \\
        & & (158) & (316) & (632) & (1263) & (2526) \\
        \midrule
        SemiVL~\cite{hoyer24eccv} {\venue{ECCV'24}} &  \textcolor{gray}{\faFire{}} \underline{ViT-B/16} & {28.1} & \text{33.7} & {35.1} & {37.2} & {39.4} \\
        {SemiVL}$^\dagger$~\cite{hoyer24eccv} {\venue{ECCV'24}} &  \faSnowflake{} CN-B-L2B &
        \bm3{30.2} & \text{33.3} & \bm3{36.3} & \text{37.7} & \bm3{40.7} \\
        {SemiVL}$^\dagger$~\cite{hoyer24eccv} {\venue{ECCV'24}} &  \faFire{} CN-B-L2B &
        \text{29.5} & \bm3{33.8} & \text{36.1} & \bm3{38.3} & {40.0} \\[0.6em]
                
        {\textbf{\ours}} &   {\faSnowflake{} CN-B-L2B} &\bm2{36.5} &\bm2{40.5} &\bm2{42.8} & \bm1{45.8} &\bm2{47.5} \\
        \midrule
        {\textbf{\ourspp}}   & {\faSnowflake{} CN-B-L2B} &\bm1{38.9} &\bm1{42.0} &\bm1{44.3} & \bm2{45.0} &\bm1{48.1} \\
        \bottomrule
    \end{tabular}
        \caption{
            Comparison with the state of the art in semi-supervised semantic segmentation on \textbf{ADE20K}.
            $\dagger$ indicates our experiments with public source code. \underline{Underline} denotes CLIP-WiT~\cite{radford21icml} initialization. \textcolor{gray}{\faFire{}} denotes partial fine-tuning (attention weights only).
        }
    \label{tab:semivl-reproduction-ade}
\end{table}

%% file: tables/04_appendix_semivl_reproduction_coco-objects.tex
\begin{table}[tbh]
    \centering
    \scriptsize
    \setlength{\tabcolsep}{4.5pt}
    \begin{tabular}{llc@{\hspace{3pt}}c@{\hspace{3pt}}c@{\hspace{3pt}}c@{\hspace{3pt}}c}
        \toprule
        \multirow{2}{*}{Method}               &     \multirow{2}{*}{Net} &     1/512 &    1/256 &    1/128 &    1/64 &    1/32 \\
        & & (232) & (463) & (925) & (1849) & (3697) \\
        \midrule
        {SemiVL}~\cite{hoyer24eccv} \venue{ECCV'24} & \textcolor{gray}{\faFire{}} \underline{ViT-B/16} &            \bm3{50.1} &           \bm3{52.8} &           \bm3{53.6} &           \bm3{55.4} &           \bm3{56.5} \\
        {SemiVL}$^\dagger$~\cite{hoyer24eccv} \venue{ECCV'24} & {\faSnowflake{} {CN-B-L2B}} &            \text{47.6} &           \text{49.1} &           \text{50.1} &           \text{52.6} &           \text{52.9} \\
                {SemiVL}$^\dagger$~\cite{hoyer24eccv} \venue{ECCV'24} & {\faFire{} {CN-B-L2B}} &            \text{47.2} &           \text{47.3} &           \text{50.0} &           \text{51.4} &           \text{52.6} \\[0.6em]
        {\textbf{\ours}} & {\faSnowflake{} {CN-B-L2B}} &  \bm2{52.9} &           \bm2{53.9} &           \bm2{56.2} &           \bm2{58.7} &           \bm2{59.3} \\
        \midrule
        {\textbf{\ourspp}}  & {\faSnowflake{} {CN-B-L2B}} &           \bm1{54.6} &           \bm1{55.1} &           \bm1{57.0} & \bm1{59.1} &           \bm1{60.2} \\
       \bottomrule
    \end{tabular}
        \caption{
            Comparison with the state of the art in semi-supervised semantic segmentation on \textbf{COCO-Objects}.
            $\dagger$ indicates our experiments with public source code.
            \underline{Underline} denotes CLIP-WiT~\cite{radford21icml} initialization.
            \textcolor{gray}{\faFire{}} denotes partial fine-tuning.
        }
    \label{tab:reproduction-coco-objects-semivl}
\end{table}

%% file: tables/04_appendix_panoptic_predictions.tex
\begin{figure*}[]
    \setlength{\tabcolsep}{1pt}
    \centering
    
    \begin{tabular}{ccccc}
         Image & Ground truth & M2F+SSL Baseline & \textbf{\ours{}} & \textbf{\ourspp{}} \\
         \includegraphics[width=0.195\linewidth]{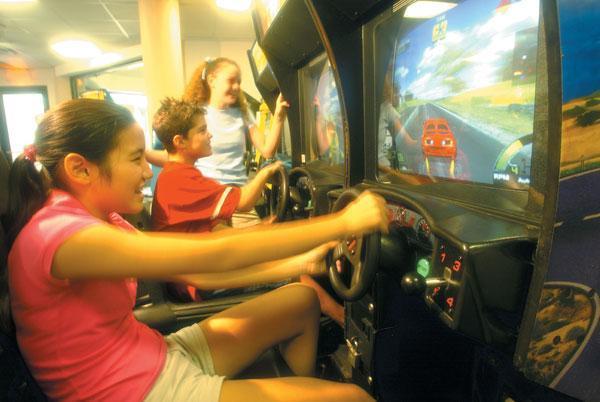} &
         \includegraphics[width=0.195\linewidth]{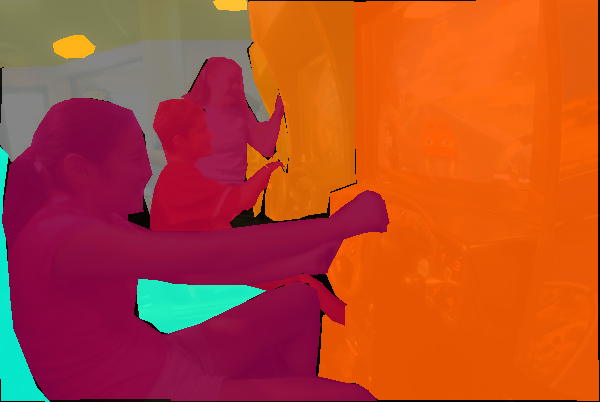} &
         \includegraphics[width=0.195\linewidth]{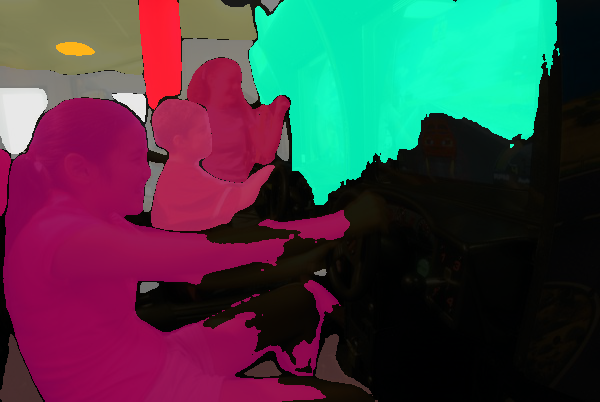} &
         \includegraphics[width=0.195\linewidth]{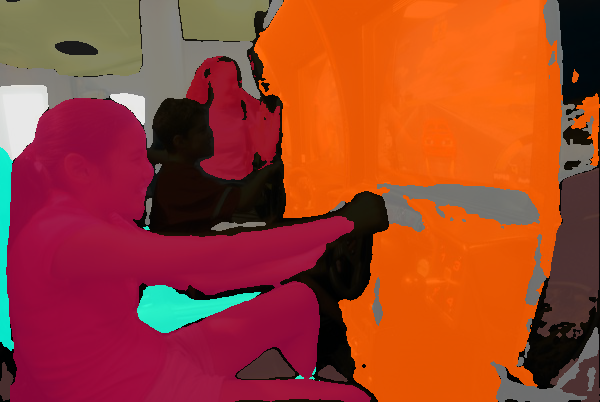} &
         \includegraphics[width=0.195\linewidth]{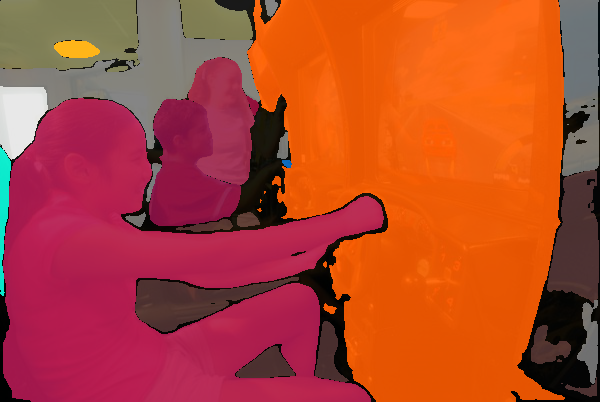} \\
         \includegraphics[width=0.195\linewidth]{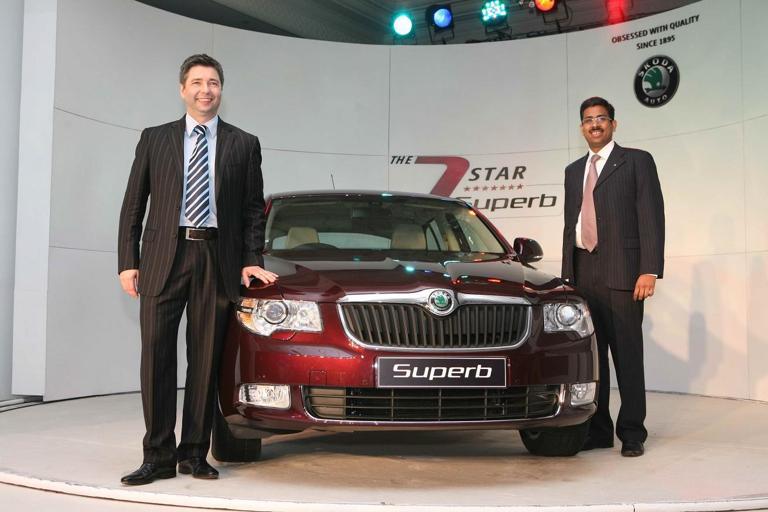} &
         \includegraphics[width=0.195\linewidth]{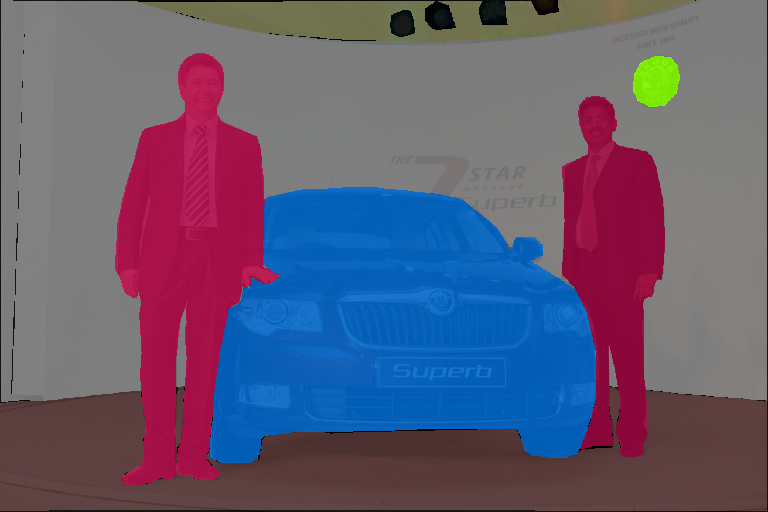} &
         \includegraphics[width=0.195\linewidth]{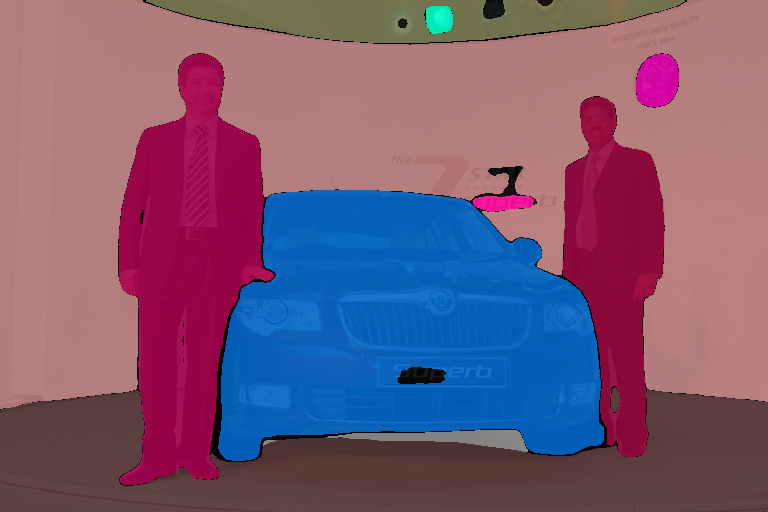} &
         \includegraphics[width=0.195\linewidth]{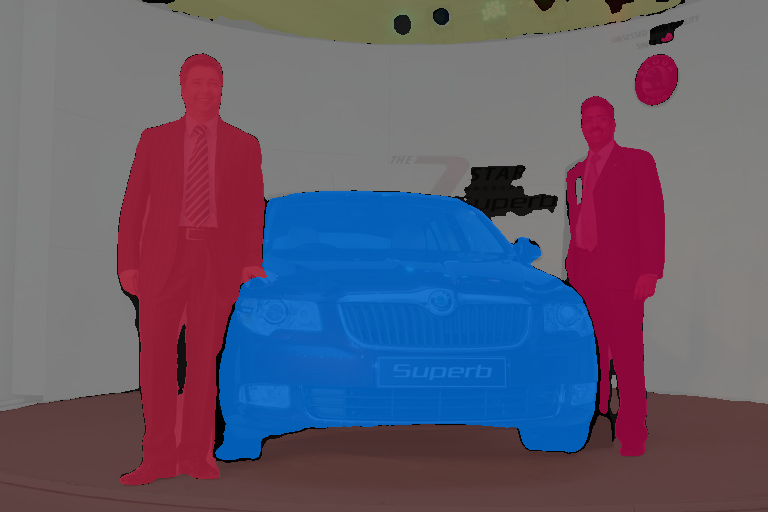} &
         \includegraphics[width=0.195\linewidth]{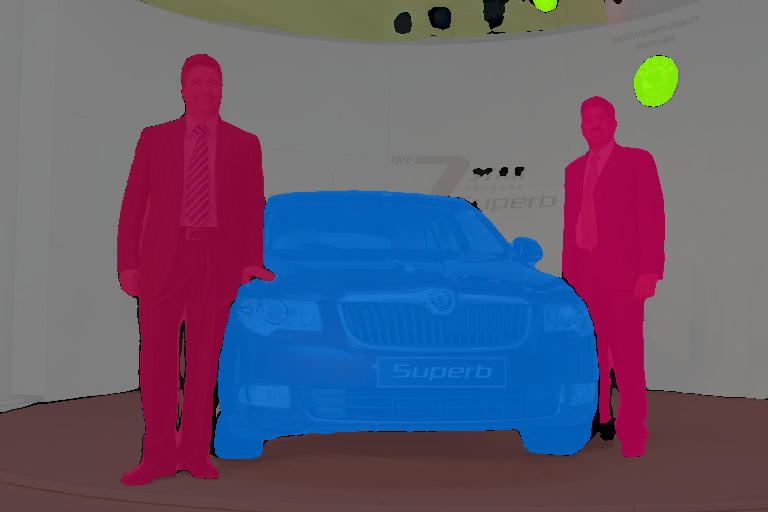} \\
         \includegraphics[width=0.195\linewidth]{fig/fix_supp/ade_panoptic/rgb/ADE_val_00000267.jpg} &
         \includegraphics[width=0.195\linewidth]{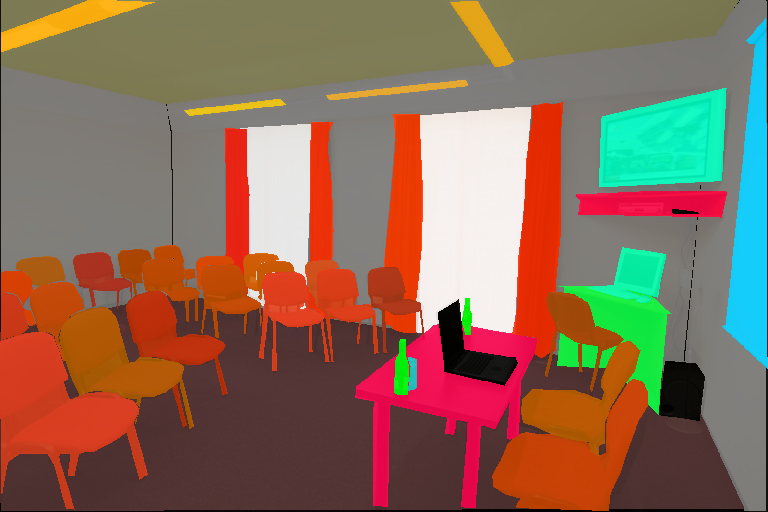} &
         \includegraphics[width=0.195\linewidth]{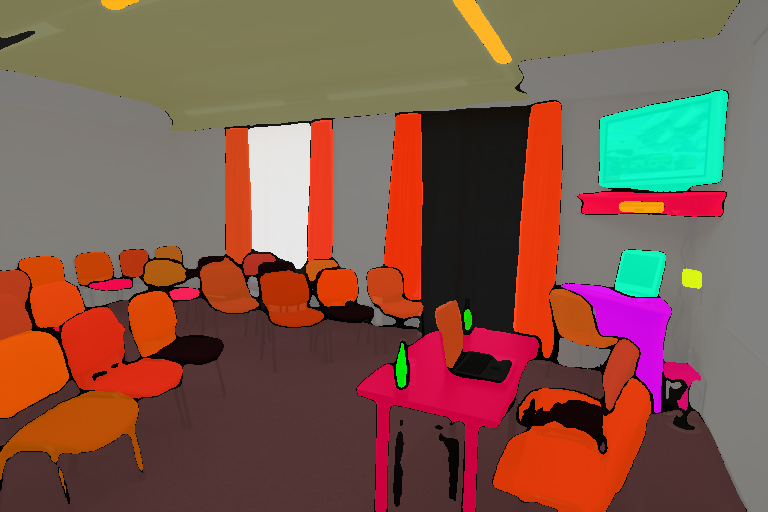} &
         \includegraphics[width=0.195\linewidth]{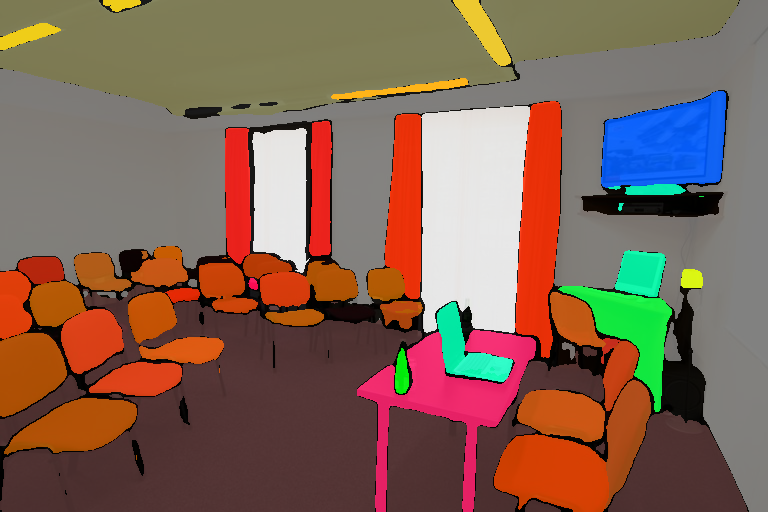} &
         \includegraphics[width=0.195\linewidth]{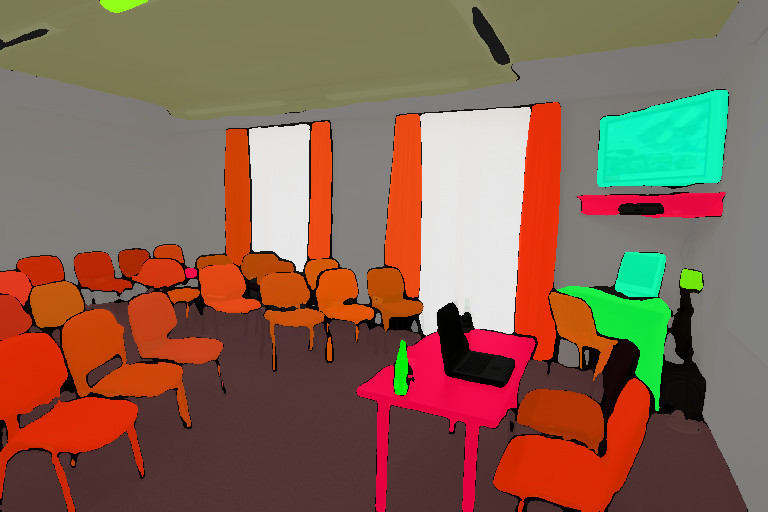} \\
         \includegraphics[width=0.195\linewidth]{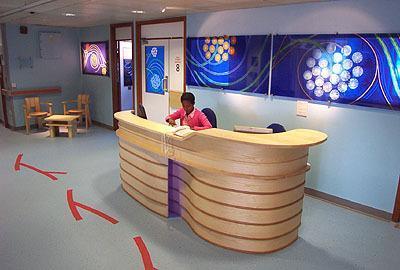} &
         \includegraphics[width=0.195\linewidth]{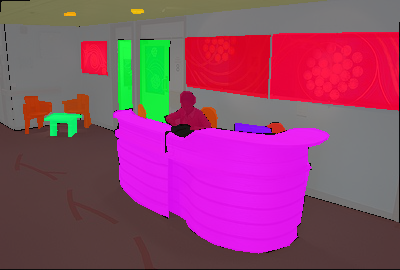} &
         \includegraphics[width=0.195\linewidth]{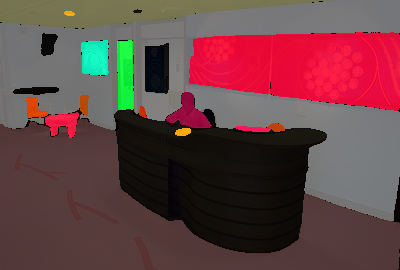} &
         \includegraphics[width=0.195\linewidth]{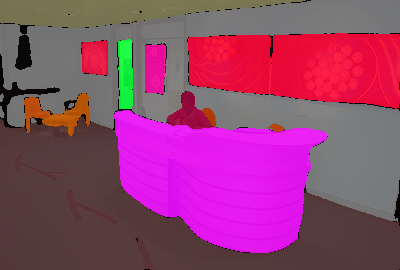} &
         \includegraphics[width=0.195\linewidth]{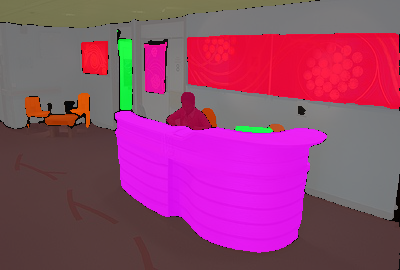} \\
         \includegraphics[width=0.195\linewidth]{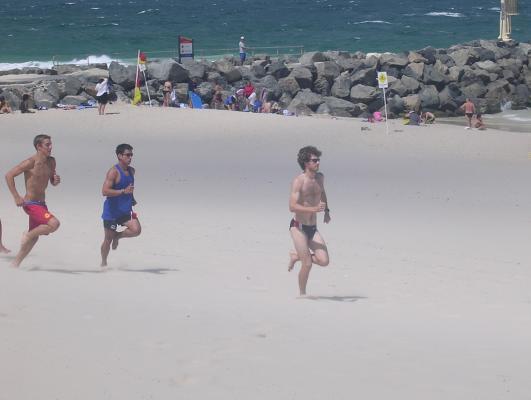} &
         \includegraphics[width=0.195\linewidth]{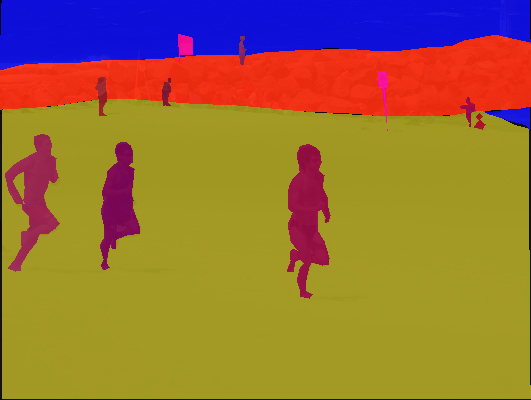} &
         \includegraphics[width=0.195\linewidth]{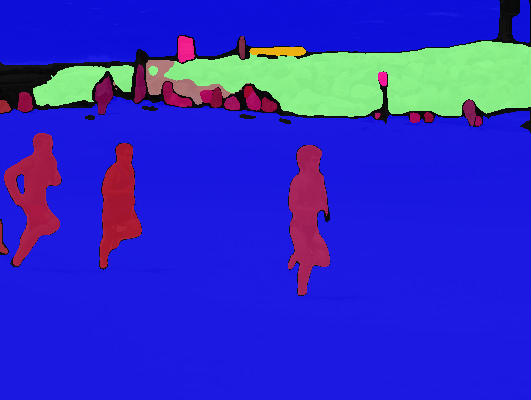} &
         \includegraphics[width=0.195\linewidth]{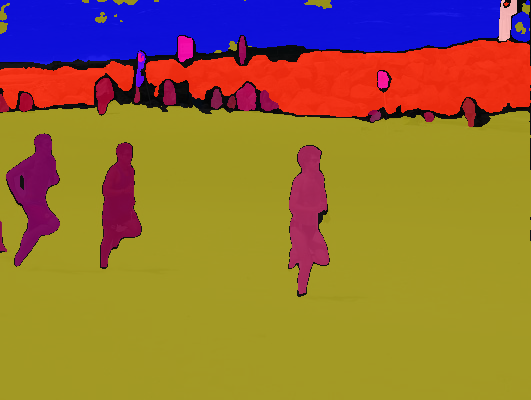} &
         \includegraphics[width=0.195\linewidth]{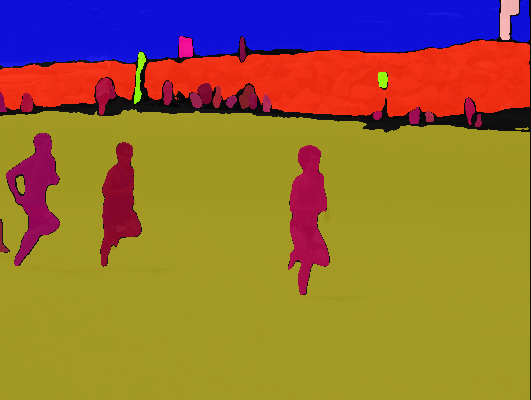} \\
                  \includegraphics[width=0.195\linewidth,height=7em]{fig/fix_supp/ade_panoptic/rgb/ADE_val_00001761.jpg} &
         \includegraphics[width=0.195\linewidth,height=7em]{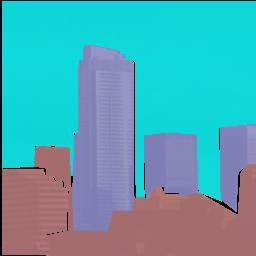} &
         \includegraphics[width=0.195\linewidth,height=7em]{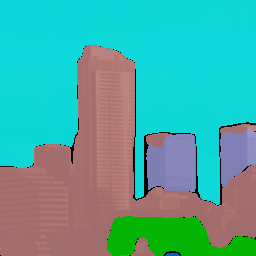} &
         \includegraphics[width=0.195\linewidth,height=7em]{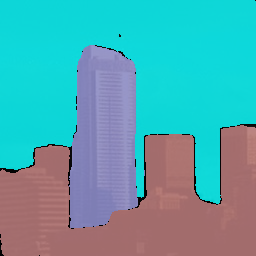} &
         \includegraphics[width=0.195\linewidth,height=7em]{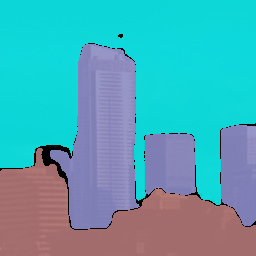} \\
         \includegraphics[width=0.195\linewidth]{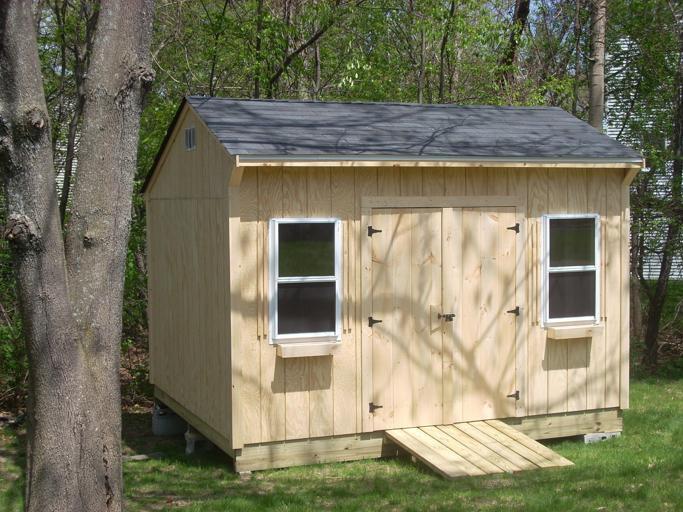} &
         \includegraphics[width=0.195\linewidth]{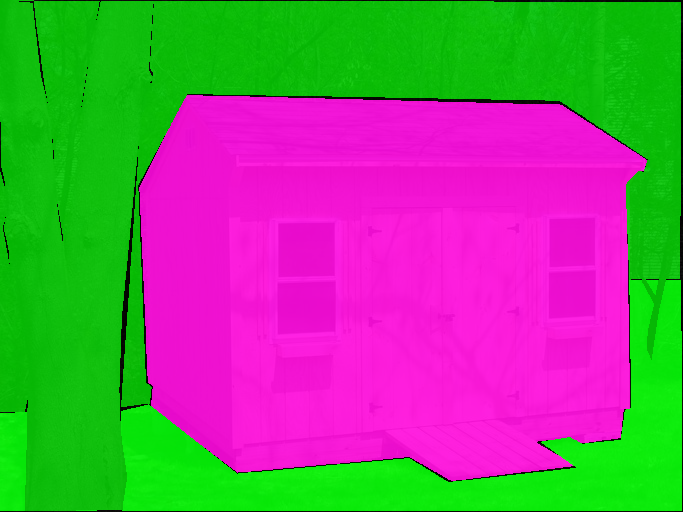} &
         \includegraphics[width=0.195\linewidth]{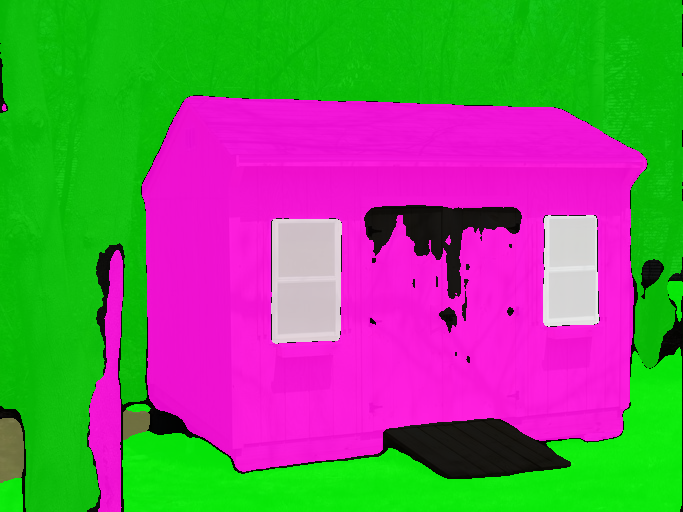} &
         \includegraphics[width=0.195\linewidth]{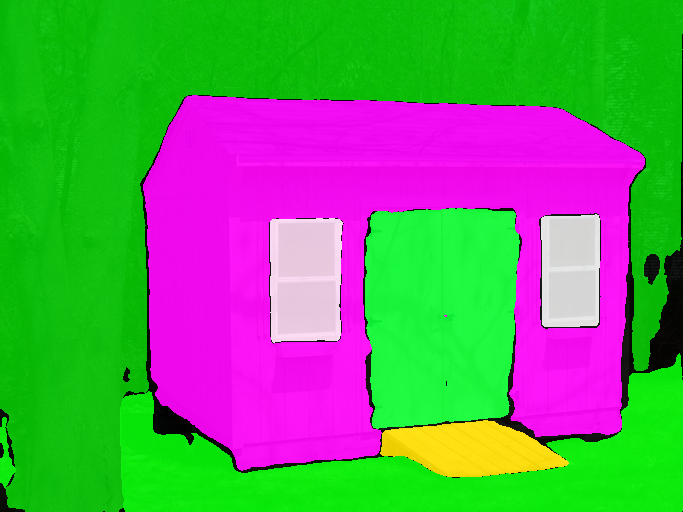} &
         \includegraphics[width=0.195\linewidth]{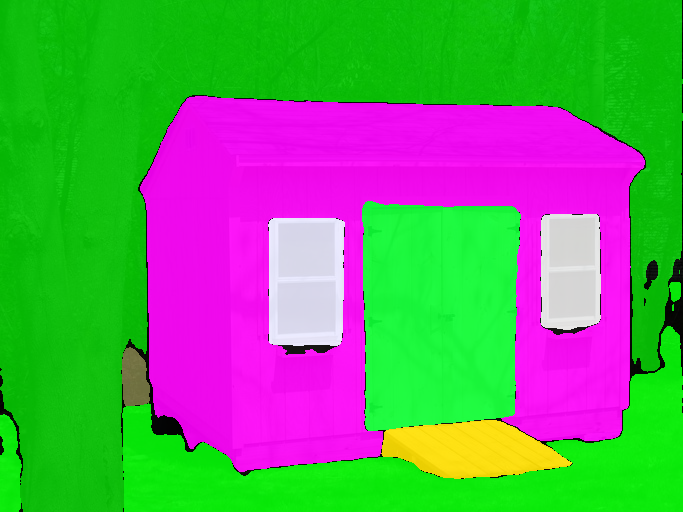} \\
    \end{tabular}
    \caption{Panoptic predictions of the baseline M2F+SSL CN-B-L2B (\cf \cref{tab:pq_ablations} in the main manuscript), \ours{}, and \ourspp{} on few examples from the ADE20K validation set. All models are trained on the ADE20K 1/128 data partition (\ie, only 158 labeled images).}
    \label{tab:ade-visualizations-panoptic}
\end{figure*}

%% file: tables/04_appendix_panoptic_predictions_coco-panoptic.tex
\begin{figure*}[]
    \setlength{\tabcolsep}{1pt}
    \centering
    
    \begin{tabular}{ccccc}
         Image & Ground truth & M2F+SSL Baseline & \textbf{\ours{}} & \textbf{\ourspp{}} \\
         \includegraphics[width=0.195\linewidth]{fig/fix_supp/coco-pan-panoptic/rgb/000000000139.jpg} &
         \includegraphics[width=0.195\linewidth]{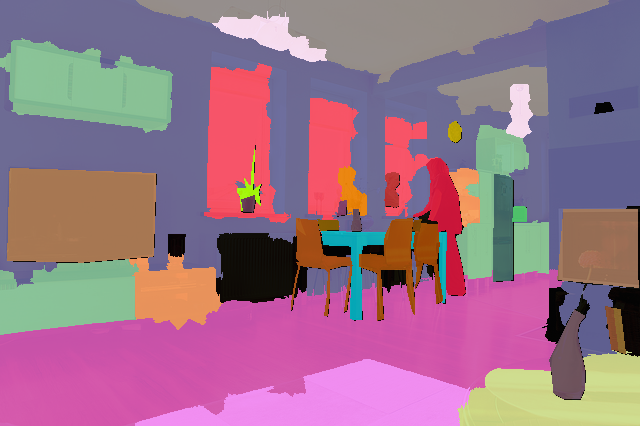} &
         \includegraphics[width=0.195\linewidth]{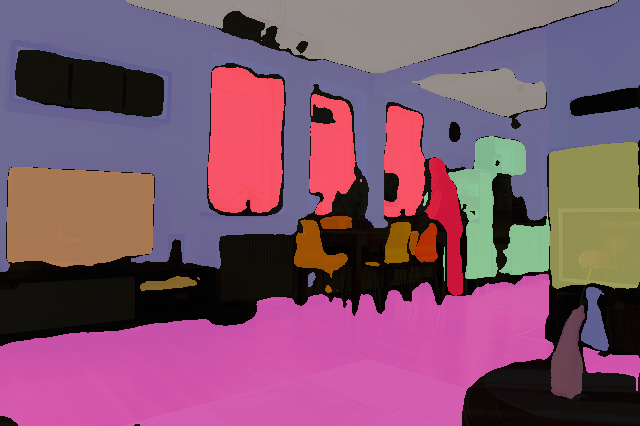} &
         \includegraphics[width=0.195\linewidth]{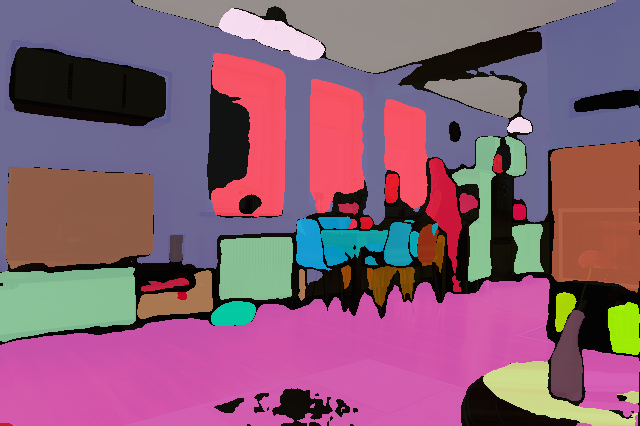} &
         \includegraphics[width=0.195\linewidth]{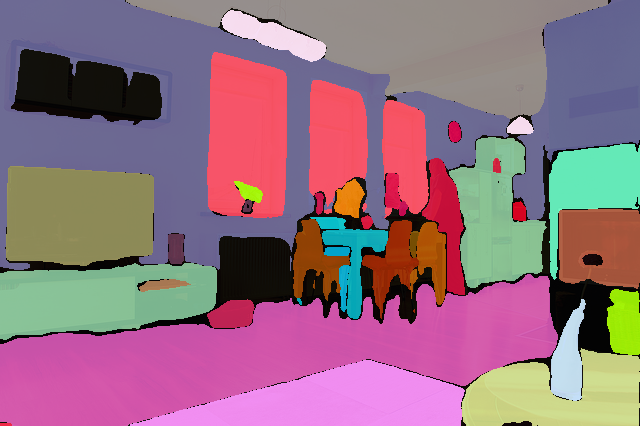} \\
         \includegraphics[width=0.195\linewidth,height=7em]{fig/fix_supp/coco-pan-panoptic/rgb/000000001761.jpg} &
         \includegraphics[width=0.195\linewidth,height=7em]{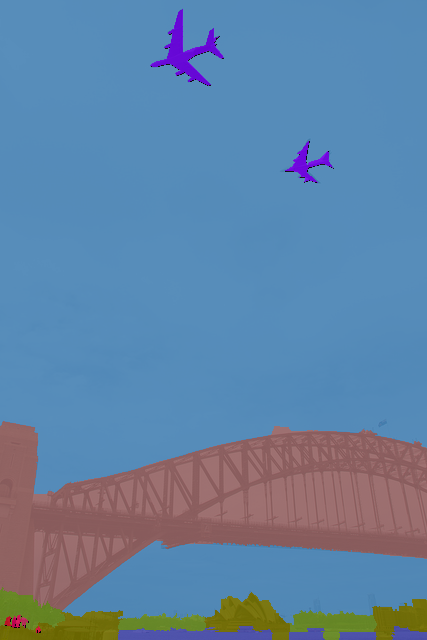} &
         \includegraphics[width=0.195\linewidth,height=7em]{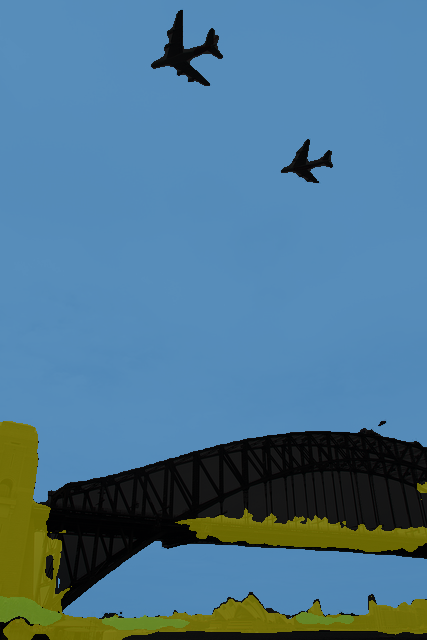} &
         \includegraphics[width=0.195\linewidth,height=7em]{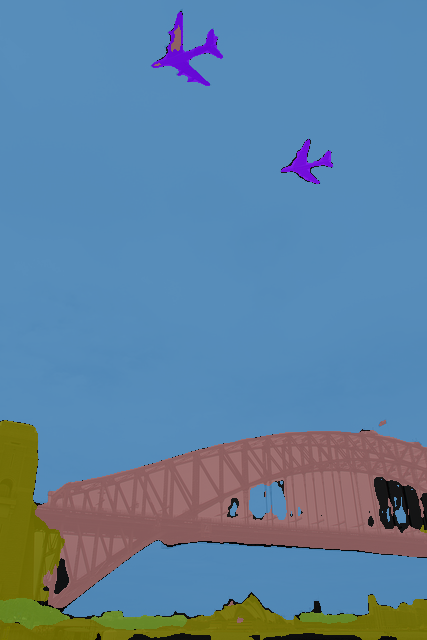} &
         \includegraphics[width=0.195\linewidth,height=7em]{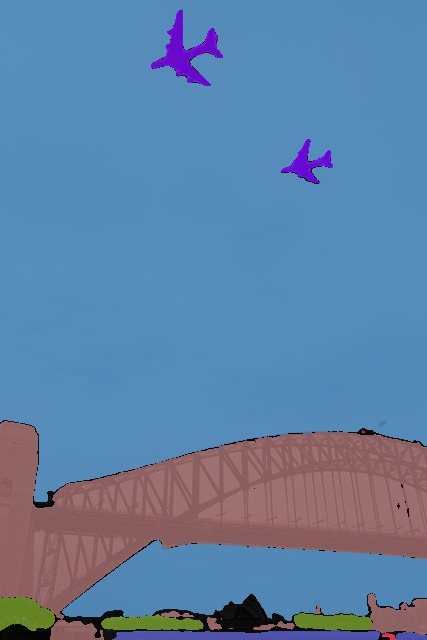} \\
         \includegraphics[width=0.195\linewidth]{fig/fix_supp/coco-pan-panoptic/rgb/000000005992.jpg} &
         \includegraphics[width=0.195\linewidth]{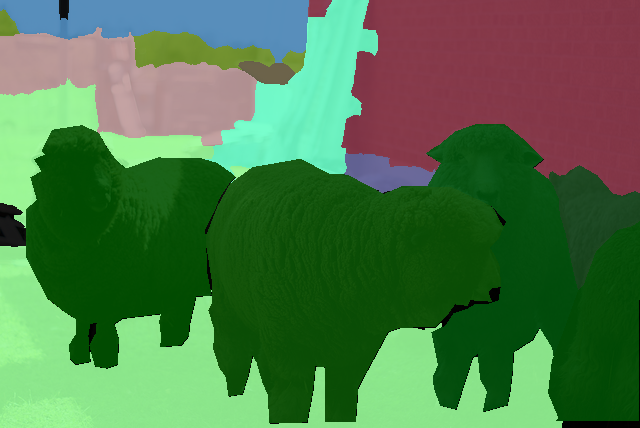} &
         \includegraphics[width=0.195\linewidth]{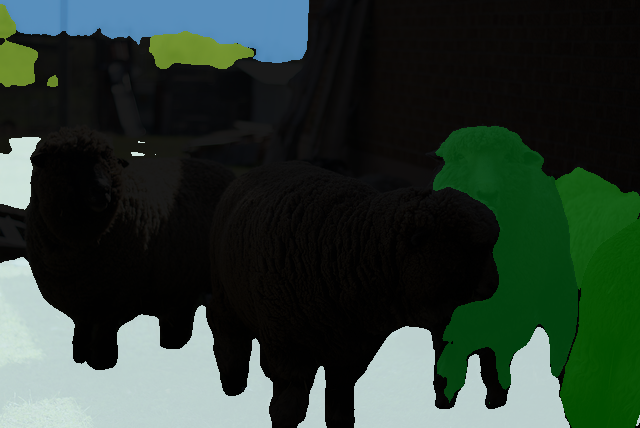} &
         \includegraphics[width=0.195\linewidth]{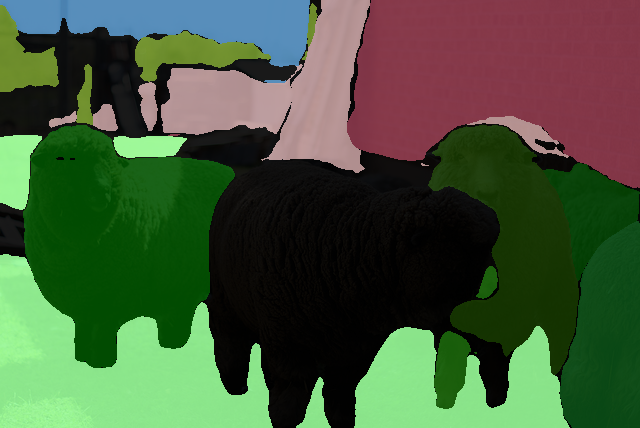} &
         \includegraphics[width=0.195\linewidth]{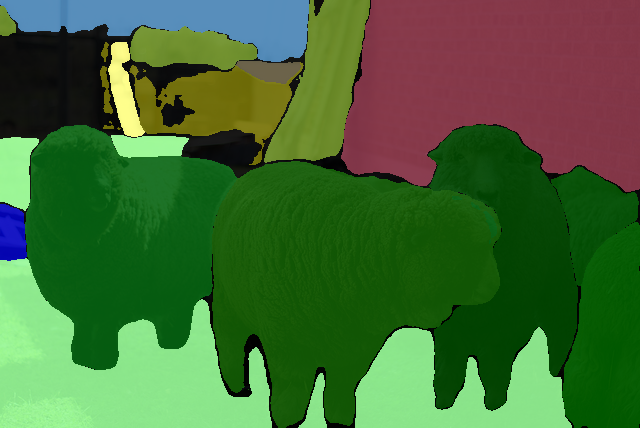} \\
         \includegraphics[width=0.195\linewidth]{fig/fix_supp/coco-pan-panoptic/rgb/000000006954.jpg} &
         \includegraphics[width=0.195\linewidth]{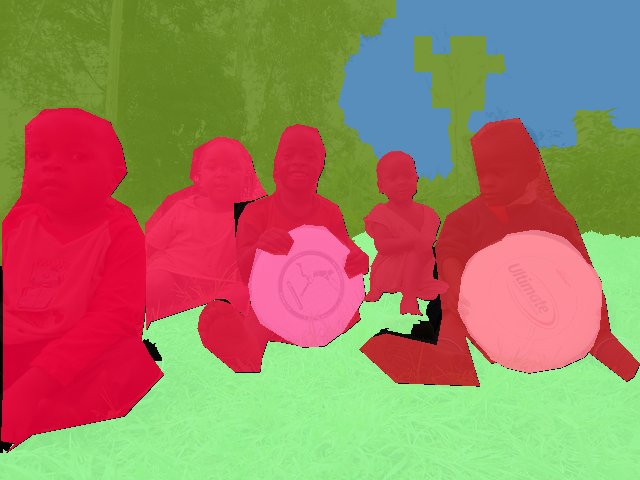} &
         \includegraphics[width=0.195\linewidth]{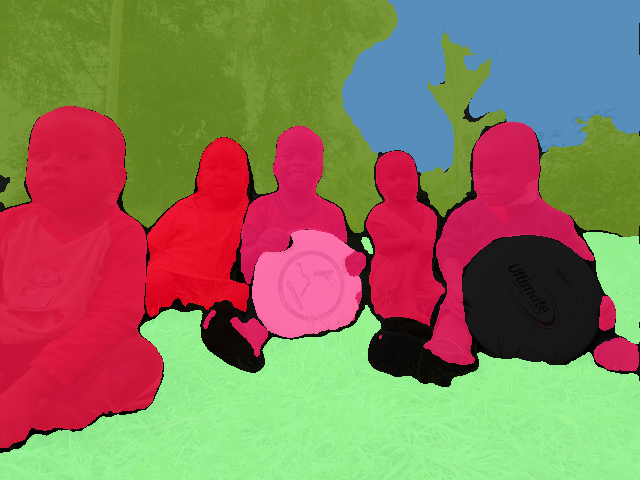} &
         \includegraphics[width=0.195\linewidth]{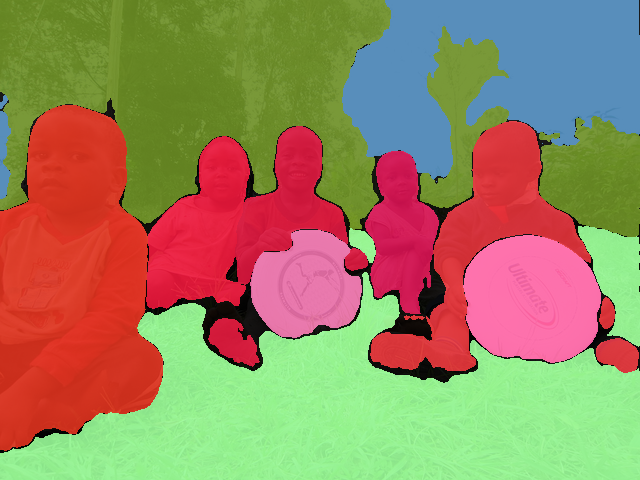} &
         \includegraphics[width=0.195\linewidth]{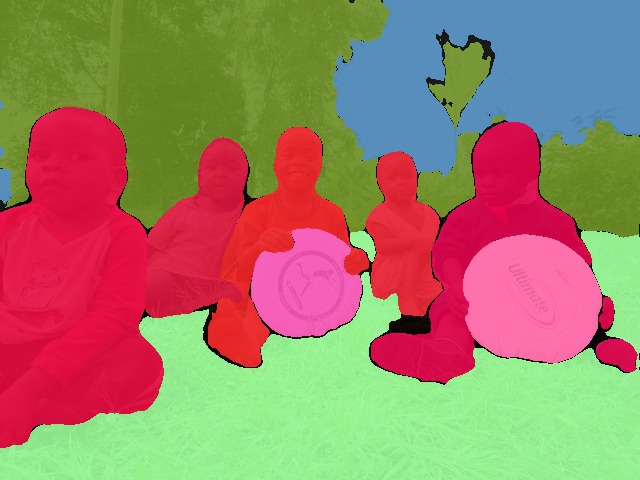} \\
         \includegraphics[width=0.195\linewidth]{fig/fix_supp/coco-pan-panoptic/rgb/000000026941.jpg} &
         \includegraphics[width=0.195\linewidth]{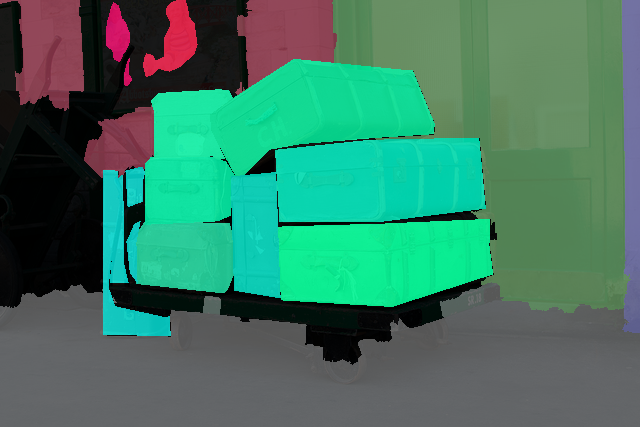} &
         \includegraphics[width=0.195\linewidth]{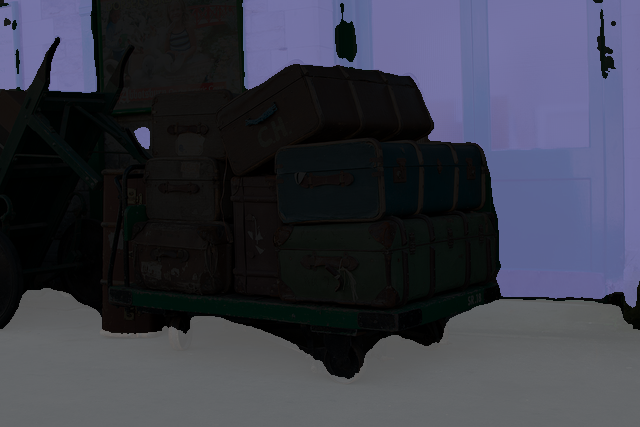} &
         \includegraphics[width=0.195\linewidth]{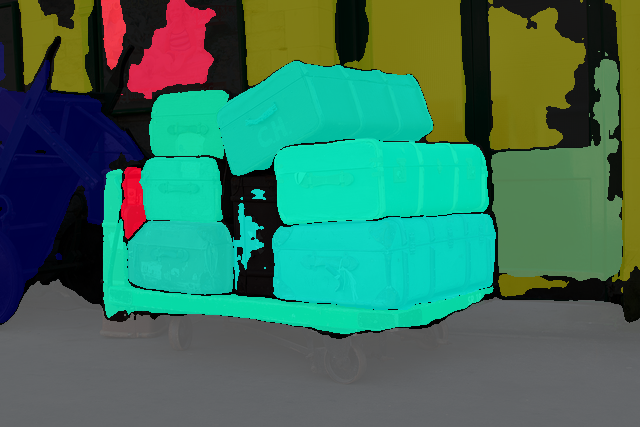} &
         \includegraphics[width=0.195\linewidth]{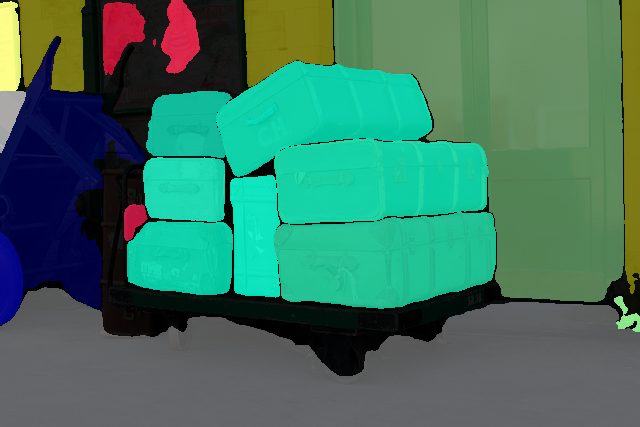} \\
         \includegraphics[width=0.195\linewidth]{fig/fix_supp/coco-pan-panoptic/rgb/000000032610.jpg} &
         \includegraphics[width=0.195\linewidth]{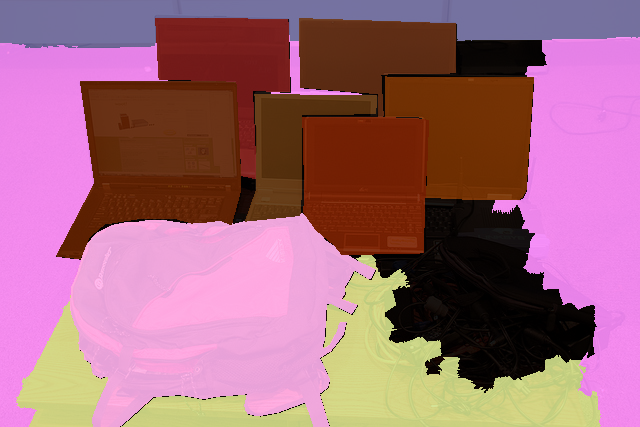} &
         \includegraphics[width=0.195\linewidth]{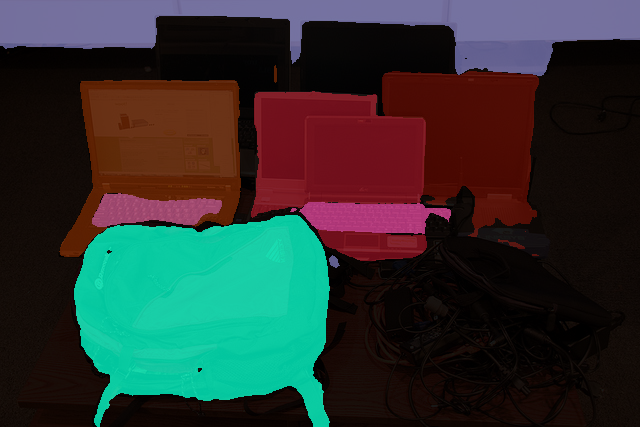} &
         \includegraphics[width=0.195\linewidth]{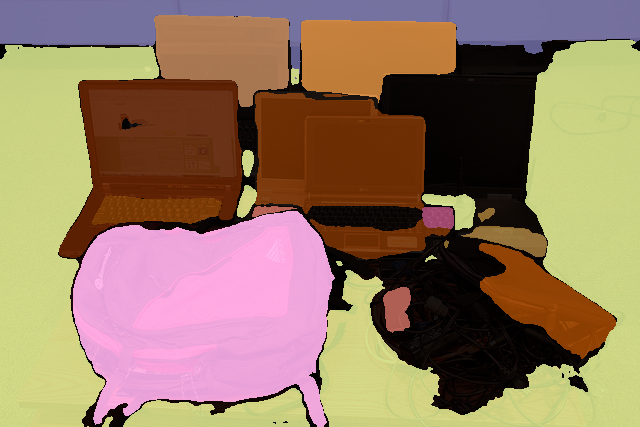} &
         \includegraphics[width=0.195\linewidth]{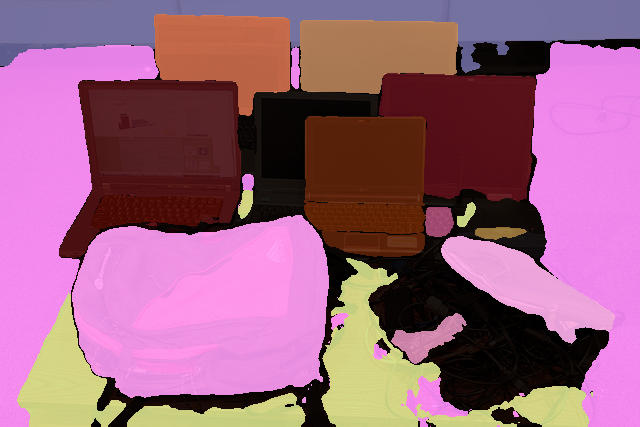} \\
         \includegraphics[width=0.195\linewidth]{fig/fix_supp/coco-pan-panoptic/rgb/000000041888.jpg} &
         \includegraphics[width=0.195\linewidth]{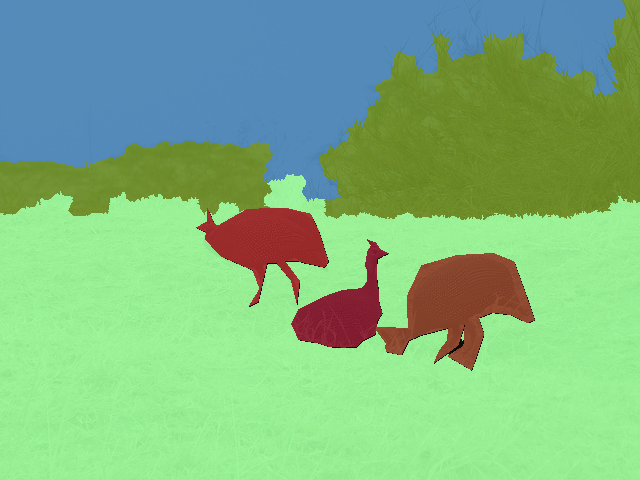} &
         \includegraphics[width=0.195\linewidth]{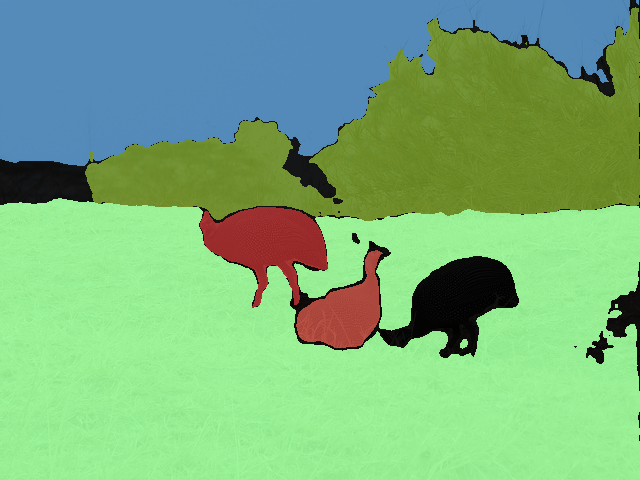} &
         \includegraphics[width=0.195\linewidth]{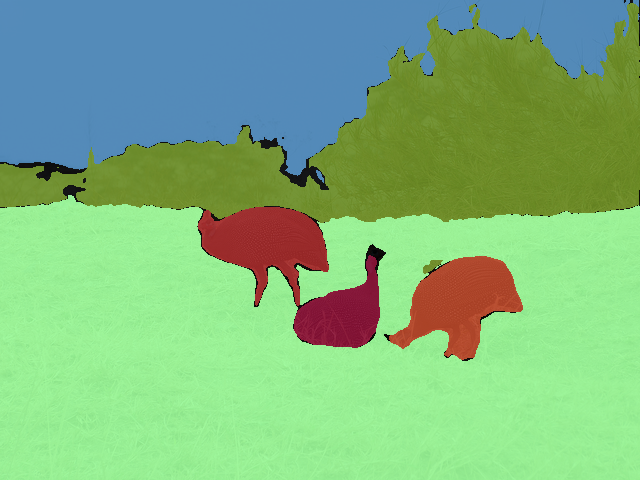} &
         \includegraphics[width=0.195\linewidth]{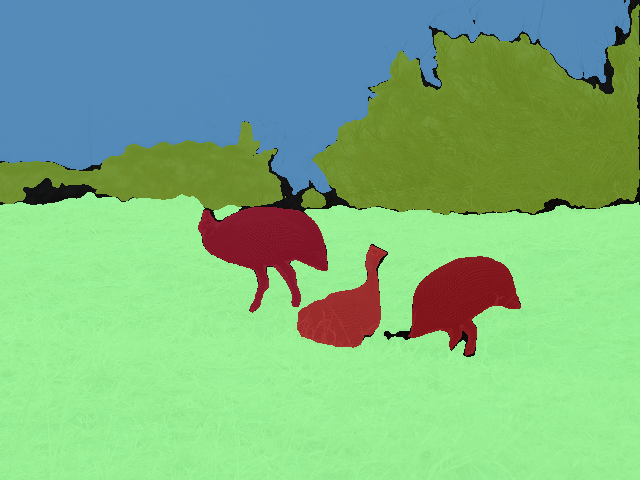} \\
         \includegraphics[width=0.195\linewidth,height=9em]{fig/fix_supp/coco-pan-panoptic/rgb/000000005060.jpg} &
         \includegraphics[width=0.195\linewidth,height=9em]{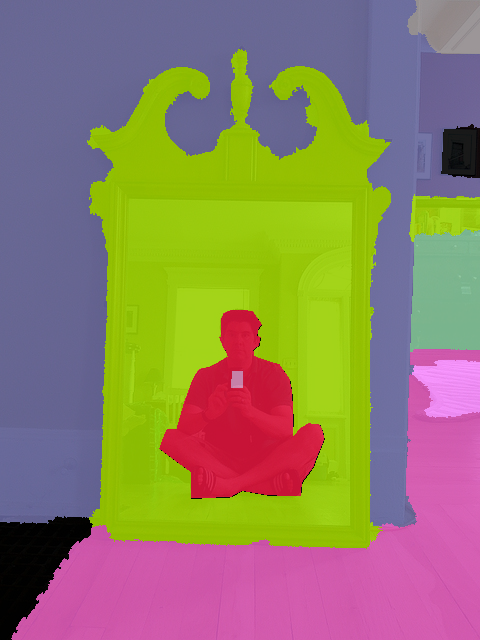} &
         \includegraphics[width=0.195\linewidth,height=9em]{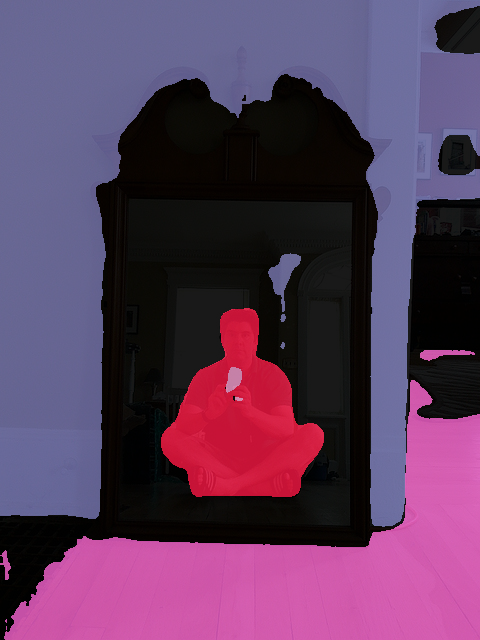} &
         \includegraphics[width=0.195\linewidth,height=9em]{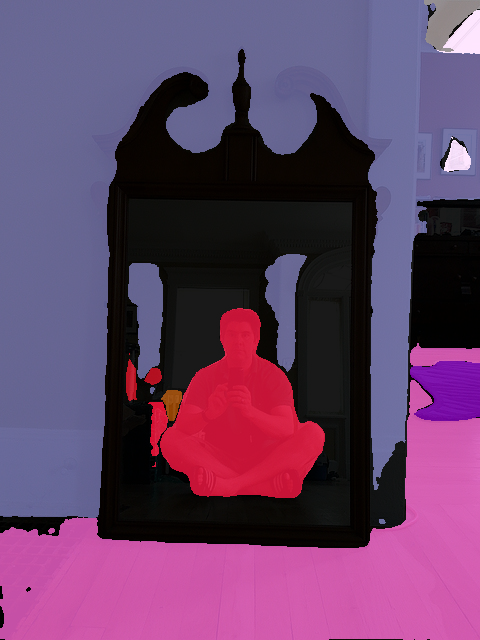} &
         \includegraphics[width=0.195\linewidth,height=9em]{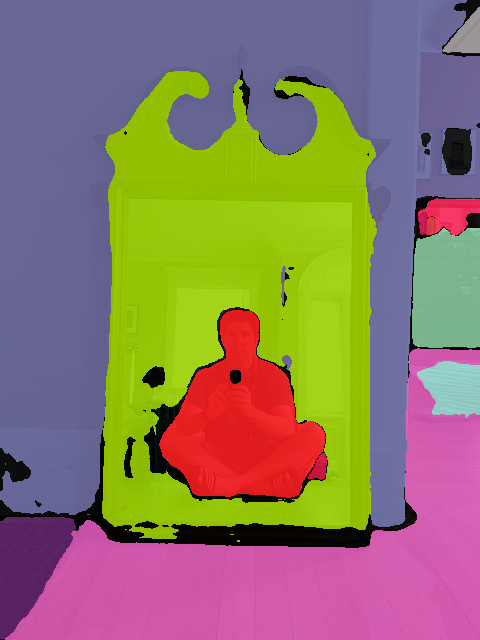} \\
         
    \end{tabular}
    \caption{Panoptic predictions of the baseline M2F+SSL CN-B-L2B (\cf \cref{tab:pq_ablations} in the main manuscript), \ours{}, and \ourspp{} on few examples from the COCO-Panoptic val. All models are trained on the COCO-Panoptic 1/512 data partition (\ie, only 232 labeled images).}
    \label{tab:coco-panoptic-visualizations-panoptic}
\end{figure*}

%% file: tables/04_appendix_sota_comparison_ade_visualizations.tex
\begin{figure*}[]
    \setlength{\tabcolsep}{1pt}
    \centering
    
    \begin{tabular}{cccc}
         Image & Ground truth & SemiVL~\cite{hoyer24eccv} & \textbf{\ours{}} \\
         \includegraphics[width=0.245\linewidth]{fig/fix_supp/ade_semantic/rgbs/ADE_val_00000084.jpg} &
         \includegraphics[width=0.245\linewidth]{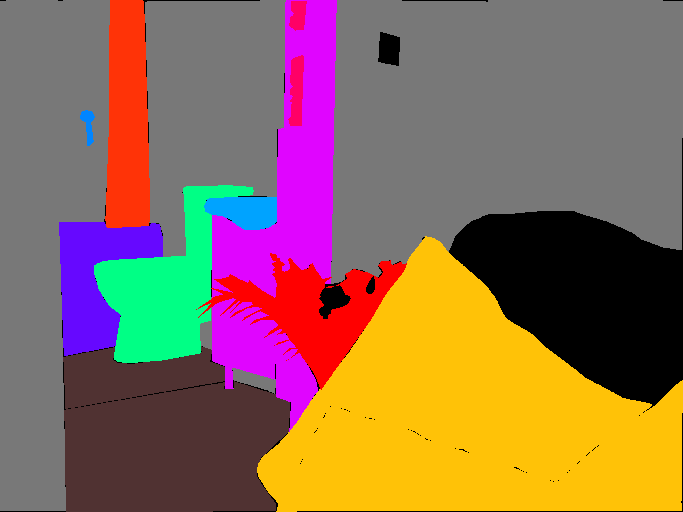} &
         \includegraphics[width=0.245\linewidth]{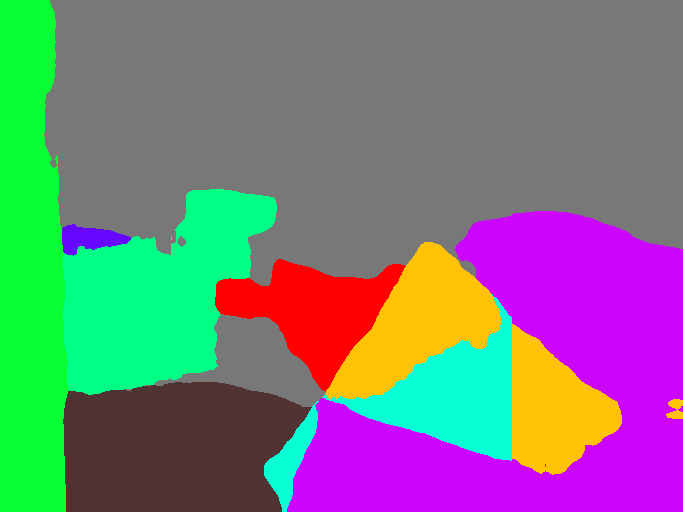} &
         \includegraphics[width=0.245\linewidth]{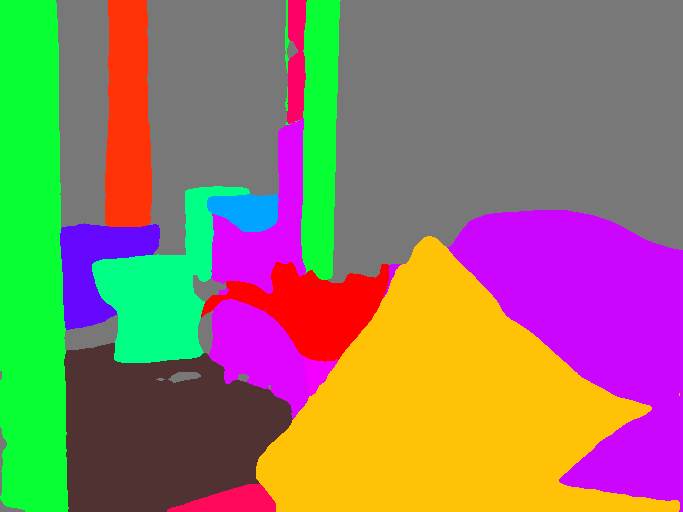}  \\
         \includegraphics[width=0.245\linewidth]{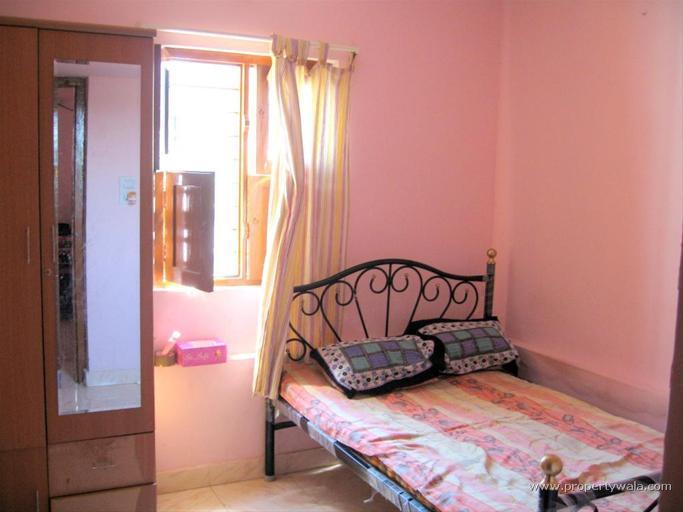} &
         \includegraphics[width=0.245\linewidth]{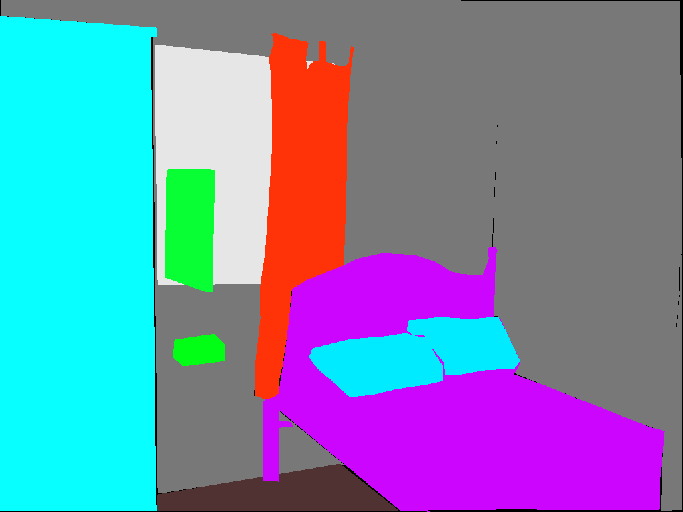} &
         \includegraphics[width=0.245\linewidth]{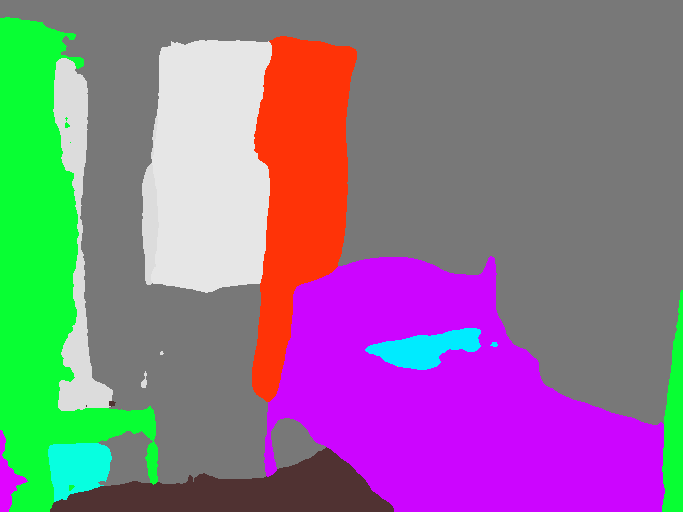} &
         \includegraphics[width=0.245\linewidth]{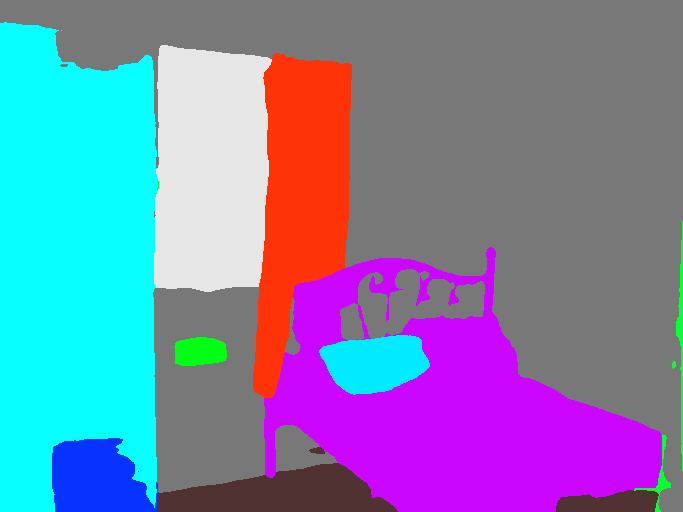}  \\
         \includegraphics[width=0.245\linewidth]{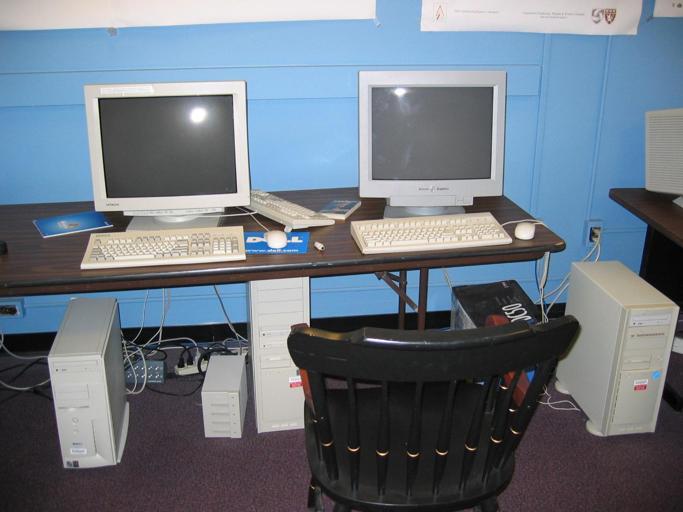} &
         \includegraphics[width=0.245\linewidth]{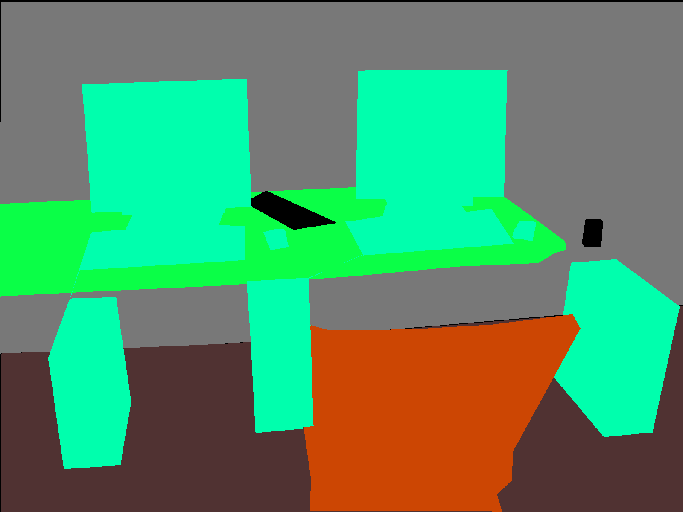} &
         \includegraphics[width=0.245\linewidth]{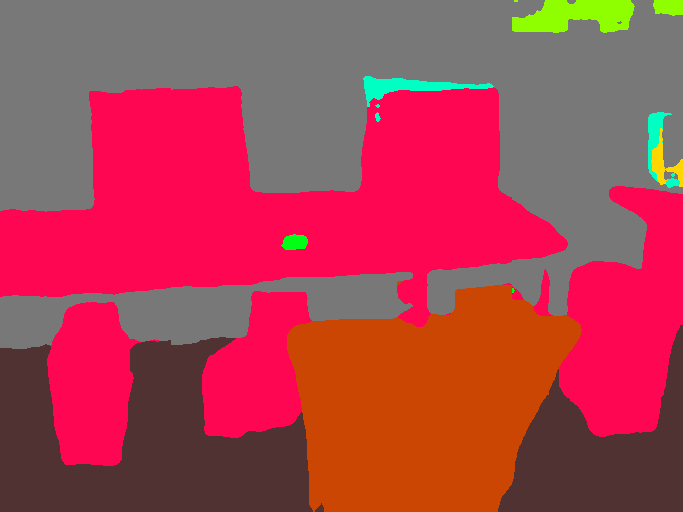} &
         \includegraphics[width=0.245\linewidth]{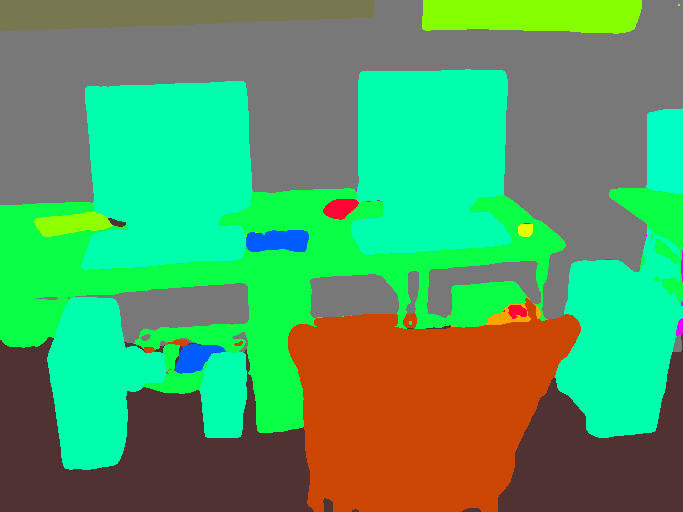}  \\
         \includegraphics[width=0.245\linewidth]{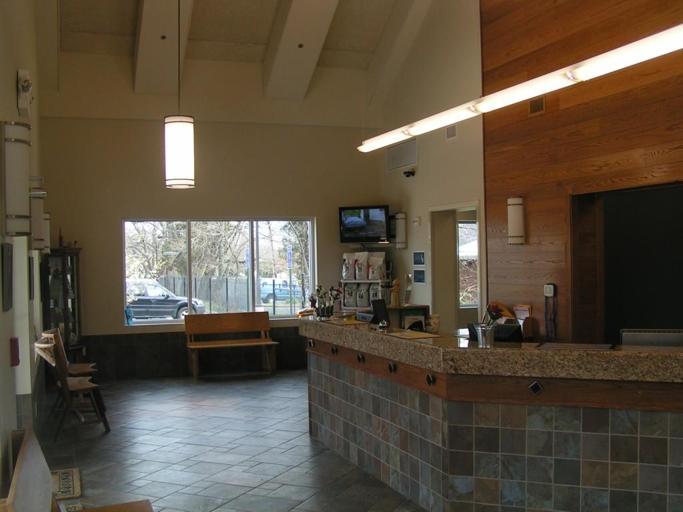} &
         \includegraphics[width=0.245\linewidth]{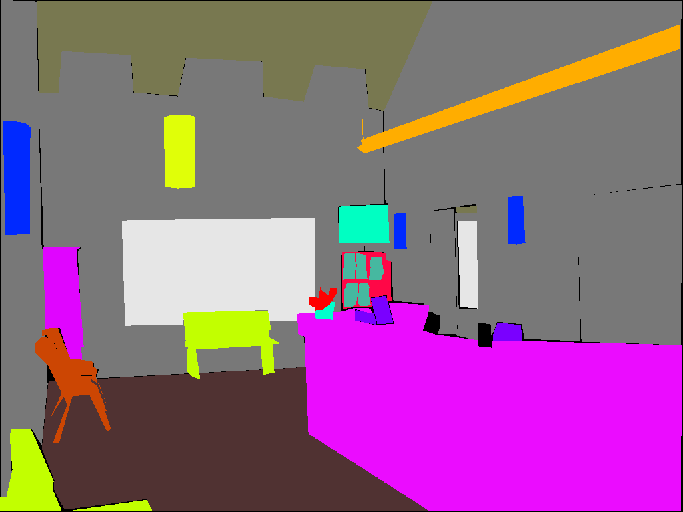} &
         \includegraphics[width=0.245\linewidth]{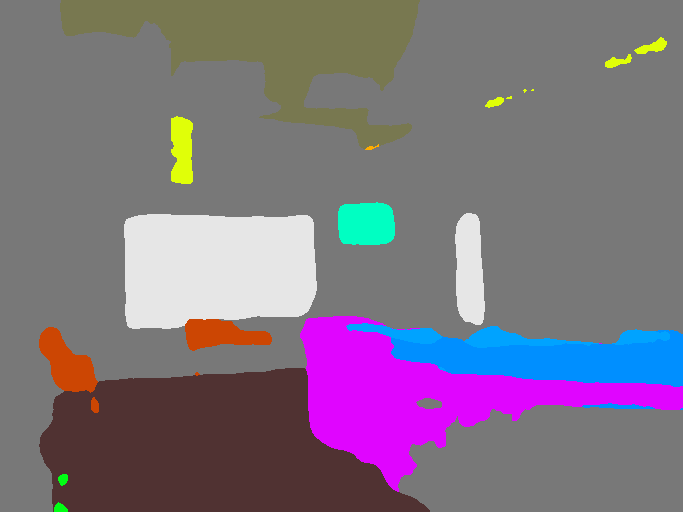} &
         \includegraphics[width=0.245\linewidth]{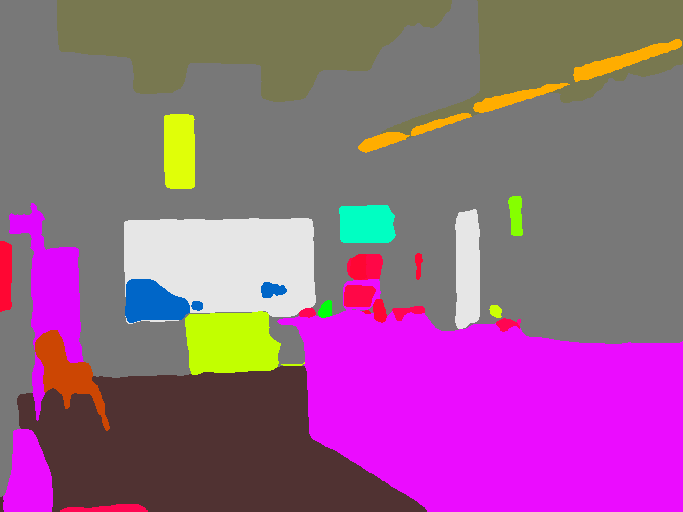}  \\
         \includegraphics[width=0.245\linewidth]{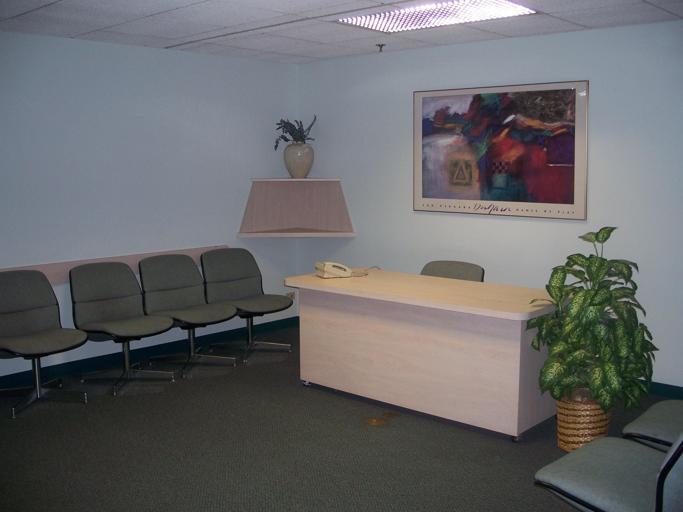} &
         \includegraphics[width=0.245\linewidth]{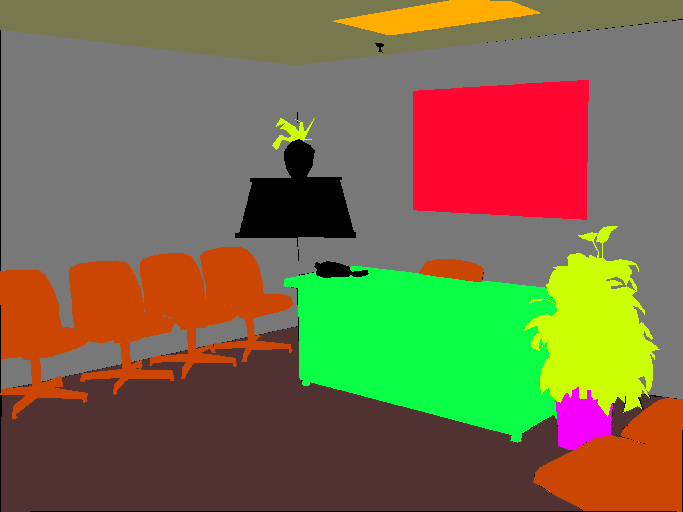} &
         \includegraphics[width=0.245\linewidth]{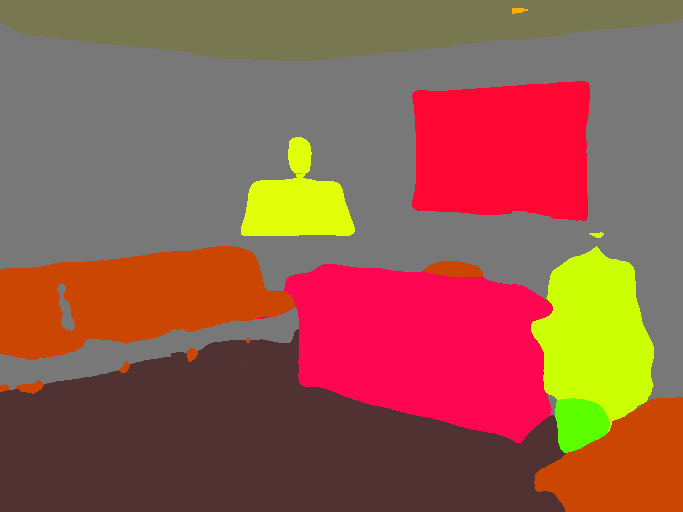} &
         \includegraphics[width=0.245\linewidth]{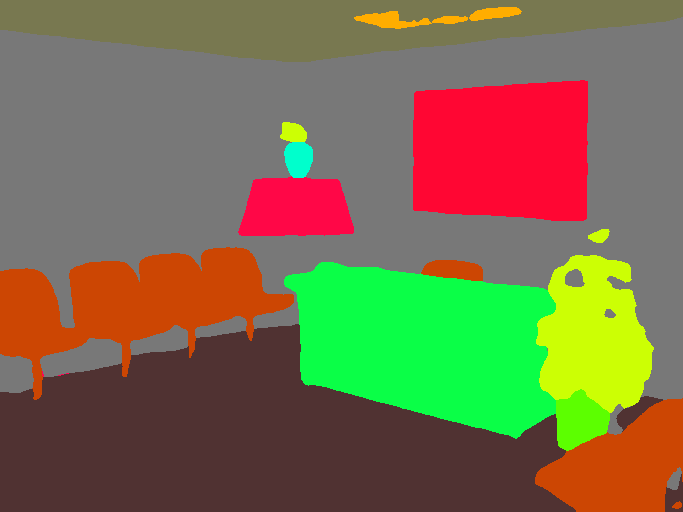}  \\
         \includegraphics[width=0.245\linewidth]{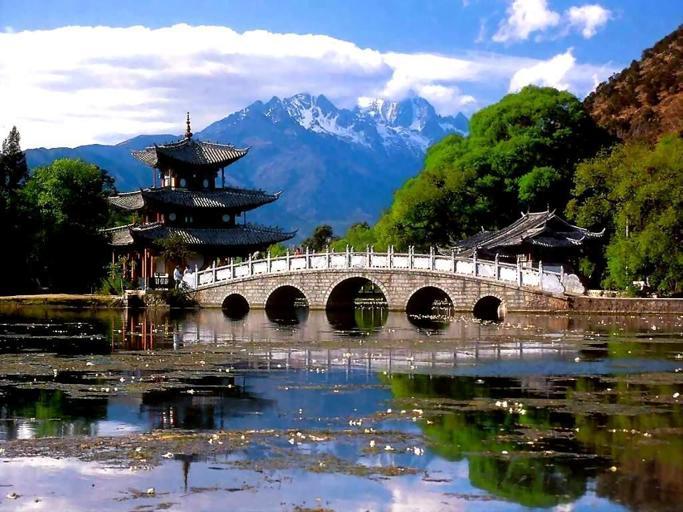} &
         \includegraphics[width=0.245\linewidth]{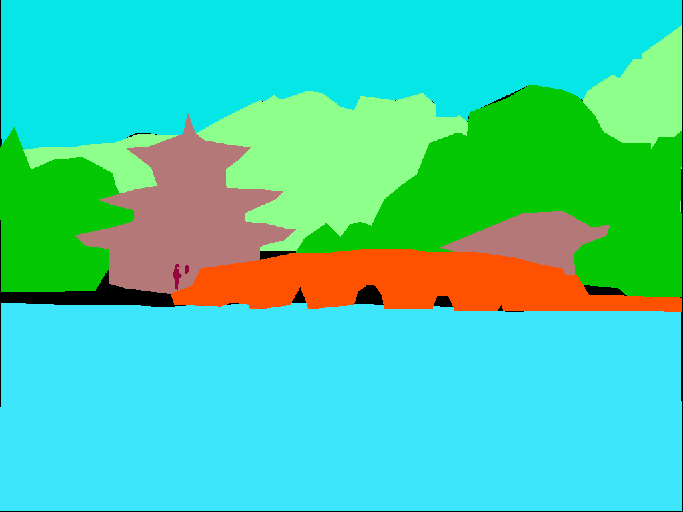} &
         \includegraphics[width=0.245\linewidth]{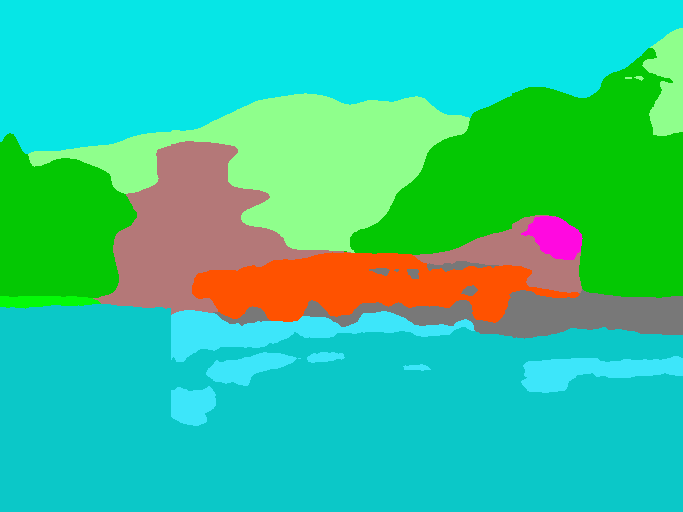} &
         \includegraphics[width=0.245\linewidth]{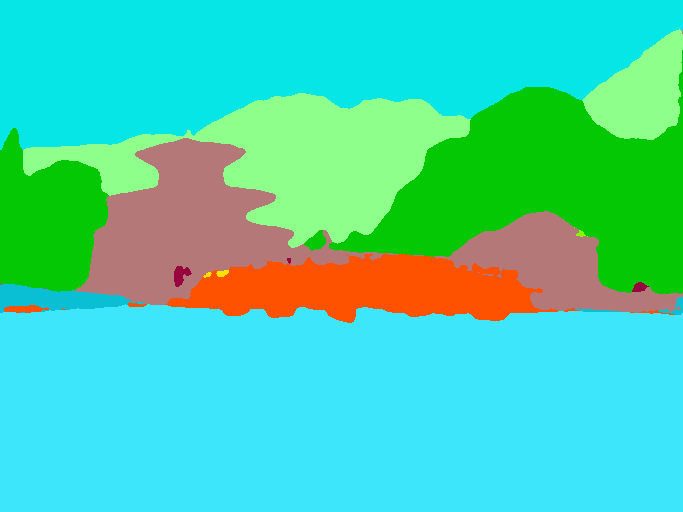}  \\
          
    \end{tabular}
    \caption{Qualitative comparison of \ours{} with the state-of-the-art method SemiVL~\cite{hoyer24eccv} for semantic segmentation. Models are trained on the ADE20K 1/128 partition (\ie, 158 labeled images), with predictions visualized on examples from the ADE20K validation set. Predictions for SemiVL~\cite{hoyer24eccv} are generated using the publicly released checkpoint.}
    \label{tab:ade-visualizations-comparison-sota}
\end{figure*}

%% file: tables/04_apendix_sota_comparison_coco-objects-visualization.tex
\begin{figure*}[]
    \setlength{\tabcolsep}{1pt}
    \centering
    
    \begin{tabular}{cccc}
         Image & Ground truth & SemiVL~\cite{hoyer24eccv} & \textbf{\ours{}} \\
         \includegraphics[width=0.245\linewidth]{fig/fix_supp/coco-objects-semantic/rgb/000000001490.jpg} &
         \includegraphics[width=0.245\linewidth]{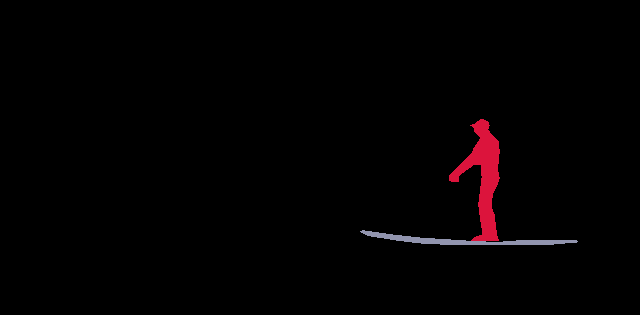} &
         \includegraphics[width=0.245\linewidth]{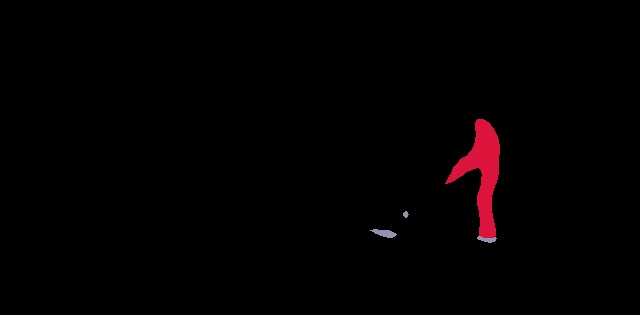} &
         \includegraphics[width=0.245\linewidth]{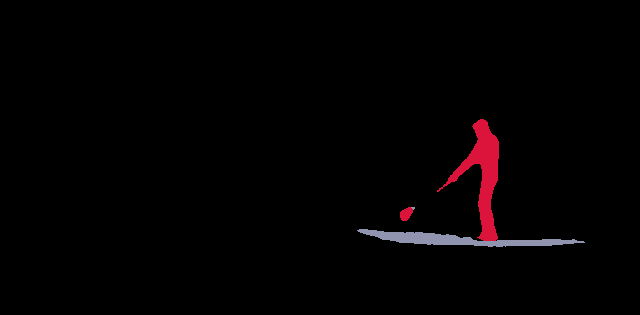}  \\
         \includegraphics[width=0.245\linewidth]{fig/fix_supp/coco-objects-semantic/rgb/000000001532.jpg} &
         \includegraphics[width=0.245\linewidth]{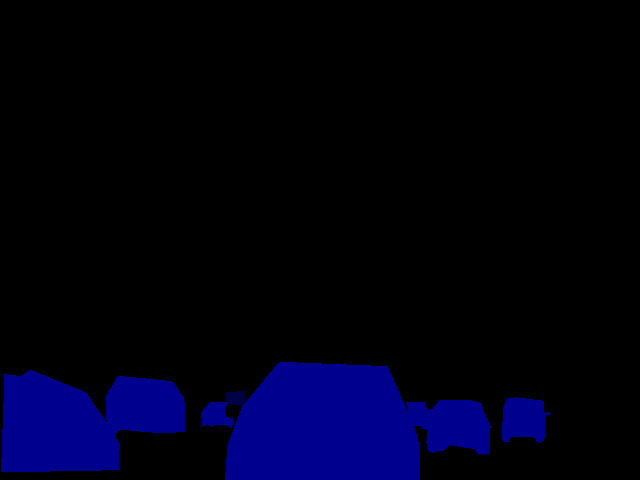} &
         \includegraphics[width=0.245\linewidth]{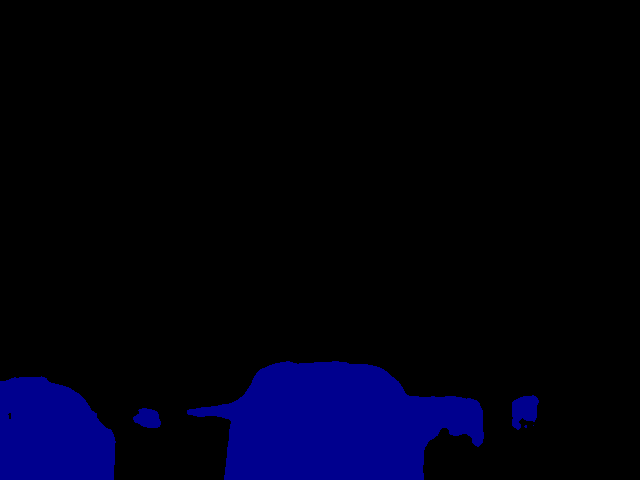} &
         \includegraphics[width=0.245\linewidth]{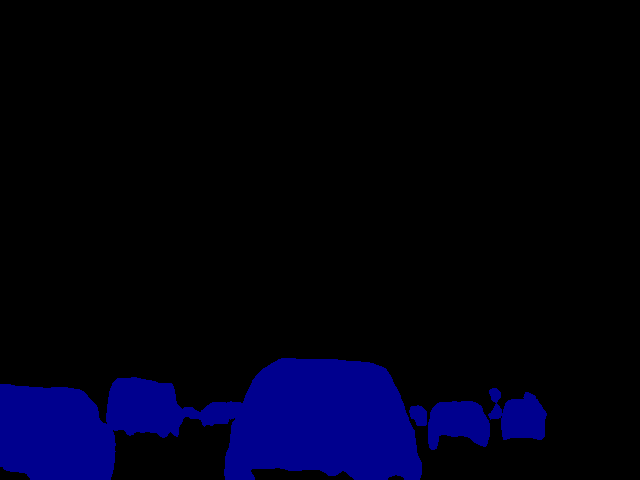}  \\
         \includegraphics[width=0.245\linewidth]{fig/fix_supp/coco-objects-semantic/rgb/000000007818.jpg} &
         \includegraphics[width=0.245\linewidth]{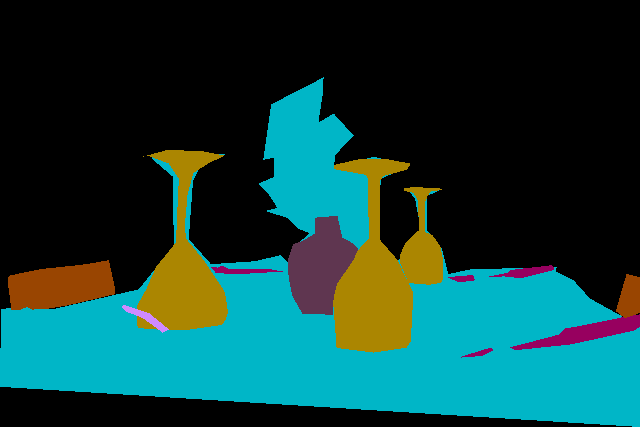} &
         \includegraphics[width=0.245\linewidth]{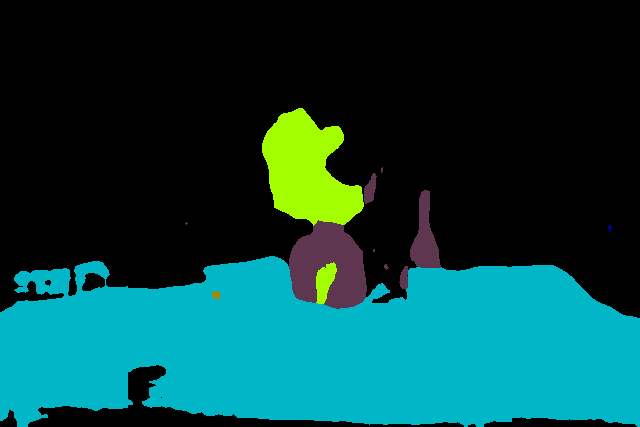} &
         \includegraphics[width=0.245\linewidth]{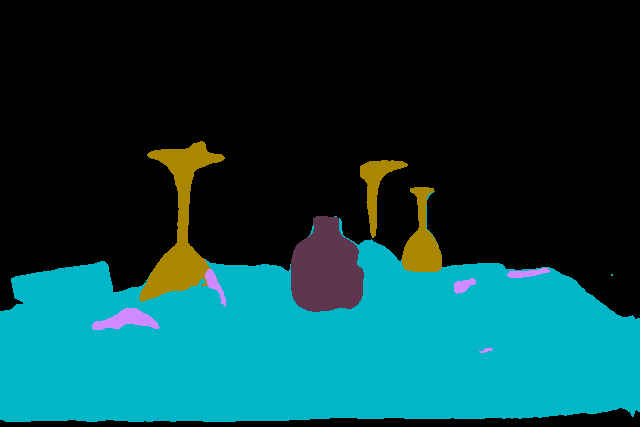}  \\
         \includegraphics[width=0.245\linewidth]{fig/fix_supp/coco-objects-semantic/rgb/000000009590.jpg} &
         \includegraphics[width=0.245\linewidth]{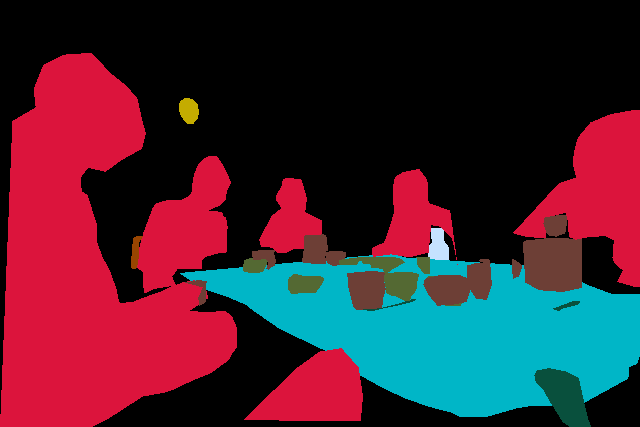} &
         \includegraphics[width=0.245\linewidth]{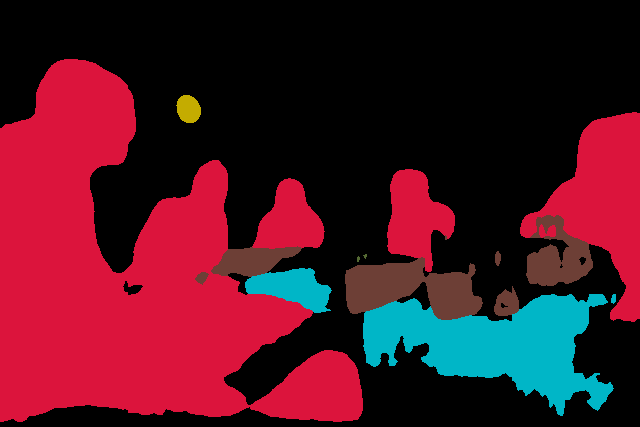} &
         \includegraphics[width=0.245\linewidth]{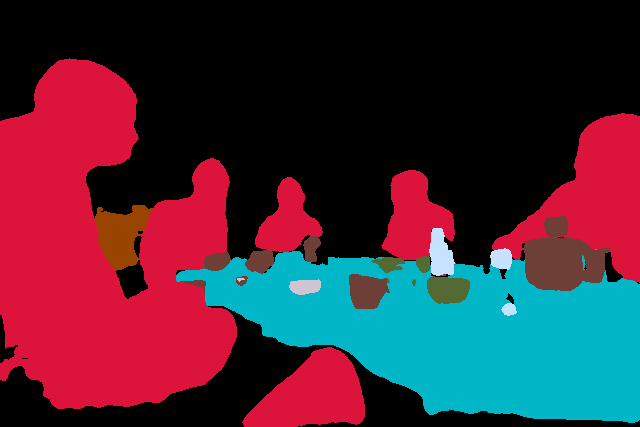}  \\
         \includegraphics[width=0.245\linewidth]{fig/fix_supp/coco-objects-semantic/rgb/000000011149.jpg} &
         \includegraphics[width=0.245\linewidth]{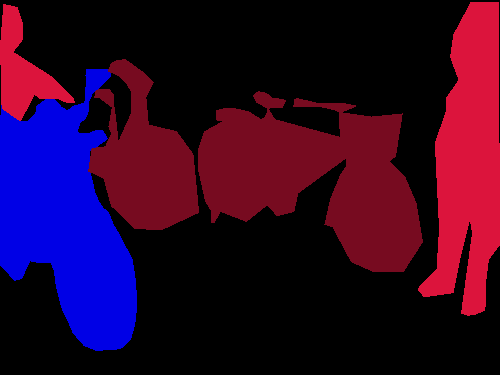} &
         \includegraphics[width=0.245\linewidth]{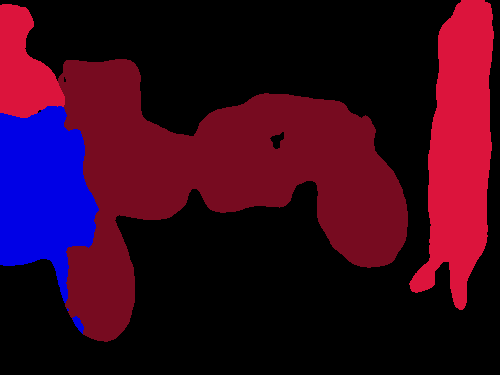} &
         \includegraphics[width=0.245\linewidth]{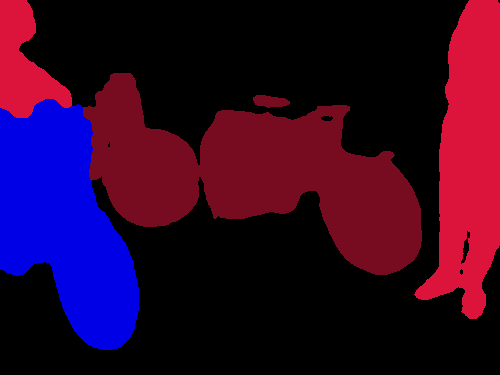}  \\
         \includegraphics[width=0.245\linewidth]{fig/fix_supp/coco-objects-semantic/rgb/000000022589.jpg} &
         \includegraphics[width=0.245\linewidth]{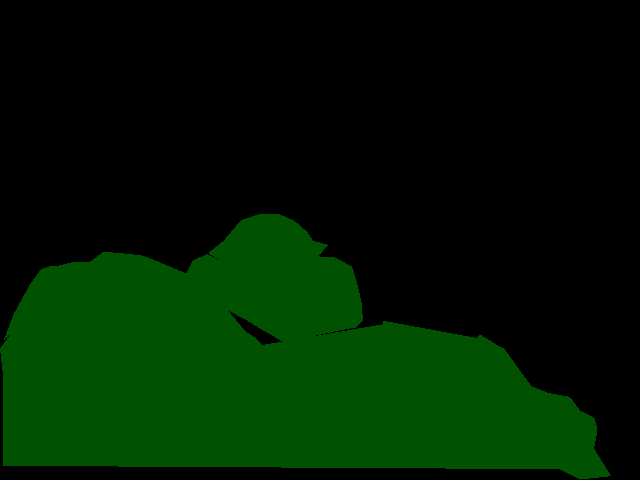} &
         \includegraphics[width=0.245\linewidth]{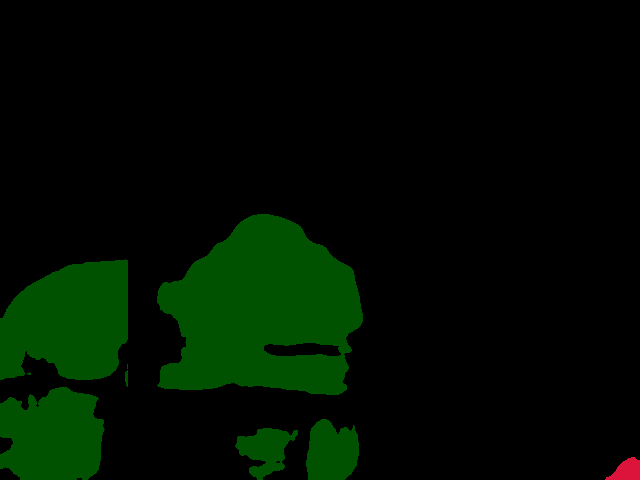} &
         \includegraphics[width=0.245\linewidth]{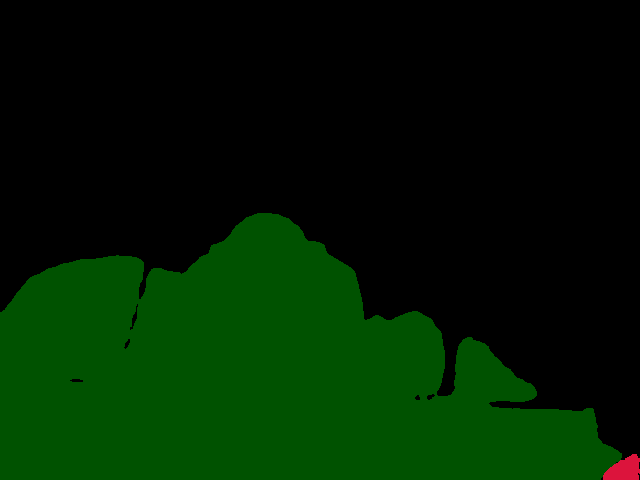}  \\
          
    \end{tabular}
    \caption{Qualitative comparison of \ours{} with the state-of-the-art method SemiVL~\cite{hoyer24eccv} for semantic segmentation. Models are trained on the COCO-Objects 1/512 partition (\ie, 232 labeled images), with predictions visualized on examples from the COCO-Objects validation subset. In COCO-Objects, the \voidtt{} class is represented in black and included in the evaluation. Predictions for SemiVL~\cite{hoyer24eccv} are generated using the publicly released checkpoint.}
    \label{tab:coco-objects-visualizations-comparison-sota}
\end{figure*}

%% file: main.bbl
\begin{thebibliography}{85}
\providecommand{\natexlab}[1]{#1}
\providecommand{\url}[1]{\texttt{#1}}
\expandafter\ifx\csname urlstyle\endcsname\relax
  \providecommand{\doi}[1]{doi: #1}\else
  \providecommand{\doi}{doi: \begingroup \urlstyle{rm}\Url}\fi

\bibitem[Blum and Mitchell(1998)]{blum98cclt}
Avrim Blum and Tom Mitchell.
\newblock Combining labeled and unlabeled data with co-training.
\newblock In \emph{Proceedings of the eleventh annual conference on Computational learning theory}, pages 92--100, 1998.

\bibitem[Bortsova et~al.(2019)Bortsova, Dubost, Hogeweg, Katramados, and De~Bruijne]{bortsova19miccai}
Gerda Bortsova, Florian Dubost, Laurens Hogeweg, Ioannis Katramados, and Marleen De~Bruijne.
\newblock Semi-supervised medical image segmentation via learning consistency under transformations.
\newblock In \emph{Medical Image Computing and Computer Assisted Intervention--MICCAI 2019: 22nd International Conference, Shenzhen, China, October 13--17, 2019, Proceedings, Part VI 22}, pages 810--818. Springer, 2019.

\bibitem[Carion et~al.(2020)Carion, Massa, Synnaeve, Usunier, Kirillov, and Zagoruyko]{carion20eccv}
Nicolas Carion, Francisco Massa, Gabriel Synnaeve, Nicolas Usunier, Alexander Kirillov, and Sergey Zagoruyko.
\newblock End-to-end object detection with transformers.
\newblock In \emph{European conference on computer vision}, pages 213--229. Springer, 2020.

\bibitem[Cha et~al.(2021)Cha, Yoon, Yeo, Huh, and Han]{cha21jcm}
Jun-Young Cha, Hyung-In Yoon, In-Sung Yeo, Kyung-Hoe Huh, and Jung-Suk Han.
\newblock Panoptic segmentation on panoramic radiographs: Deep learning-based segmentation of various structures including maxillary sinus and mandibular canal.
\newblock \emph{Journal of Clinical Medicine}, 10\penalty0 (12):\penalty0 2577, 2021.

\bibitem[Chen et~al.(2018)Chen, Zhu, Papandreou, Schroff, and Adam]{chen18eccv}
Liang-Chieh Chen, Yukun Zhu, George Papandreou, Florian Schroff, and Hartwig Adam.
\newblock Encoder-decoder with atrous separable convolution for semantic image segmentation.
\newblock In \emph{Proceedings of the European conference on computer vision (ECCV)}, pages 801--818, 2018.

\bibitem[Chen et~al.(2020)Chen, Lopes, Cheng, Collins, Cubuk, Zoph, Adam, and Shlens]{chen20eccv}
Liang-Chieh Chen, Raphael~Gontijo Lopes, Bowen Cheng, Maxwell~D Collins, Ekin~D Cubuk, Barret Zoph, Hartwig Adam, and Jonathon Shlens.
\newblock Naive-student: Leveraging semi-supervised learning in video sequences for urban scene segmentation.
\newblock In \emph{Computer Vision--ECCV 2020: 16th European Conference, Glasgow, UK, August 23--28, 2020, Proceedings, Part IX 16}, pages 695--714. Springer, 2020.

\bibitem[Cheng et~al.(2020)Cheng, Collins, Zhu, Liu, Huang, Adam, and Chen]{cheng20cvpr}
Bowen Cheng, Maxwell~D Collins, Yukun Zhu, Ting Liu, Thomas~S Huang, Hartwig Adam, and Liang-Chieh Chen.
\newblock Panoptic-deeplab: A simple, strong, and fast baseline for bottom-up panoptic segmentation.
\newblock In \emph{Proceedings of the IEEE/CVF conference on computer vision and pattern recognition}, pages 12475--12485, 2020.

\bibitem[Cheng et~al.(2021)Cheng, Schwing, and Kirillov]{cheng21neurips}
Bowen Cheng, Alex Schwing, and Alexander Kirillov.
\newblock Per-pixel classification is not all you need for semantic segmentation.
\newblock \emph{Advances in neural information processing systems}, 34:\penalty0 17864--17875, 2021.

\bibitem[Cheng et~al.(2022)Cheng, Misra, Schwing, Kirillov, and Girdhar]{cheng22cvpr}
Bowen Cheng, Ishan Misra, Alexander~G Schwing, Alexander Kirillov, and Rohit Girdhar.
\newblock Masked-attention mask transformer for universal image segmentation.
\newblock In \emph{Proceedings of the IEEE/CVF conference on computer vision and pattern recognition}, pages 1290--1299, 2022.

\bibitem[Cherti et~al.(2023)Cherti, Beaumont, Wightman, Wortsman, Ilharco, Gordon, Schuhmann, Schmidt, and Jitsev]{cherti23cvpr}
Mehdi Cherti, Romain Beaumont, Ross Wightman, Mitchell Wortsman, Gabriel Ilharco, Cade Gordon, Christoph Schuhmann, Ludwig Schmidt, and Jenia Jitsev.
\newblock Reproducible scaling laws for contrastive language-image learning.
\newblock In \emph{Proceedings of the IEEE/CVF Conference on Computer Vision and Pattern Recognition}, pages 2818--2829, 2023.

\bibitem[Cordts et~al.(2016)Cordts, Omran, Ramos, Rehfeld, Enzweiler, Benenson, Franke, Roth, and Schiele]{cordts16cvpr}
Marius Cordts, Mohamed Omran, Sebastian Ramos, Timo Rehfeld, Markus Enzweiler, Rodrigo Benenson, Uwe Franke, Stefan Roth, and Bernt Schiele.
\newblock The cityscapes dataset for semantic urban scene understanding.
\newblock In \emph{Proceedings of the IEEE conference on computer vision and pattern recognition}, pages 3213--3223, 2016.

\bibitem[de~Carvalho et~al.(2022)de~Carvalho, de~Carvalho~J{\'u}nior, Silva, de~Albuquerque, Santana, Borges, Gomes, and Guimar{\~a}es]{carvalho22rsensing}
Osmar Luiz~Ferreira de Carvalho, Osmar~Ab{\'\i}lio de Carvalho~J{\'u}nior, Cristiano Rosa~e Silva, Anesmar~Olino de Albuquerque, Nickolas~Castro Santana, Dibio~Leandro Borges, Roberto Arnaldo~Trancoso Gomes, and Renato~Fontes Guimar{\~a}es.
\newblock Panoptic segmentation meets remote sensing.
\newblock \emph{Remote Sensing}, 14\penalty0 (4):\penalty0 965, 2022.

\bibitem[Deng et~al.(2009)Deng, Dong, Socher, Li, Li, and Fei-Fei]{deng09cvpr}
Jia Deng, Wei Dong, Richard Socher, Li-Jia Li, Kai Li, and Li Fei-Fei.
\newblock Imagenet: A large-scale hierarchical image database.
\newblock In \emph{2009 IEEE conference on computer vision and pattern recognition}, pages 248--255. Ieee, 2009.

\bibitem[Ding et~al.(2022)Ding, Xue, Xia, and Dai]{ding22cvpr}
Jian Ding, Nan Xue, Gui-Song Xia, and Dengxin Dai.
\newblock Decoupling zero-shot semantic segmentation.
\newblock In \emph{Proceedings of the IEEE/CVF conference on computer vision and pattern recognition}, pages 11583--11592, 2022.

\bibitem[Dosovitskiy et~al.(2021)Dosovitskiy, Beyer, Kolesnikov, Weissenborn, Zhai, Unterthiner, Dehghani, Minderer, Heigold, Gelly, Uszkoreit, and Houlsby]{dosovitskiy21iclr}
Alexey Dosovitskiy, Lucas Beyer, Alexander Kolesnikov, Dirk Weissenborn, Xiaohua Zhai, Thomas Unterthiner, Mostafa Dehghani, Matthias Minderer, Georg Heigold, Sylvain Gelly, Jakob Uszkoreit, and Neil Houlsby.
\newblock An image is worth 16x16 words: Transformers for image recognition at scale.
\newblock 2021.

\bibitem[French et~al.(2020)French, Laine, Aila, Mackiewicz, and Finlayson]{french20bmvc}
Geoffrey French, Samuli Laine, Timo Aila, Michal Mackiewicz, and Graham~D. Finlayson.
\newblock Semi-supervised semantic segmentation needs strong, varied perturbations.
\newblock In \emph{31st British Machine Vision Conference 2020, {BMVC} 2020, Virtual Event, UK, September 7-10, 2020}, 2020.

\bibitem[Ghiasi et~al.(2022)Ghiasi, Gu, Cui, and Lin]{ghiasi22eccv}
Golnaz Ghiasi, Xiuye Gu, Yin Cui, and Tsung-Yi Lin.
\newblock Scaling open-vocabulary image segmentation with image-level labels.
\newblock In \emph{European Conference on Computer Vision}, pages 540--557. Springer, 2022.

\bibitem[Goodfellow et~al.(2014)Goodfellow, Pouget-Abadie, Mirza, Xu, Warde-Farley, Ozair, Courville, and Bengio]{goodfellow14neurips}
Ian Goodfellow, Jean Pouget-Abadie, Mehdi Mirza, Bing Xu, David Warde-Farley, Sherjil Ozair, Aaron Courville, and Yoshua Bengio.
\newblock Generative adversarial nets.
\newblock \emph{Advances in neural information processing systems}, 27, 2014.

\bibitem[Grubi{\v{s}}i{\'c} et~al.(2023)Grubi{\v{s}}i{\'c}, Or{\v{s}}i{\'c}, and {\v{S}}egvi{\'c}]{grubisic23sensors}
Ivan Grubi{\v{s}}i{\'c}, Marin Or{\v{s}}i{\'c}, and Sini{\v{s}}a {\v{S}}egvi{\'c}.
\newblock Revisiting consistency for semi-supervised semantic segmentation.
\newblock \emph{Sensors}, 23\penalty0 (2):\penalty0 940, 2023.

\bibitem[Gu et~al.(2022)Gu, Lin, Kuo, and Cui]{gu2022iclr}
Xiuye Gu, Tsung-Yi Lin, Weicheng Kuo, and Yin Cui.
\newblock Open-vocabulary object detection via vision and language knowledge distillation.
\newblock In \emph{International Conference on Learning Representations}, 2022.

\bibitem[He et~al.(2017)He, Gkioxari, Doll{\'a}r, and Girshick]{he17iccv}
Kaiming He, Georgia Gkioxari, Piotr Doll{\'a}r, and Ross Girshick.
\newblock Mask r-cnn.
\newblock In \emph{Proceedings of the IEEE international conference on computer vision}, pages 2961--2969, 2017.

\bibitem[Hoyer et~al.(2025)Hoyer, Tan, Naeem, Van~Gool, and Tombari]{hoyer24eccv}
Lukas Hoyer, David~Joseph Tan, Muhammad~Ferjad Naeem, Luc Van~Gool, and Federico Tombari.
\newblock Semivl: Semi-supervised semantic segmentation with vision-language guidance.
\newblock In \emph{European Conference on Computer Vision}, pages 257--275. Springer, 2025.

\bibitem[Hu et~al.(2021)Hu, Wei, Hu, Ye, Cui, and Wang]{hu21neurips}
Hanzhe Hu, Fangyun Wei, Han Hu, Qiwei Ye, Jinshi Cui, and Liwei Wang.
\newblock Semi-supervised semantic segmentation via adaptive equalization learning.
\newblock \emph{Advances in Neural Information Processing Systems}, 34:\penalty0 22106--22118, 2021.

\bibitem[Hu et~al.(2023)Hu, Chen, Cao, Zhang, Shu, Jiang, and Ji]{hu23iccv}
Jie Hu, Chen Chen, Liujuan Cao, Shengchuan Zhang, Annan Shu, Guannan Jiang, and Rongrong Ji.
\newblock Pseudo-label alignment for semi-supervised instance segmentation.
\newblock In \emph{Proceedings of the IEEE/CVF International Conference on Computer Vision}, pages 16337--16347, 2023.

\bibitem[Hu et~al.(2024)Hu, Jiang, and Schiele]{hu24cvpr}
Xinting Hu, Li Jiang, and Bernt Schiele.
\newblock Training vision transformers for semi-supervised semantic segmentation.
\newblock In \emph{Proceedings of the IEEE/CVF Conference on Computer Vision and Pattern Recognition}, pages 4007--4017, 2024.

\bibitem[Jia et~al.(2021)Jia, Yang, Xia, Chen, Parekh, Pham, Le, Sung, Li, and Duerig]{jia21pmlr}
Chao Jia, Yinfei Yang, Ye Xia, Yi-Ting Chen, Zarana Parekh, Hieu Pham, Quoc Le, Yun-Hsuan Sung, Zhen Li, and Tom Duerig.
\newblock Scaling up visual and vision-language representation learning with noisy text supervision.
\newblock In \emph{International conference on machine learning}, pages 4904--4916. PMLR, 2021.

\bibitem[Jiao et~al.(2023)Jiao, Wei, Wang, Zhao, and Shi]{jiao23neurips}
Siyu Jiao, Yunchao Wei, Yaowei Wang, Yao Zhao, and Humphrey Shi.
\newblock Learning mask-aware clip representations for zero-shot segmentation.
\newblock \emph{Advances in Neural Information Processing Systems}, 36:\penalty0 35631--35653, 2023.

\bibitem[Ke et~al.(2022)Ke, Aviles-Rivero, Pandey, Reddy, and Sch{\"o}nlieb]{ke22itip}
Rihuan Ke, Angelica~I Aviles-Rivero, Saurabh Pandey, Saikumar Reddy, and Carola-Bibiane Sch{\"o}nlieb.
\newblock A three-stage self-training framework for semi-supervised semantic segmentation.
\newblock \emph{IEEE Transactions on Image Processing}, 31:\penalty0 1805--1815, 2022.

\bibitem[Kingma and Ba(2015)]{kingma15iclr}
Diederik~P. Kingma and Jimmy Ba.
\newblock Adam: {A} method for stochastic optimization.
\newblock In \emph{3rd International Conference on Learning Representations, {ICLR} 2015, San Diego, CA, USA, May 7-9, 2015, Conference Track Proceedings}, 2015.

\bibitem[Kirillov et~al.(2019{\natexlab{a}})Kirillov, Girshick, He, and Doll{\'a}r]{kirillov19cvpr2}
Alexander Kirillov, Ross Girshick, Kaiming He, and Piotr Doll{\'a}r.
\newblock Panoptic feature pyramid networks.
\newblock In \emph{Proceedings of the IEEE/CVF conference on computer vision and pattern recognition}, pages 6399--6408, 2019{\natexlab{a}}.

\bibitem[Kirillov et~al.(2019{\natexlab{b}})Kirillov, He, Girshick, Rother, and Doll{\'a}r]{kirillov19cvpr}
Alexander Kirillov, Kaiming He, Ross Girshick, Carsten Rother, and Piotr Doll{\'a}r.
\newblock Panoptic segmentation.
\newblock In \emph{Proceedings of the IEEE/CVF conference on computer vision and pattern recognition}, pages 9404--9413, 2019{\natexlab{b}}.

\bibitem[Kirillov et~al.(2023)Kirillov, Mintun, Ravi, Mao, Rolland, Gustafson, Xiao, Whitehead, Berg, Lo, et~al.]{kirillov23iccv}
Alexander Kirillov, Eric Mintun, Nikhila Ravi, Hanzi Mao, Chloe Rolland, Laura Gustafson, Tete Xiao, Spencer Whitehead, Alexander~C Berg, Wan-Yen Lo, et~al.
\newblock Segment anything.
\newblock In \emph{Proceedings of the IEEE/CVF International Conference on Computer Vision}, pages 4015--4026, 2023.

\bibitem[Kuo et~al.(2023)Kuo, Cui, Gu, Piergiovanni, and Angelova]{kuo2023iclr}
Weicheng Kuo, Yin Cui, Xiuye Gu, AJ Piergiovanni, and Anelia Angelova.
\newblock Open-vocabulary object detection upon frozen vision and language models.
\newblock In \emph{The Eleventh International Conference on Learning Representations}, 2023.

\bibitem[Lai et~al.(2017)Lai, Huang, and Yang]{lai17neurips}
Wei-Sheng Lai, Jia-Bin Huang, and Ming-Hsuan Yang.
\newblock Semi-supervised learning for optical flow with generative adversarial networks.
\newblock \emph{Advances in neural information processing systems}, 30, 2017.

\bibitem[Lai et~al.(2021)Lai, Tian, Jiang, Liu, Zhao, Wang, and Jia]{lai21cvpr}
Xin Lai, Zhuotao Tian, Li Jiang, Shu Liu, Hengshuang Zhao, Liwei Wang, and Jiaya Jia.
\newblock Semi-supervised semantic segmentation with directional context-aware consistency.
\newblock In \emph{Proceedings of the IEEE/CVF Conference on Computer Vision and Pattern Recognition}, pages 1205--1214, 2021.

\bibitem[Lan et~al.(2024)Lan, Chen, Ke, Wang, Feng, and Zhang]{lan24eccv}
Mengcheng Lan, Chaofeng Chen, Yiping Ke, Xinjiang Wang, Litong Feng, and Wayne Zhang.
\newblock Proxyclip: Proxy attention improves clip for open-vocabulary segmentation.
\newblock In \emph{ECCV}, 2024.

\bibitem[Li et~al.(2022)Li, Weinberger, Belongie, Koltun, and Ranftl]{li22iclr}
Boyi Li, Kilian~Q Weinberger, Serge Belongie, Vladlen Koltun, and Rene Ranftl.
\newblock Language-driven semantic segmentation.
\newblock In \emph{International Conference on Learning Representations}, 2022.

\bibitem[Li et~al.(2023)Li, Zhang, Xu, Liu, Zhang, Ni, and Shum]{li23cvpr}
Feng Li, Hao Zhang, Huaizhe Xu, Shilong Liu, Lei Zhang, Lionel~M Ni, and Heung-Yeung Shum.
\newblock Mask dino: Towards a unified transformer-based framework for object detection and segmentation.
\newblock In \emph{Proceedings of the IEEE/CVF Conference on Computer Vision and Pattern Recognition}, pages 3041--3050, 2023.

\bibitem[Li et~al.(2018)Li, Arnab, and Torr]{li18eccv}
Qizhu Li, Anurag Arnab, and Philip~HS Torr.
\newblock Weakly-and semi-supervised panoptic segmentation.
\newblock In \emph{Proceedings of the European conference on computer vision (ECCV)}, pages 102--118, 2018.

\bibitem[Liang et~al.(2023{\natexlab{a}})Liang, Wang, Miao, and Yang]{liang23iccv}
Chen Liang, Wenguan Wang, Jiaxu Miao, and Yi Yang.
\newblock Logic-induced diagnostic reasoning for semi-supervised semantic segmentation.
\newblock In \emph{Proceedings of the IEEE/CVF International Conference on Computer Vision}, pages 16197--16208, 2023{\natexlab{a}}.

\bibitem[Liang et~al.(2023{\natexlab{b}})Liang, Wu, Dai, Li, Zhao, Zhang, Zhang, Vajda, and Marculescu]{liang23cvpr}
Feng Liang, Bichen Wu, Xiaoliang Dai, Kunpeng Li, Yinan Zhao, Hang Zhang, Peizhao Zhang, Peter Vajda, and Diana Marculescu.
\newblock Open-vocabulary semantic segmentation with mask-adapted clip.
\newblock In \emph{Proceedings of the IEEE/CVF conference on computer vision and pattern recognition}, pages 7061--7070, 2023{\natexlab{b}}.

\bibitem[Lin et~al.(2014)Lin, Maire, Belongie, Hays, Perona, Ramanan, Doll{\'a}r, and Zitnick]{lin14eccv}
Tsung-Yi Lin, Michael Maire, Serge Belongie, James Hays, Pietro Perona, Deva Ramanan, Piotr Doll{\'a}r, and C~Lawrence Zitnick.
\newblock Microsoft coco: Common objects in context.
\newblock In \emph{Computer Vision--ECCV 2014: 13th European Conference, Zurich, Switzerland, September 6-12, 2014, Proceedings, Part V 13}, pages 740--755. Springer, 2014.

\bibitem[Liu et~al.(2022)Liu, Mao, Wu, Feichtenhofer, Darrell, and Xie]{liu22cvpr}
Zhuang Liu, Hanzi Mao, Chao-Yuan Wu, Christoph Feichtenhofer, Trevor Darrell, and Saining Xie.
\newblock A convnet for the 2020s.
\newblock In \emph{Proceedings of the IEEE/CVF conference on computer vision and pattern recognition}, pages 11976--11986, 2022.

\bibitem[Loshchilov and Hutter(2019)]{loshchilov19iclr}
Ilya Loshchilov and Frank Hutter.
\newblock Decoupled weight decay regularization.
\newblock In \emph{7th International Conference on Learning Representations, {ICLR} 2019, New Orleans, LA, USA, May 6-9, 2019}. OpenReview.net, 2019.

\bibitem[Mai et~al.(2024)Mai, Sun, Zhang, and Wu]{mai24cvpr}
Huayu Mai, Rui Sun, Tianzhu Zhang, and Feng Wu.
\newblock Rankmatch: Exploring the better consistency regularization for semi-supervised semantic segmentation.
\newblock In \emph{Proceedings of the IEEE/CVF Conference on Computer Vision and Pattern Recognition}, pages 3391--3401, 2024.

\bibitem[Martinovi{\'c} et~al.(2024)Martinovi{\'c}, {\v{S}}ari{\'c}, and {\v{S}}egvi{\'c}]{martinovic2024mc}
Ivan Martinovi{\'c}, Josip {\v{S}}ari{\'c}, and Sini{\v{s}}a {\v{S}}egvi{\'c}.
\newblock Mc-panda: Mask confidence for panoptic domain adaptation.
\newblock In \emph{European Conference on Computer Vision}, pages 167--185. Springer, 2024.

\bibitem[Mittal et~al.(2019)Mittal, Tatarchenko, and Brox]{mittal19pami}
Sudhanshu Mittal, Maxim Tatarchenko, and Thomas Brox.
\newblock Semi-supervised semantic segmentation with high-and low-level consistency.
\newblock \emph{IEEE transactions on pattern analysis and machine intelligence}, 43\penalty0 (4):\penalty0 1369--1379, 2019.

\bibitem[Ouali et~al.(2020)Ouali, Hudelot, and Tami]{ouali20cvpr}
Yassine Ouali, C{\'e}line Hudelot, and Myriam Tami.
\newblock Semi-supervised semantic segmentation with cross-consistency training.
\newblock In \emph{Proceedings of the IEEE/CVF conference on computer vision and pattern recognition}, pages 12674--12684, 2020.

\bibitem[Qi et~al.(2022)Qi, Kuen, Lin, Gu, Rao, Li, Guo, Wen, Yang, and Jia]{qi22eccv}
Lu Qi, Jason Kuen, Zhe Lin, Jiuxiang Gu, Fengyun Rao, Dian Li, Weidong Guo, Zhen Wen, Ming-Hsuan Yang, and Jiaya Jia.
\newblock Ca-ssl: Class-agnostic semi-supervised learning for detection and segmentation.
\newblock In \emph{European Conference on Computer Vision}, pages 59--77. Springer, 2022.

\bibitem[Qi et~al.(2019)Qi, Wang, Qin, and Li]{qi19cvpr}
Mengshi Qi, Yunhong Wang, Jie Qin, and Annan Li.
\newblock Ke-gan: Knowledge embedded generative adversarial networks for semi-supervised scene parsing.
\newblock In \emph{Proceedings of the IEEE/CVF Conference on Computer Vision and Pattern Recognition}, pages 5237--5246, 2019.

\bibitem[Qiao et~al.(2018)Qiao, Shen, Zhang, Wang, and Yuille]{qiao18eccv}
Siyuan Qiao, Wei Shen, Zhishuai Zhang, Bo Wang, and Alan Yuille.
\newblock Deep co-training for semi-supervised image recognition.
\newblock In \emph{Proceedings of the european conference on computer vision (eccv)}, pages 135--152, 2018.

\bibitem[Radford et~al.(2021)Radford, Kim, Hallacy, Ramesh, Goh, Agarwal, Sastry, Askell, Mishkin, Clark, et~al.]{radford21icml}
Alec Radford, Jong~Wook Kim, Chris Hallacy, Aditya Ramesh, Gabriel Goh, Sandhini Agarwal, Girish Sastry, Amanda Askell, Pamela Mishkin, Jack Clark, et~al.
\newblock Learning transferable visual models from natural language supervision.
\newblock In \emph{International conference on machine learning}, pages 8748--8763. PMLR, 2021.

\bibitem[Ramesh et~al.(2021)Ramesh, Pavlov, Goh, Gray, Voss, Radford, Chen, and Sutskever]{ramesh21icml}
Aditya Ramesh, Mikhail Pavlov, Gabriel Goh, Scott Gray, Chelsea Voss, Alec Radford, Mark Chen, and Ilya Sutskever.
\newblock Zero-shot text-to-image generation.
\newblock In \emph{International conference on machine learning}, pages 8821--8831. Pmlr, 2021.

\bibitem[Ridnik et~al.(2021)Ridnik, Baruch, Noy, and Zelnik]{ridnik21neurips}
Tal Ridnik, Emanuel~Ben Baruch, Asaf Noy, and Lihi Zelnik.
\newblock Imagenet-21k pretraining for the masses.
\newblock In \emph{Proceedings of the Neural Information Processing Systems Track on Datasets and Benchmarks 1, NeurIPS Datasets and Benchmarks 2021, December 2021, virtual}, 2021.

\bibitem[Rombach et~al.(2022)Rombach, Blattmann, Lorenz, Esser, and Ommer]{rombach22cvpr}
Robin Rombach, Andreas Blattmann, Dominik Lorenz, Patrick Esser, and Bj{\"o}rn Ommer.
\newblock High-resolution image synthesis with latent diffusion models.
\newblock In \emph{Proceedings of the IEEE/CVF conference on computer vision and pattern recognition}, pages 10684--10695, 2022.

\bibitem[Sakaridis et~al.(2021)Sakaridis, Dai, and Van~Gool]{sakaridis21iccv}
Christos Sakaridis, Dengxin Dai, and Luc Van~Gool.
\newblock Acdc: The adverse conditions dataset with correspondences for semantic driving scene understanding.
\newblock In \emph{Proceedings of the IEEE/CVF International Conference on Computer Vision}, pages 10765--10775, 2021.

\bibitem[Schuhmann et~al.(2022)Schuhmann, Beaumont, Vencu, Gordon, Wightman, Cherti, Coombes, Katta, Mullis, Wortsman, et~al.]{schuhmann22neurips}
Christoph Schuhmann, Romain Beaumont, Richard Vencu, Cade Gordon, Ross Wightman, Mehdi Cherti, Theo Coombes, Aarush Katta, Clayton Mullis, Mitchell Wortsman, et~al.
\newblock Laion-5b: An open large-scale dataset for training next generation image-text models.
\newblock \emph{Advances in Neural Information Processing Systems}, 35:\penalty0 25278--25294, 2022.

\bibitem[Sohn et~al.(2020)Sohn, Berthelot, Carlini, Zhang, Zhang, Raffel, Cubuk, Kurakin, and Li]{sohn20neurips}
Kihyuk Sohn, David Berthelot, Nicholas Carlini, Zizhao Zhang, Han Zhang, Colin Raffel, Ekin~Dogus Cubuk, Alexey Kurakin, and Chun{-}Liang Li.
\newblock Fixmatch: Simplifying semi-supervised learning with consistency and confidence.
\newblock In \emph{Advances in Neural Information Processing Systems 33: Annual Conference on Neural Information Processing Systems 2020, NeurIPS 2020, December 6-12, 2020, virtual}, 2020.

\bibitem[Souly et~al.(2017)Souly, Spampinato, and Shah]{souly17iccv}
Nasim Souly, Concetto Spampinato, and Mubarak Shah.
\newblock Semi supervised semantic segmentation using generative adversarial network.
\newblock In \emph{Proceedings of the IEEE international conference on computer vision}, pages 5688--5696, 2017.

\bibitem[Sun et~al.(2024)Sun, Yang, Zhang, Cheng, and Hou]{sun24cvpr}
Boyuan Sun, Yuqi Yang, Le Zhang, Ming-Ming Cheng, and Qibin Hou.
\newblock Corrmatch: Label propagation via correlation matching for semi-supervised semantic segmentation.
\newblock In \emph{Proceedings of the IEEE/CVF Conference on Computer Vision and Pattern Recognition}, pages 3097--3107, 2024.

\bibitem[Tarvainen and Valpola(2017)]{tarvainen17neurips}
Antti Tarvainen and Harri Valpola.
\newblock Mean teachers are better role models: Weight-averaged consistency targets improve semi-supervised deep learning results.
\newblock \emph{Advances in neural information processing systems}, 30, 2017.

\bibitem[Vaswani et~al.(2017)Vaswani, Shazeer, Parmar, Uszkoreit, Jones, Gomez, Kaiser, and Polosukhin]{vaswani17neurips}
Ashish Vaswani, Noam Shazeer, Niki Parmar, Jakob Uszkoreit, Llion Jones, Aidan~N Gomez, \L~ukasz Kaiser, and Illia Polosukhin.
\newblock Attention is all you need.
\newblock In \emph{Advances in Neural Information Processing Systems}, 2017.

\bibitem[Wang et~al.(2024{\natexlab{a}})Wang, Mei, and Yuille]{wang24eccv}
Feng Wang, Jieru Mei, and Alan Yuille.
\newblock Sclip: Rethinking self-attention for dense vision-language inference.
\newblock In \emph{European Conference on Computer Vision}, pages 315--332. Springer, 2024{\natexlab{a}}.

\bibitem[Wang et~al.(2021)Wang, Zhu, Adam, Yuille, and Chen]{wang21cvpr}
Huiyu Wang, Yukun Zhu, Hartwig Adam, Alan Yuille, and Liang-Chieh Chen.
\newblock Max-deeplab: End-to-end panoptic segmentation with mask transformers.
\newblock In \emph{Proceedings of the IEEE/CVF conference on computer vision and pattern recognition}, pages 5463--5474, 2021.

\bibitem[Wang et~al.(2024{\natexlab{b}})Wang, Zhang, Li, and Li]{wang24cvpr}
Haonan Wang, Qixiang Zhang, Yi Li, and Xiaomeng Li.
\newblock Allspark: Reborn labeled features from unlabeled in transformer for semi-supervised semantic segmentation.
\newblock In \emph{Proceedings of the IEEE/CVF Conference on Computer Vision and Pattern Recognition}, pages 3627--3636, 2024{\natexlab{b}}.

\bibitem[Wang et~al.(2024{\natexlab{c}})Wang, Bai, Yu, Zhao, and Xiao]{wang24cvpr2}
Xiaoyang Wang, Huihui Bai, Limin Yu, Yao Zhao, and Jimin Xiao.
\newblock Towards the uncharted: Density-descending feature perturbation for semi-supervised semantic segmentation.
\newblock In \emph{Proceedings of the IEEE/CVF Conference on Computer Vision and Pattern Recognition}, pages 3303--3312, 2024{\natexlab{c}}.

\bibitem[Wortsman et~al.(2022)Wortsman, Ilharco, Kim, Li, Kornblith, Roelofs, Lopes, Hajishirzi, Farhadi, Namkoong, et~al.]{wortsman22cvpr}
Mitchell Wortsman, Gabriel Ilharco, Jong~Wook Kim, Mike Li, Simon Kornblith, Rebecca Roelofs, Raphael~Gontijo Lopes, Hannaneh Hajishirzi, Ali Farhadi, Hongseok Namkoong, et~al.
\newblock Robust fine-tuning of zero-shot models.
\newblock In \emph{Proceedings of the IEEE/CVF conference on computer vision and pattern recognition}, pages 7959--7971, 2022.

\bibitem[Wu et~al.(2023)Wu, Fang, He, He, Ma, and Zhong]{wu23pami}
Linshan Wu, Leyuan Fang, Xingxin He, Min He, Jiayi Ma, and Zhun Zhong.
\newblock Querying labeled for unlabeled: Cross-image semantic consistency guided semi-supervised semantic segmentation.
\newblock \emph{IEEE Transactions on Pattern Analysis and Machine Intelligence}, 45\penalty0 (7):\penalty0 8827--8844, 2023.

\bibitem[Xie et~al.(2020)Xie, Dai, Hovy, Luong, and Le]{xie20neurips}
Qizhe Xie, Zihang Dai, Eduard Hovy, Thang Luong, and Quoc Le.
\newblock Unsupervised data augmentation for consistency training.
\newblock \emph{Advances in neural information processing systems}, 33:\penalty0 6256--6268, 2020.

\bibitem[Xu et~al.(2023)Xu, Liu, Vahdat, Byeon, Wang, and De~Mello]{xu23cvpr}
Jiarui Xu, Sifei Liu, Arash Vahdat, Wonmin Byeon, Xiaolong Wang, and Shalini De~Mello.
\newblock Open-vocabulary panoptic segmentation with text-to-image diffusion models.
\newblock In \emph{Proceedings of the IEEE/CVF Conference on Computer Vision and Pattern Recognition}, pages 2955--2966, 2023.

\bibitem[Xu et~al.(2021)Xu, Zhang, Hu, Wang, Wang, Wei, Bai, and Liu]{xu21iccv}
Mengde Xu, Zheng Zhang, Han Hu, Jianfeng Wang, Lijuan Wang, Fangyun Wei, Xiang Bai, and Zicheng Liu.
\newblock End-to-end semi-supervised object detection with soft teacher.
\newblock In \emph{Proceedings of the IEEE/CVF international conference on computer vision}, pages 3060--3069, 2021.

\bibitem[Yang et~al.(2023{\natexlab{a}})Yang, Nagrani, Seo, Miech, Pont-Tuset, Laptev, Sivic, and Schmid]{yang23cvpr2}
Antoine Yang, Arsha Nagrani, Paul~Hongsuck Seo, Antoine Miech, Jordi Pont-Tuset, Ivan Laptev, Josef Sivic, and Cordelia Schmid.
\newblock Vid2seq: Large-scale pretraining of a visual language model for dense video captioning.
\newblock In \emph{CVPR}, 2023{\natexlab{a}}.

\bibitem[Yang et~al.(2022)Yang, Zhuo, Qi, Shi, and Gao]{yang22cvpr}
Lihe Yang, Wei Zhuo, Lei Qi, Yinghuan Shi, and Yang Gao.
\newblock St++: Make self-training work better for semi-supervised semantic segmentation.
\newblock In \emph{Proceedings of the IEEE/CVF conference on computer vision and pattern recognition}, pages 4268--4277, 2022.

\bibitem[Yang et~al.(2023{\natexlab{b}})Yang, Qi, Feng, Zhang, and Shi]{yang23cvpr}
Lihe Yang, Lei Qi, Litong Feng, Wayne Zhang, and Yinghuan Shi.
\newblock Revisiting weak-to-strong consistency in semi-supervised semantic segmentation.
\newblock In \emph{Proceedings of the IEEE/CVF Conference on Computer Vision and Pattern Recognition}, pages 7236--7246, 2023{\natexlab{b}}.

\bibitem[Yu et~al.(2022)Yu, Wang, Qiao, Collins, Zhu, Adam, Yuille, and Chen]{yu22eccv}
Qihang Yu, Huiyu Wang, Siyuan Qiao, Maxwell Collins, Yukun Zhu, Hartwig Adam, Alan Yuille, and Liang-Chieh Chen.
\newblock k-means mask transformer.
\newblock In \emph{European Conference on Computer Vision}, pages 288--307. Springer, 2022.

\bibitem[Yu et~al.(2023)Yu, He, Deng, Shen, and Chen]{yu23neurips}
Qihang Yu, Ju He, Xueqing Deng, Xiaohui Shen, and Liang-Chieh Chen.
\newblock Convolutions die hard: Open-vocabulary segmentation with single frozen convolutional clip.
\newblock \emph{Advances in Neural Information Processing Systems}, 36:\penalty0 32215--32234, 2023.

\bibitem[Yuan et~al.(2021)Yuan, Liu, Shen, Wang, and Li]{yuan21iccv}
Jianlong Yuan, Yifan Liu, Chunhua Shen, Zhibin Wang, and Hao Li.
\newblock A simple baseline for semi-supervised semantic segmentation with strong data augmentation.
\newblock In \emph{Proceedings of the IEEE/CVF International Conference on Computer Vision}, pages 8229--8238, 2021.

\bibitem[Yun et~al.(2019)Yun, Han, Oh, Chun, Choe, and Yoo]{yun19iccv}
Sangdoo Yun, Dongyoon Han, Seong~Joon Oh, Sanghyuk Chun, Junsuk Choe, and Youngjoon Yoo.
\newblock Cutmix: Regularization strategy to train strong classifiers with localizable features.
\newblock In \emph{Proceedings of the IEEE/CVF international conference on computer vision}, pages 6023--6032, 2019.

\bibitem[Zendel et~al.(2022)Zendel, Sch{\"o}rghuber, Rainer, Murschitz, and Beleznai]{zendel22cvpr}
Oliver Zendel, Matthias Sch{\"o}rghuber, Bernhard Rainer, Markus Murschitz, and Csaba Beleznai.
\newblock Unifying panoptic segmentation for autonomous driving.
\newblock In \emph{Proceedings of the IEEE/CVF Conference on Computer Vision and Pattern Recognition}, pages 21351--21360, 2022.

\bibitem[Zhai et~al.(2023)Zhai, Mustafa, Kolesnikov, and Beyer]{zhai23iccv}
Xiaohua Zhai, Basil Mustafa, Alexander Kolesnikov, and Lucas Beyer.
\newblock Sigmoid loss for language image pre-training.
\newblock In \emph{Proceedings of the IEEE/CVF International Conference on Computer Vision}, pages 11975--11986, 2023.

\bibitem[Zhong et~al.(2021)Zhong, Yuan, Wu, Yuan, Peng, and Wang]{zhong21iccv}
Yuanyi Zhong, Bodi Yuan, Hong Wu, Zhiqiang Yuan, Jian Peng, and Yu-Xiong Wang.
\newblock Pixel contrastive-consistent semi-supervised semantic segmentation.
\newblock In \emph{Proceedings of the IEEE/CVF International Conference on Computer Vision}, pages 7273--7282, 2021.

\bibitem[Zhou et~al.(2017)Zhou, Zhao, Puig, Fidler, Barriuso, and Torralba]{zhou17cvpr}
Bolei Zhou, Hang Zhao, Xavier Puig, Sanja Fidler, Adela Barriuso, and Antonio Torralba.
\newblock Scene parsing through ade20k dataset.
\newblock In \emph{Proceedings of the IEEE conference on computer vision and pattern recognition}, pages 633--641, 2017.

\bibitem[Zhou et~al.(2022)Zhou, Loy, and Dai]{zhou22eccv}
Chong Zhou, Chen~Change Loy, and Bo Dai.
\newblock Extract free dense labels from clip.
\newblock In \emph{European Conference on Computer Vision}, pages 696--712. Springer, 2022.

\bibitem[Zou et~al.(2021)Zou, Zhang, Zhang, Li, Bian, Huang, and Pfister]{zou21iclr}
Yuliang Zou, Zizhao Zhang, Han Zhang, Chun-Liang Li, Xiao Bian, Jia-Bin Huang, and Tomas Pfister.
\newblock Pseudoseg: Designing pseudo labels for semantic segmentation.
\newblock In \emph{International Conference on Learning Representations}, 2021.

\bibitem[{\v{Z}}ust et~al.(2023){\v{Z}}ust, Per{\v{s}}, and Kristan]{zust23iccv}
Lojze {\v{Z}}ust, Janez Per{\v{s}}, and Matej Kristan.
\newblock Lars: A diverse panoptic maritime obstacle detection dataset and benchmark.
\newblock In \emph{Proceedings of the IEEE/CVF International Conference on Computer Vision}, pages 20304--20314, 2023.

\end{thebibliography}
